\pdfoutput=1
\documentclass{article}
\usepackage{iclr2027_conference,times}
\usepackage{enumitem}
\usepackage{visual/preprint}
\hypersetup{pdftitle={Act on the Low Frequencies: Prefix Execution of Vision-Language-Action Chunks}}
\hypersetup{pdfauthor={Ziming Wang, Jingmin Zhu, Chutian Wang, Wenhao Zhang, Xiuyu Li}}
\author{\parbox{\dimexpr\textwidth-2\tabcolsep\relax}{\raggedright
Ziming Wang\textsuperscript{1}\quad Jingmin Zhu\textsuperscript{2}\quad Chutian Wang\textsuperscript{3}\\Wenhao Zhang\textsuperscript{4}\quad Xiuyu Li\textsuperscript{5}\\[0.6em]
\normalfont\small
\textsuperscript{1}The Hong Kong University of Science and Technology\\\textsuperscript{2}Monash University\\\textsuperscript{3}Sun Yat-sen University\\\textsuperscript{4}Xidian University\\\textsuperscript{5}Renmin University of China}}
\newcommand{\preprintauthornames}{Ziming Wang\textsuperscript{1}, Jingmin Zhu\textsuperscript{2}, Chutian Wang\textsuperscript{3}, Wenhao Zhang\textsuperscript{4}, and Xiuyu Li\textsuperscript{5}}
\newcommand{\preprintaffiliations}{\textsuperscript{1}The Hong Kong University of Science and Technology, \textsuperscript{2}Monash University, \textsuperscript{3}Sun Yat-sen University, \textsuperscript{4}Xidian University, \textsuperscript{5}Renmin University of China}

\theoremstyle{plain}
\newtheorem{result}{Result}
\newtheorem{theorem}[result]{Theorem}
\newtheorem{proposition}[result]{Proposition}

\crefname{result}{Result}{Results}

\newcommand{\dist}{\operatorname{dist}}
\newcommand{\pairgap}{PairGap}

\newcommand{\PairedEdgeCount}{24}
\newcommand{\PairedSeedsHigh}{64}
\newcommand{\PairedSeedsLow}{27}
\newcommand{\PairedShortFortyEight}{10}
\newcommand{\PairedShortSixtyFour}{19}

\newcommand{\DepthCertificatePrefixBits}{560}
\newcommand{\DepthCertificateFullBits}{1,328}
\newcommand{\DepthCertificateRelationalLow}{0.12}
\newcommand{\DepthCertificateRelationalHigh}{0.26}
\newcommand{\DepthCertificateDepthEdgeHigh}{0.006}
\newcommand{\RelationDepthDiscordance}{0.036}
\newcommand{\RelationCandidateDiscordance}{0.059}
\newcommand{\RelationSaturatedStates}{73}

\newcommand{\LaraReconstructionErrorMaxTwo}{6.1}
\newcommand{\LaraReconstructionErrorMaxEight}{4.7}
\newcommand{\LaraDctShareTwo}{99.69}
\newcommand{\LaraKltShareTwo}{99.86}
\newcommand{\LaraSubspaceOverlapTwo}{0.992}
\newcommand{\LaraLagOneLow}{0.82}
\newcommand{\LaraLagOneHigh}{0.89}
\newcommand{\LaraDemeanedLagOneLow}{0.43}
\newcommand{\LaraDemeanedLagOneHigh}{0.56}
\newcommand{\LaraFastTwoRowSaving}{22}
\newcommand{\FlexPiReconstructionErrorMaxTen}{4.1}
\newcommand{\FlexPiReconstructionErrorAllTen}{3.3}
\newcommand{\FlexPiReconstructionErrorMaxTwo}{10.0}
\newcommand{\FlexPiDctShareTwo}{99.31}
\newcommand{\FlexPiKltShareTwo}{99.78}
\newcommand{\FlexPiReconstructionErrorTwo}{5.6}
\newcommand{\FlexPiSubspaceOverlapTwo}{0.990}
\newcommand{\FlexPiLagOneLow}{0.88}
\newcommand{\FlexPiLagOneHigh}{0.91}
\newcommand{\FlexPiDemeanedLagOneLow}{0.61}
\newcommand{\FlexPiDemeanedLagOneHigh}{0.71}
\newcommand{\FlexPiFastTwoRowSaving}{27}
\newcommand{\LadderFlexPiPrefixSuccesses}{255}
\newcommand{\LadderFlexPiFullSuccesses}{253}
\newcommand{\LadderLaraTrajectories}{256}
\newcommand{\LadderLaraTwoSuccesses}{254}
\newcommand{\LadderLaraUnquantizedSuccesses}{253}
\newcommand{\PanelTwoRowRollouts}{10,240}
\newcommand{\PanelTwoRowFailures}{386}
\newcommand{\PanelEightRowFailures}{416}
\newcommand{\OftGoalFreqOneVersusFreqTwoWins}{13}
\newcommand{\OftGoalFreqOneVersusFreqTwoLosses}{9}
\newcommand{\OftGoalFreqOneVersusFreqTwoP}{0.52}
\newcommand{\OftGoalFreqOneVersusQuantizedWins}{11}
\newcommand{\OftGoalFreqOneVersusQuantizedLosses}{7}
\newcommand{\OftGoalFreqOneVersusQuantizedP}{0.48}
\newcommand{\OftSpatialFreqOneVersusFreqTwoWins}{3}
\newcommand{\OftSpatialFreqOneVersusFreqTwoLosses}{8}
\newcommand{\OftSpatialFreqOneVersusFreqTwoP}{0.23}

\newcommand{\CertificateAbsoluteBoundFortyEight}{0.0740}
\newcommand{\CertificateAbsoluteFirstBudget}{48}
\newcommand{\CertificateAbsoluteRollouts}{48}
\newcommand{\CertificateFigureRegret}{0.076}
\newcommand{\CertificateHeldOutMax}{0.034}
\newcommand{\CertificateReceiptRequests}{eight}
\newcommand{\CertificateUnifiedBoundFortyEight}{0.1003}
\newcommand{\CertificateUnifiedFirstBudget}{64}
\newcommand{\CertificateUnifiedRollouts}{319}

\newcommand{\EnergyKeptTwoHigh}{99.7}
\newcommand{\EnergyKeptTwoLow}{98.8}
\newcommand{\EnergySuccessPFour}{0.29}
\newcommand{\EnergySuccessPTwo}{0.84}
\newcommand{\EnergySuccessRhoFour}{0.25}
\newcommand{\EnergySuccessRhoTwo}{0.05}
\newcommand{\EnergySuccessTasks}{20}
\newcommand{\TeaserCalls}{17}
\newcommand{\TeaserMaxShare}{2.9}
\newcommand{\TeaserZeroCalls}{13}

\newcommand{\FixedTwoRowSavingEight}{75}
\newcommand{\FixedTwoRowSavingTen}{80}
\newcommand{\GripperGoalDelta}{+0.2}
\newcommand{\GripperGoalLosses}{0}
\newcommand{\GripperGoalP}{1.0}
\newcommand{\GripperGoalWins}{1}
\newcommand{\GripperSpatialDelta}{\ensuremath{-}1.0}
\newcommand{\GripperSpatialLosses}{7}
\newcommand{\GripperSpatialP}{0.34}
\newcommand{\GripperSpatialWins}{3}
\newcommand{\OftFastChunks}{36,834}
\newcommand{\OftFastFixedBits}{448}
\newcommand{\OftFastFourRatio}{0.87}
\newcommand{\OftFastFullBits}{104.5}
\newcommand{\OftFastFullShare}{0.23}
\newcommand{\OftFastTwoRatio}{0.71}
\newcommand{\OftFastTwoSaving}{29}

\newcommand{\SharedSeedCertifiedAudit}{0}
\newcommand{\SharedSeedCertifiedBound}{0.088}
\newcommand{\SharedSeedCertifiedRollouts}{256}
\newcommand{\SharedSeedCertifiedStates}{2}
\newcommand{\SharedSeedCompletedStates}{2 and 4}
\newcommand{\SharedSeedControls}{4}
\newcommand{\SharedSeedControlsReproduced}{4}
\newcommand{\SharedSeedNewRollouts}{319}
\newcommand{\SharedSeedOpenBound}{0.146}
\newcommand{\SharedSeedOpenFailures}{2}
\newcommand{\SharedSeedOpenStates}{4}

\newcommand{\OftLiberoTenRerunIdentical}{265}
\newcommand{\OftLiberoTenRerunLosses}{14}
\newcommand{\OftLiberoTenRerunSuccess}{93.0}
\newcommand{\OftLiberoTenRerunWins}{10}

\newcommand{\OftLiberoTenTimeRerunDelta}{\ensuremath{-}5.6}
\newcommand{\OftLiberoTenTimeRerunLosses}{41}
\newcommand{\OftLiberoTenTimeRerunWins}{13}

\newcommand{\GoalSeedEightTwoRowsP}{0.17}

\newcommand{\GoalSeedNineTwoRowsP}{0.0039}
\newcommand{\GoalThreeSeedFourRowsP}{1.0\times10^{-5}}
\newcommand{\GoalThreeSeedTwoRowsP}{0.0015}

\newcommand{\MainLiberoTenQuantizedDelta}{\ensuremath{-}1.6}
\newcommand{\MainLiberoTenTwoRowsDrop}{2.2}
\newcommand{\MainLiberoTenTwoRowsLosses}{25}
\newcommand{\MainLiberoTenTwoRowsWins}{14}
\newcommand{\MainSpatialTwoRowsLosses}{4}
\newcommand{\MainSpatialTwoRowsWins}{5}
\newcommand{\SecondSeedLiberoTenFourRowsP}{9.8\times10^{-4}}

\newcommand{\SecondSeedSpatialTwoRowsP}{0.0020}

\newcommand{\PiFiveEnergyP}{0.33}
\newcommand{\PiFiveEnergySpearman}{0.16}
\newcommand{\PiFiveEnergyTasks}{40}
\newcommand{\PiFiveFifthRowShare}{0.0003}

\newcommand{\PiFiveFullBits}{1,112}

\newcommand{\PiFiveSeedEightEnergyP}{0.85}
\newcommand{\PiFiveSeedEightEnergySpearman}{\ensuremath{-}0.03}

\newcommand{\PiFiveSeedSevenPolicyHighMs}{101}
\newcommand{\PiFiveSeedSevenPolicyLowMs}{54}

\newcommand{\CallTimeCalls}{119,038}
\newcommand{\CallTimeCodecMedianMs}{0.31}
\newcommand{\CallTimePolicyMedianMs}{62}
\newcommand{\CallTimePolicyNinetyMs}{122}
\newcommand{\CodecBreakEvenMbps}{2}
\newcommand{\CoderCalls}{38,011}
\newcommand{\CoderFullBestStream}{131}
\newcommand{\CoderFullFast}{104}
\newcommand{\CoderKltGapPercent}{0.04}
\newcommand{\CoderTenFourBinningBest}{192}
\newcommand{\CoderTenFourRowsBest}{106}
\newcommand{\CoderTenThreeBinningBest}{120}

\newcommand{\CoderTwoBestStream}{99}
\newcommand{\CoderTwoFast}{74}
\newcommand{\CoderTwoOverFull}{0.76}
\newcommand{\CoderTwoSaving}{24}
\newcommand{\LinkBreakEvenKbps}{21}
\newcommand{\RateSuccessGoalBinningSmallestPass}{224}

\newcommand{\RateSuccessLiberoTenCosineSmallestPass}{224}

\newcommand{\RateSuccessSpatialBinningSmallestPass}{448}
\newcommand{\RateSuccessSpatialCosineFourP}{0.031}
\newcommand{\RateSuccessSpatialCosineSmallestPass}{112}

\newcommand{\FlexPiFailureFreeBound}{0.109}
\newcommand{\FlexPiHardAbsolute}{0.999}
\newcommand{\FlexPiHardInit}{49}
\newcommand{\FlexPiHardTask}{6}
\newcommand{\FlexPiHardUnified}{0.407}

\newcommand{\FlexPiPairedDepthTwenty}{34}
\newcommand{\FlexPiReceiptBits}{3,488}
\newcommand{\FlexPiReceiptFifteen}{34}

\newcommand{\FlexPiReceiveFullBits}{5,312}
\newcommand{\FlexPiReceiveFullFifteen}{32}

\newcommand{\FlexPiUnifiedFifteen}{34}

\newcommand{\OneCandidateFlexPiDepthMax}{0.010}
\newcommand{\OneCandidateGaussianDepthMax}{0.076}
\newcommand{\OneCandidateLaraDepthMax}{0.012}

\title{Act on the Low Frequencies:\\Prefix Execution of \mbox{Vision-Language-Action} Chunks}
\preprintlink{\faGithub}{Code and data}{https://github.com/ZenAlexa/act-on-the-low-frequencies}
\iclrfinalcopy
\begin{document}
\lhead{Preprint}
\begin{abstract}
Vision-language-action (VLA) policies control a robot by predicting short sequences of future actions, called action
chunks. When the policy runs on a remote server, every chunk travels to the robot as a message. Existing ways to shorten
this message compress the chunk so that it can be reconstructed closely, or execute fewer of its steps per call, and how
much of a chunk the robot needs to complete its task has rarely been measured. We propose frequency-prefix execution
(FPE), a training-free method that sends only the low-frequency part of each chunk, which carries its overall motion. The
robot rebuilds and executes the whole chunk, and the policy and its call schedule stay unchanged. Before quantization,
the error of the rebuilt chunk equals the energy of the frequencies left out, and under a stable control loop these
errors bound how far the robot drifts over an episode. Because the part of a chunk a robot needs varies with the state,
we also propose PairGap, a certificate computed before deployment from simulator runs at saved states, which bounds how
much task success the robot can lose at one state by executing the partial chunk in place of a full one. On the LIBERO
benchmark with the OpenVLA-OFT and $\pi_{0.5}$ policies, in episodes paired with the full chunk and under a
non-inferiority test, the two lowest frequencies keep task success on the Spatial and Object suites and four keep it on
the long-horizon LIBERO-10 suite. With OpenVLA-OFT, executing half of each chunk per call about doubles the policy calls and lowers
success on LIBERO-10, and under FAST+ the two-frequency coefficient stream is \OftFastTwoSaving\% smaller than the
full chunk's.
\end{abstract}

\section{Introduction}
\label{sec:intro}
Vision-language-action (VLA) policies control robots from camera images and language
instructions~\citep{Black2024Pi0,Kim2025OpenVLAOFT}. At each call the policy predicts an action chunk, a short sequence of
future controls~\citep{Zhao2023ACT}, and the robot executes a fixed number of those controls before it calls the policy
again~\citep{Zhao2026DynamicHorizon}. Large policies often run on a remote server~\citep{GeminiRobotics2025,Liu2026ComVLA},
so every chunk travels to the robot as a message over a link. On a narrow or shared link, the size of this message sets
how long the robot waits and how many robots the link serves. We propose frequency-prefix execution (FPE), which sends
only the low-frequency part of each chunk and keeps the policy and its call schedule unchanged.

Two existing ways shrink this message. Action tokenizers compress a chunk with a frequency transform or a learned code and
are judged by how closely they reconstruct it~\citep{Pertsch2025FAST,Liu2026OAT,Lian2026ActionPiece}, and a shorter
execution horizon sends fewer steps per call and makes the robot call the policy more
often~\citep{Zhao2026DynamicHorizon,Chen2026DynamicCommitment}. Neither tells how much of a chunk the robot needs. The
robot acts in closed loop: a small error in one executed chunk moves the state, the moved state changes the next camera
image and every later prediction, and the error can grow over an episode or be corrected at the next call. We therefore
ask how much of each predicted chunk a robot needs to complete its task when the policy, the number of steps executed per
call and the replanning schedule stay fixed.

\begin{figure}[t]
\centering
\includegraphics[width=\textwidth]{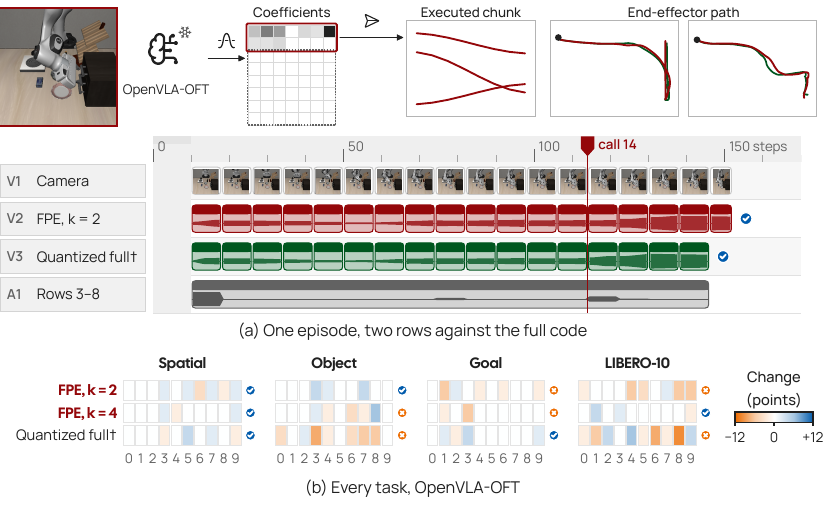}
\caption{Frequency-prefix execution on OpenVLA-OFT. (a) Episode 0 of Goal task 0 from a separate twenty-episode run,
one clip per call after the runner's ten settling steps (grey); waveforms show each executed step's translation norm, ticks a successful end, and the playhead the
call shown above (two rows sent, decoded translation axes, end-effector path from above and the side). The track Rows 3--8
gives the share of each full-code call's energy in those rows, zero in \TeaserZeroCalls{} of \TeaserCalls{} calls and at most
\TeaserMaxShare\%. (b) Success change per task over 50
paired episodes at seed 7, FPE against the quantized full code and that code against native execution; ticks and crosses
give the suite's two-point test.}
\label{fig:teaser}
\end{figure}

FPE answers with a code that needs no training. A chunk describes a smooth motion over a
short time, so a few slow oscillations describe it well. FPE applies the cosine transform along time, as FAST
does~\citep{Pertsch2025FAST}, and sends the lowest-frequency rows of coefficients; the robot sets the missing rows to
zero, rebuilds every step of the chunk and executes it (\cref{fig:teaser}). The number of rows sent sets the size of
every message, and the policy and its call schedule stay those of the full chunk. The cosine basis makes the dropped
detail accountable: before quantization, the squared error of the rebuilt chunk equals the energy of the rows left out, and when the control
loop under full chunks is stable, these one-chunk errors keep the robot within a bounded distance of the path it follows
with full chunks.

We evaluate FPE on the LIBERO benchmark~\citep{Liu2023LIBERO} with two released policies,
OpenVLA-OFT~\citep{Kim2025OpenVLAOFT} and $\pi_{0.5}$~\citep{Intelligence2025Pi05}, and pair every episode by task,
initial state and seed across the full chunk, the partial codes and a shorter execution horizon. Under a paired
non-inferiority test, two rows keep task success on the Spatial and Object suites and four rows keep it on the
long-horizon LIBERO-10 suite, for both policies. The shorter horizon, run with OpenVLA-OFT, about doubles the policy calls
and lowers success on LIBERO-10. At the size of two rows, keeping two of the chunk's steps and interpolating between them
fails the test where two rows pass it, and rounding every step to two bits loses success, so the low frequencies carry
what the task needs.

The number of rows a robot needs varies: LIBERO-10 needs more than the other suites, and within a suite the lost episodes
fall on a few tasks. To decide at one state, we propose \pairgap{}, a certificate computed before deployment. At a state
saved in a simulator, it runs the partial chunk the robot holds and the full chunks the policy could send, and bounds how
much task success the partial chunk can lose against the best of them, full chunks that have not arrived included.
Each simulation bounds a difference of failure rates, and chaining these bounds along the shortest paths of a graph gives
the exact worst-case regret that the simulations allow.

\section{Related work}
\label{sec:related}
Chunked policies predict several control steps per call~\citep{Zhao2023ACT,Black2024Pi0,Kim2025OpenVLAOFT} and decide how
many to execute before the next observation. Real-time chunking overlaps inference with execution by inpainting or
conditioning on the committed prefix~\citep{Black2025RTC,Black2025TTRTC}, asynchronous inference rolls the state forward
to the next chunk's start~\citep{Tang2025VLASH}, and other methods correct a chunk at every step, choose the horizon from
the state, from uncertainty or from the agreement of several horizons, truncate a chunk once its attention plateaus,
verify speculative chunks or stream
decoding~\citep{Sendai2025A2C2,Zhao2026DynamicHorizon,Chen2026DynamicCommitment,Liang2026AAC,Jing2025MoH,Xu2026AdaptiveChunking,Wang2026SVVLA,Li2026FlashVLA,Song2025PDVLA}.
These methods change when feedback reaches the controller, a shorter horizon raises the number of policy calls, and the
benefit of executing open loop has been traced to non-Markovian demonstrations~\citep{Zeng2026OpenLoop}. FPE keeps the horizon and the replanning schedule, changes only what is sent for each
chunk, and uses a shorter horizon under the same frozen policy as a control.

Action tokenizers turn a continuous chunk into a compact code that a policy generates and a robot decodes: FAST applies a
cosine transform, rounding and byte-pair encoding~\citep{Pertsch2025FAST}, and other codes use B-spline control points,
ordered latent tokens whose prefixes decode coarse motion, residual vector quantization or rank-preserving
losses~\citep{Zhou2025BEAST,Liu2026OAT,Jawaid2026NAC,Lian2026ActionPiece}. Spectral structure also shapes policies that
generate coarse motion first or learn frequency-aware
actions~\citep{Cao2026CSP,Su2025FreqPolicy,Guo2026FAFM,Dong2026TFGCA}. These codes are trained into the policy and judged
by reconstruction and by the success of the trained policy, and reconstruction accuracy explains only part of that
success~\citep{Chen2026ProAct}. FPE codes the finished chunk of a frozen policy, fixes the number of sent rows for an
episode, and measures closed-loop success with row removal and rounding separated.

Classical results explain why a cosine prefix carries most of a smooth chunk and how execution errors propagate. The
cosine basis approaches the Karhunen--Lo\`eve basis of a strongly correlated first-order Markov
signal~\citep{Ahmed1974DCT,Clarke1981KLT}, analyses of action chunking bound how execution errors compound under
incremental stability~\citep{Zhang2025Chunking}, and input-to-state stability bounds a state by a gain of the largest
input~\citep{Jiang2001ISS}. These results hold for a generic action error; the orthonormal cosine basis gives the error
of a truncated chunk in closed form, and we combine it with input-to-state stability and then measure task success
directly.

Control under communication constraints asks how few bits per sample keep a feedback loop
stable~\citep{Tatikonda2004CommunicationConstraints,Nair2007DataRate}, task-oriented communication selects transmissions
from a downstream task model~\citep{Shao2022TaskOriented,Gunduz2023BeyondBits,Liu2026ComVLA}, and edge--cloud
deployments split a policy across devices or keep a local head that tolerates network
delay~\citep{Zheng2026RoboECC,Peng2026CloudEdgeVLA}. Each derives its code or split from a model of the controlled
system, the channel or the task, which a pretrained VLA policy does not provide. FPE fixes a standard temporal code,
measures how many rows closed-loop success requires, and shrinks the message such a link carries.

Statistical procedures attach guarantees to decisions about robot policies from repeated rollouts: sequential tests
compare two policies with few trials, betting confidence sequences give sim-to-real certificates, anytime-valid paired
tests stop agent comparisons early, and simulators and learned surrogates reduce the real trials an evaluation
needs~\citep{Snyder2026BeyondBinary,Chen2026SimToRealBetting,Li2026AVAIVAT,Badithela2025SureSim,Luo2026X4Val}.
Best-arm identification selects an $\varepsilon$-optimal option with fixed confidence and designs where to
sample~\citep{EvenDar2006ActionElimination,Gabillon2012UGapE,Jourdan2023EBTC,Russo2020TTTS}. Their unit of decision is a
policy, an arm or a comparison of the two, with every option observable. \pairgap{} decides at one state against every
candidate at every evaluated depth, codes that have not arrived included, reports the received bits as a separate cost, and composes
the comparisons with shortest-path algorithms~\citep{Bellman1958Routing,Floyd1962Shortest} into the exact worst-case
regret the evidence allows, in the spirit of minimax-regret reasoning~\citep{Regan2009RegretElicitation}.

Test-time methods spend generation, selection, verification and evaluation on a policy's outputs: they decide how much
generation to spend~\citep{Salla2026POGP,Li2026ELASTIC,Li2026VLAATTC}, rank candidate chunks with learned verifiers or
values~\citep{Kwok2025RoboMonkey,Dai2025RoVer,Xu2026VGAS,Nakamoto2024VGPS,Jang2025VerifierFreeSelection,Yang2025AntiExploration,Kwok2026ScalingVerification,Guo2025VLAReasoner},
flag or roll back unreliable
chunks~\citep{Singh2026BOKBO,Yang2026FabriVLA,Roemer2026FlowVLAUncertainty,Yuan2026UPS,Chen2026ReconVLA,Sun2026PreVLA,Liu2026CheckVLA,Lou2026SafeLoop,Wan2026NoFreeChecker},
and allocate evaluation trials where they most shrink
uncertainty~\citep{Fei2025LIBEROPlus,Li2026LIBEROVPro,Wang2026SimRealCorrelation,XPolicyLab2026XPolicyLab,Liao2026ActiveFactorEval,Zouhar2026DynamicAllocation}.
FPE decides how much of a generated chunk to send, and \pairgap{} certifies that decision from rollouts paired by seed, as
common random numbers are~\citep{Glasserman1992CommonRandomNumbers,Kim2006SelectingBestSystem}.

\section{Frequency-prefix execution}
\label{sec:method}
\subsection{A partial code that keeps the horizon}
\label{sec:code}
At every call the policy observes the robot's state and predicts a chunk $X\in\mathbb R^{L\times m}$ of $L$ control steps
and $m$ action dimensions. The sender applies the orthonormal type-II cosine transform $C\in\mathbb R^{L\times L}$ along
time, $Z=CX$, so row $Z_r$, $r=1,\dots,L$, holds frequency $r-1$ of every action dimension, and its energy is the
squared norm $\|Z_r\|^2$. The sender rounds the coefficients at scale $s>0$, $\widetilde Z=s^{-1}\operatorname{round}(sZ)$,
and transmits the rows in increasing frequency. A receiver that holds the first $k$ rows sets the others to zero and
decodes
\begin{equation}
\widetilde X_k=C^{\top}\Pi_k\widetilde Z,\qquad \Pi_k=\operatorname{diag}(\underbrace{1,\dots,1}_{k},0,\dots,0),
\label{eq:prefix}
\end{equation}
a chunk of all $L$ steps, which the robot executes after clipping it to the action box and converting the gripper
channel. FPE fixes the depth $k$, the number of rows sent, for a whole episode, so the policy, the horizon, the stopping rule and the step limit stay those of
the full code. Since $C$ is orthogonal, the error of the decoded chunk splits into the energy of the dropped rows and the
rounding error of the kept rows,
\begin{equation}
\|X-\widetilde X_k\|_F^2=\sum_{r>k}\|Z_r\|^2+\sum_{r\le k}\|Z_r-\widetilde Z_r\|^2 ,
\label{eq:quantized-residual}
\end{equation}
where the second sum is at most $km/(4s^2)$ for nearest-integer rounding. Measured against the quantized full code
$\widetilde X_L$, the rounding of the kept rows cancels,
\begin{equation}
\|\widetilde X_L-\widetilde X_k\|_F^2=\sum_{r>k}\|\widetilde Z_r\|^2 .
\label{eq:quantized-tail}
\end{equation}
Both identities hold for the decoded chunk before clipping; clipping both chunks by the same Euclidean projection onto the
convex action box turns them into upper bounds, and the gripper conversion lies outside them. The experiments use $s=10$. The quantized full code against the
policy's own chunk measures rounding alone, and FPE against the quantized full code measures row removal alone. Time
truncation, the other lever, shortens the execution horizon: it executes the leading $d<L$ steps of each chunk and
then calls the policy again.

\subsection{Bits and calls}
\label{sec:accounting}
A message of $k$ rows carries a header of $\beta_h$ bits (shape, scale and candidate binding), one segment record of
$\beta_s$ bits per geometric band of rows, and for each row a $\beta_r$-bit index and $m$ coefficients of $\beta_c$ bits.
A prefix of $k=2^j\le L$ rows spans $1+\log_2k$ bands, so
\begin{equation}
\ell(k)=\beta_h+\beta_s\,(1+\log_2k)+(\beta_r+m\,\beta_c)\,k .
\label{eq:exact-price}
\end{equation}
\begin{restatable}[Serialized payload]{proposition}{proppayload}
\label{prop:payload}
With the implemented fields, $\beta_h=400$, $\beta_s=88$, $\beta_r=16$, and $\beta_c=8$ whenever every quantized
integer $q=\operatorname{round}(sZ_{rj})$ satisfies $-64\le q\le63$. The leading $d$ steps coded as a chunk of horizon $d$
also carry $\ell(d)$ bits.
\end{restatable}
At $m=7$ a message of one, two, four and eight rows carries 560, 720, 952 and 1,328 bits. Over $H$
executed steps, a controller that executes $d$ steps per call issues $N(d)=\lceil H/d\rceil$ calls and receives
$B(d)=N(d)\,\ell(d)$ bits.
\begin{restatable}[Calls and bits]{proposition}{propcounts}
\label{prop:counts}
If $d<L$ and $L$ both divide $H$, then $N(d)/N(L)=L/d$ and $B(d)/B(L)=(L/d)\,\ell(d)/\ell(L)$, and the bit ratio is at most
one exactly when $\ell(d)/d\le \ell(L)/L$.
\end{restatable}
FPE keeps the calls and scales the bits of every call by $\ell(k)/\ell(L)$, $0.54$ at $k=2$ and $0.72$ at $k=4$. Time
truncation at $d=4$ doubles the calls and multiplies the bits by $2\cdot952/1{,}328=1.43$. Episode lengths vary with
success, so the experiments report measured totals. Native execution, which sends the policy's own unquantized chunk
and runs no codec, is priced as a full-chunk message, the common reference. These are action-code bits, to which transport
headers add, and \cref{sec:spectra} sets them against the time of a call.

\subsection{Reconstruction error and closed-loop deviation}
\label{sec:theory}
Let $\Sigma=\mathbb E[XX^{\top}]\in\mathbb R^{L\times L}$ be the temporal second-moment matrix of the chunks pooled over
action dimensions, $\phi_r$ the $r$-th cosine basis vector, so that $Z_r=\phi_r^\top X$, and $e_r=\phi_r^{\top}\Sigma \phi_r$ the
expected energy of row $r$.
\begin{restatable}[Energy of a frequency prefix]{lemma}{lemenergy}
\label{lem:energy}
Before quantization, $\|X-C^\top \Pi_kZ\|_F^2=\sum_{r>k}\|Z_r\|^2$ for every chunk, and the expected squared error is
$\sum_{r>k}e_r$. Every rank-$k$ orthogonal projection $Q$ along time has $\mathbb E\|X-QX\|_F^2\ge\sum_{r>k}\lambda_r$,
where $\lambda_1\ge\dots\ge\lambda_L$ are the eigenvalues of $\Sigma$, with equality for the projection onto the leading
eigenvectors. If $e_1\ge\dots\ge e_L$, the first $k$ rows have the smallest expected error among all sets of $k$ cosine
rows.
\end{restatable}
The first part is Parseval's identity and the second the Ky Fan maximum principle. The gap between $\sum_{r>k}e_r$ and
$\sum_{r>k}\lambda_r$ is what a data-dependent basis could save, and \cref{sec:spectra} measures it.

Repeated truncation changes the states the robot visits. Let $x_t$ be the state at the $t$-th chunk boundary, where the
policy is queried, and $F(x,u)$ the state after chunk $u$ is clipped, converted and executed from $x$ under shared
exogenous randomness. The reference chunk $\bar u(x)$ is the policy's own chunk $X$ or the quantized full code
$\widetilde X_L$ predicted at $x$, and $u_k(x)=\widetilde X_k$ is decoded from $k$ rows of the same prediction. The
one-chunk residual at the visited state is $\zeta_t=\|u_k(x_t)-\bar u(x_t)\|_F$, which
\cref{eq:quantized-residual,eq:quantized-tail} evaluate.
\begin{restatable}[Deviation of every-chunk truncation]{proposition}{propiss}
\label{prop:iss}
Suppose that, for every realization of the shared randomness and in a norm $\|\cdot\|_M$ on the state, there is a set
$\mathcal S$ that contains the truncated trajectory $x_t$ and the reference trajectory $\bar x_t$ from a common initial
state, on which the reference loop $G(x)=F(x,\bar u(x))$ contracts, $\|G(x)-G(y)\|_M\le\rho\|x-y\|_M$ for
$x,y\in\mathcal S$ with $0\le\rho<1$, and on which $F$ is $\gamma$-Lipschitz in the chunk,
$\|F(x,u)-F(x,u')\|_M\le\gamma\|u-u'\|_F$ for $x\in\mathcal S$. Then
\begin{equation}
\|x_t-\bar x_t\|_M\le\gamma\sum_{t'<t}\rho^{\,t-1-t'}\zeta_{t'}\le\frac{\gamma\,\sup_{t'}\zeta_{t'}}{1-\rho}.
\label{eq:iss}
\end{equation}
\end{restatable}
This is the linear-gain case of discrete-time input-to-state stability~\citep{Jiang2001ISS}: the largest one-chunk
residual sets the radius of a tube around the reference trajectory, and \cref{lem:energy} ties that residual to the
dropped energy. For a goal-set criterion, success carries over to the truncated loop whenever the reference terminal
state lies deeper than that radius inside the goal set. Contact can break the contraction, so the
experiments measure what the residual does to task success directly.

\section{Frequency-prefix execution in closed loop}
\label{sec:experiments}

\subsection{Setup}
\label{sec:setup}
OpenVLA-OFT~\citep{Kim2025OpenVLAOFT,Kim2025OpenVLA} runs the official LIBERO protocol~\citep{Liu2023LIBERO} with its
released checkpoints and runner on the four suites LIBERO-Spatial, LIBERO-Object, LIBERO-Goal and LIBERO-10, written
Spatial, Object, Goal and LIBERO-10, with ten tasks per suite and fifty official initial states per task. Every execution runs 500
episodes per suite and seed, paired with the others by task, initial state and seed. The policy predicts eight-step
chunks, and the outcome of an episode is task success. At seed 7 five executions run on Spatial, Goal and LIBERO-10:
native execution of the policy's own unquantized chunk, the quantized full code, FPE at depths $k=2$ and $k=4$, and
time truncation at $d=4$; Object runs the first four with its released checkpoint. A second seed
sets both the evaluation seed and the environment seed to 8, which moves object placements within each initial state,
and repeats the quantized full code and both prefixes on Spatial, Goal and LIBERO-10; Goal also ran at seed 9. The
further seeds and the later OpenVLA-OFT studies ran in a rebuilt runtime, the same checkpoints and runner installed on
other GPUs, and each pairs its arms within that runtime. $\pi_{0.5}$, a flow-matching
policy~\citep{Intelligence2025Pi05}, runs the official LIBERO client of openpi~\citep{PhysicalIntelligence2025Openpi}, which executes five of its ten predicted steps
per call. The codec acts on those five steps at the same scale, so its quantized full code holds five rows and
\PiFiveFullBits{} bits, and native execution, the quantized full code and both prefixes ran on all four suites at seeds 7
and 8. A dagger marks a baseline or control that we construct. Every call of the code arms measured its message size;
native execution and time truncation are priced by \cref{eq:exact-price}.

An execution is \emph{non-inferior} on a suite when its true mean task difference against its reference exceeds $-2$
points. With $\Delta_j$ the paired success difference of task $j$ in points and $y_j=\Delta_j+2$, the one-sided exact
sign-flip test over sign vectors $\sigma$,
\begin{equation}
P=2^{-10}\sum_{\sigma\in\{-1,1\}^{10}}\mathbf 1\Big\{\sum_{j=1}^{10}\sigma_jy_j\ge\sum_{j=1}^{10}y_j\Big\},
\label{eq:sign-flip}
\end{equation}
rejects a mean difference of $-2$ points or less when $P\le0.025$, and the execution then passes the two-point test; over
several seeds the same test runs on the pooled task--seed units. A prefix \emph{keeps task success} on a suite when it
passes. FPE is tested against the quantized full code, which isolates row removal, and the quantized full code against
native execution, which isolates rounding. The test and its tolerance were fixed before the first frequency-arm
episode, and every later suite, seed and policy ran under the same test.

\subsection{Two rows keep Spatial and Object, four rows keep LIBERO-10}
\label{sec:episodes}
FPE keeps the success of the full code at a fraction of its bits (\cref{tab:episodes}). With OpenVLA-OFT at seed 7, both
depths pass on Spatial and two rows pass on Object. On Object the quantized full code itself falls below native execution
and fails its test, so rounding at scale 10, not row removal, costs that suite. On LIBERO-10 four rows pass, also after a
Bonferroni correction for the two depths, and two rows lose \MainLiberoTenTwoRowsDrop{} points and fail; the quantized
full code, the reference of the row comparisons, itself changes success by \MainLiberoTenQuantizedDelta{} points against
native execution there. On Goal both depths stay within one point of the quantized full code and neither passes at seed
7 alone.

The second seed repeats these decisions. Two rows pass on Spatial at seed 8 ($P=\SecondSeedSpatialTwoRowsP$), four rows
pass on LIBERO-10 ($P=\SecondSeedLiberoTenFourRowsP$), and two rows miss on LIBERO-10 again. On Goal two rows miss at seed
8 ($P=\GoalSeedEightTwoRowsP$) and pass at seed 9 ($P=\GoalSeedNineTwoRowsP$), and pooled over the thirty task--seed units
of seeds 7, 8 and 9 two rows ($P=\GoalThreeSeedTwoRowsP$) and four rows ($P=\GoalThreeSeedFourRowsP$) pass.

With $\pi_{0.5}$, pooled over seeds 7 and 8, two rows pass on Spatial, Object and Goal and four rows pass on all four
suites; on LIBERO-10 two rows miss, as with OpenVLA-OFT. The fifth row of the $\pi_{0.5}$ code holds at most
\PiFiveFifthRowShare\% of its energy, so four rows nearly equal the full code.

Time truncation pays about twice the calls and more bits than native execution, and on LIBERO-10, against native
execution rerun in the same runtime, its success changes by \OftLiberoTenTimeRerunDelta{} points
(\OftLiberoTenTimeRerunLosses{} paired losses, \OftLiberoTenTimeRerunWins{} wins).

\begin{table}[t]
\centering\small\setlength{\tabcolsep}{3.2pt}
\caption{Whole episodes, 500 per execution, suite and seed. Success, for $\pi_{0.5}$ the mean of two seeds;
Bit ratio, total bits relative to native execution; $^\ddagger$priced by \cref{eq:exact-price}. Time truncation on
LIBERO-10 ran in a rebuilt runtime, and against a native rerun in that runtime its success changes by
\OftLiberoTenTimeRerunDelta{} points. \pass{} and \fail{} mark the
two-point test against the quantized full code (FPE) or native execution (quantized full code), over the ten tasks of
seed 7 for OpenVLA-OFT and the twenty task--seed units of both seeds for $\pi_{0.5}$. Among executions that use part of
the chunk, bold and underline mark the best and second-best value of a column.}
\label{tab:episodes}
\begin{tabular}{@{}lr@{\hspace{8pt}}rr@{\hspace{8pt}}rr@{\hspace{8pt}}rr@{\hspace{8pt}}rr@{}}
\toprule
& & \multicolumn{2}{c}{Spatial} & \multicolumn{2}{c}{Object} & \multicolumn{2}{c}{Goal} & \multicolumn{2}{c}{LIBERO-10} \\
\cmidrule(lr){3-4}\cmidrule(lr){5-6}\cmidrule(lr){7-8}\cmidrule(l){9-10}
Execution & Bits & Success & Bit ratio & Success & Bit ratio & Success & Bit ratio & Success & Bit ratio \\
& per call & (\%) $\uparrow$ & $\downarrow$ & (\%) $\uparrow$ & $\downarrow$ & (\%) $\uparrow$ & $\downarrow$ & (\%) $\uparrow$ & $\downarrow$ \\
\midrule
\multicolumn{10}{@{}l}{\emph{OpenVLA-OFT, seed 7, full chunk at every call}} \\
Native full & 1,328\rlap{$^{\ddagger}$} & 98.4 & 1.00 & 97.6 & 1.00 & 97.0 & 1.00 & 93.8 & 1.00 \\
Quantized full$^{\dagger}$ & 1,328 & 98.6\rlap{\,\pass} & 1.00 & 94.8\rlap{\,\fail} & 1.04 & 97.6\rlap{\,\pass} & 1.00 & 92.2\rlap{\,\fail} & 1.02 \\
\midrule
\multicolumn{10}{@{}l}{\emph{OpenVLA-OFT, seed 7, part of the chunk}} \\
FPE, $k=2$ & 720 & \textbf{98.8}\rlap{\,\pass} & \textbf{0.54} & \textbf{95.8}\rlap{\,\pass} & \textbf{0.55} & \secondbest{96.6}\rlap{\,\fail} & \textbf{0.55} & \secondbest{90.0}\rlap{\,\fail} & \textbf{0.56} \\
FPE, $k=4$ & 952 & \textbf{98.8}\rlap{\,\pass} & \secondbest{0.72} & \secondbest{94.8}\rlap{\,\fail} & \secondbest{0.74} & \textbf{96.8}\rlap{\,\fail} & \secondbest{0.72} & \textbf{92.6}\rlap{\,\pass} & \secondbest{0.73} \\
Time truncation$^{\dagger}$ & 952\rlap{$^{\ddagger}$} & 98.2 & 1.42 & -- & -- & 95.6 & 1.47 & 87.4 & 1.57 \\
\midrule
\multicolumn{10}{@{}l}{\emph{$\pi_{0.5}$, seeds 7 and 8, full chunk at every call}} \\
Native full & 1,112\rlap{$^{\ddagger}$} & 98.4 & 1.00 & 99.4 & 1.00 & 96.5 & 1.00 & 93.1 & 1.00 \\
Quantized full$^{\dagger}$ & 1,112 & 98.8\rlap{\,\pass} & 1.00 & 98.5\rlap{\,\fail} & 1.00 & 96.4\rlap{\,\pass} & 1.01 & 92.9\rlap{\,\fail} & 1.02 \\
\midrule
\multicolumn{10}{@{}l}{\emph{$\pi_{0.5}$, seeds 7 and 8, part of the chunk}} \\
FPE, $k=2$ & 720 & \textbf{98.6}\rlap{\,\pass} & \textbf{0.65} & 98.5\rlap{\,\pass} & \textbf{0.65} & \secondbest{96.1}\rlap{\,\pass} & \textbf{0.66} & 92.8\rlap{\,\fail} & \textbf{0.66} \\
FPE, $k=4$ & 952 & \secondbest{98.5}\rlap{\,\pass} & \secondbest{0.86} & 98.5\rlap{\,\pass} & \secondbest{0.86} & \textbf{96.2}\rlap{\,\pass} & \secondbest{0.87} & 92.8\rlap{\,\pass} & \secondbest{0.87} \\
\bottomrule
\end{tabular}

\end{table}

\subsection{The low rows carry what the task needs}
\label{sec:matched}
\begin{figure}[t]
\centering
\includegraphics[width=\textwidth]{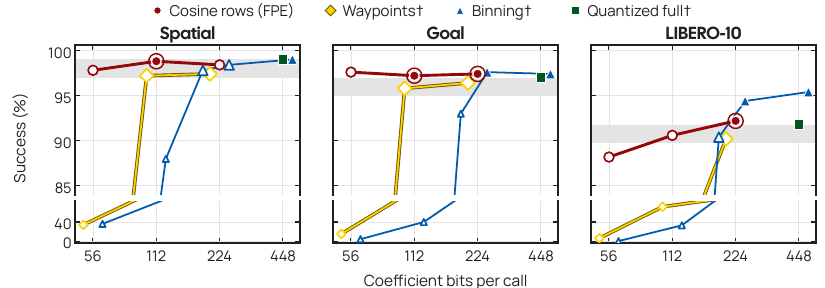}
\caption{Task success of OpenVLA-OFT against the coefficient bits of one call for three codes of a finished chunk. Filled
marks pass the two-point test against the quantized full code, whose tolerance the grey band shows; open marks miss it;
small marks are pilots on ten initial states per task; rings mark FPE. Equal-size codes are offset horizontally; the
lower strip holds the range below 84\%.}
\label{fig:rate-success}
\end{figure}

Does any code of the prefix's size keep success, or do the low rows matter? \Cref{fig:rate-success} compares FPE with two
codes of the finished chunk at the same coefficient bits per call: waypoints, which keep evenly spaced steps and
interpolate linearly between them, and binning, which rounds every step of every action dimension to a few bits between
the 1st and 99th percentiles of the policy's action statistics. A row or a waypoint costs 56 bits, and binning costs 56
bits for each bit per value.
The comparison ran OpenVLA-OFT in the rebuilt runtime, Spatial and Goal at seed 7 and LIBERO-10 at seed 8, paired with the
quantized full code on the same states, and its new arms were fixed before they ran. The cosine rows first pass the test at
\RateSuccessSpatialCosineSmallestPass{} bits on Spatial and Goal and at \RateSuccessLiberoTenCosineSmallestPass{} bits on
LIBERO-10. Waypoints pass at no size below the full code's 448 bits, and binning first passes at
\RateSuccessSpatialBinningSmallestPass{} bits on Spatial and at \RateSuccessGoalBinningSmallestPass{} bits on Goal and
LIBERO-10. One row, the constant row at the chunk's mean, misses on every suite. On Spatial four rows miss on this runtime
($P=\RateSuccessSpatialCosineFourP$) and pass on the main runs at seeds 7 and 8.

On LIBERO-10, 4-bit binning passes at the raw size of four rows, and under the best general-purpose lossless coder its
bin indices cost \CoderTenFourBinningBest{} bits per call against \CoderTenFourRowsBest{} for four rows. 3-bit binning misses the test at \CoderTenThreeBinningBest{} coded bits and 2-bit binning loses success on its
pilot, so four cosine rows are the smallest coded message tested that keeps LIBERO-10 success.

\subsection{The losses sit in a few tasks}
\label{sec:losses}
Paired by initial state, most episodes end the same way under FPE and the quantized full code, and the few that differ
gather in a few tasks (\cref{fig:teaser}b). On Spatial two rows win \MainSpatialTwoRowsWins{} and lose
\MainSpatialTwoRowsLosses{} of 500 pairs, and the LIBERO-10 loss of two rows comes from \MainLiberoTenTwoRowsLosses{}
losses against \MainLiberoTenTwoRowsWins{} wins, with wins and losses side by side in the same tasks.

If the dropped energy predicted the losses, tasks with a larger share would lose more. For a task, let
$E_r$ be the energy $\|\widetilde Z_r\|^2$ of row $r$ over the six continuous axes, summed over every call of the task's
quantized-full episodes; the dropped share at depth $k$ is $\sum_{r>k}E_r/\sum_rE_r$, which by \cref{eq:quantized-tail}
is the relative squared residual of FPE against the quantized full code on those trajectories. For OpenVLA-OFT these
spectra were recorded on Spatial and Goal. Two rows keep
\EnergyKeptTwoLow--\EnergyKeptTwoHigh\% of the energy of every Spatial and Goal task, and over these \EnergySuccessTasks{} tasks the dropped share
shows no detectable rank association with the success change (Spearman $\rho_S=\EnergySuccessRhoTwo$ at two rows and
$\EnergySuccessRhoFour$ at four, below the $|\rho_S|=0.3$ fixed before the computation). Over the \PiFiveEnergyTasks{}
tasks of $\pi_{0.5}$ at seed 7, the same association is $\rho_S=\PiFiveEnergySpearman$ at two rows (permutation
$P=\PiFiveEnergyP$).

\subsection{One row-truncated chunk rarely changes the outcome}
\label{sec:restored}
Does one row-truncated chunk change the outcome, or does the loss build up over an episode? Restored states isolate one
chunk: a simulator restores a state, the first chunk is executed from a prefix of its rows, and the policy then runs its
own chunks until the episode ends. LaRA-VLA~\citep{Bai2026LaRA} and Flex-$\pi$~\citep{Yan2026FlexPi} supply these tests
through banks, frozen sets of candidate chunks at restored states with their logged continuation outcomes. Two ladders
test one, two, four and eight rows of LaRA-VLA and one and ten rows of Flex-$\pi$ at the same sixteen states, with four
candidates and four continuation seeds per candidate and depth; the LIBERO-10 panel runs Flex-$\pi$ at one denoising step
on forty states with four candidates and 64 matched seeds, and its eight-row reference rebuilds each candidate's full
quantized chunk.

\begin{figure}[t]
\centering
\includegraphics[width=\textwidth]{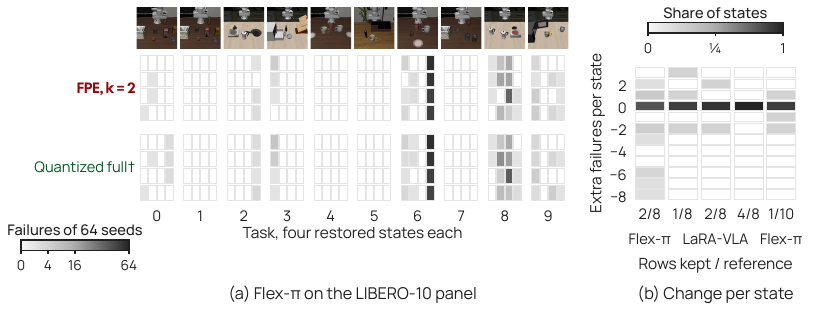}
\caption{One row-truncated chunk at restored states, after which the policy runs its own chunks. (a) Failures among
the 64 matched seeds of each candidate chunk (rows) at each of the 40 LIBERO-10 states of Flex-$\pi$ (columns, four per
task), at two rows and at the eight-row reference. (b) Share of states whose failure count changes by each amount,
truncated minus reference, on these states (256 continuations each) and on the sixteen-state ladders (16 each).}
\label{fig:restored}
\end{figure}

On the ladders two LaRA-VLA rows succeed on \LadderLaraTwoSuccesses{} of \LadderLaraTrajectories{} continuations, as the full code does, and one Flex-$\pi$ row on \LadderFlexPiPrefixSuccesses{} against \LadderFlexPiFullSuccesses{} for the ten-row code. On the forty LIBERO-10
states two rows fail \PanelTwoRowFailures{} of \PanelTwoRowRollouts{} paired continuations against \PanelEightRowFailures{}
for the eight-row reference, no state gains more than two failures, and the failures concentrate where both depths fail on
most seeds (\cref{fig:restored}). A single two-row chunk carries little risk at these states; executed at every call of
OpenVLA-OFT, two rows lose LIBERO-10 success.

\subsection{Two rows hold nearly all the energy and pay a coding overhead}
\label{sec:spectra}
A frozen policy's chunks are smooth enough that two cosine rows carry almost all of their energy. Over 64 logged
LaRA-VLA and 64 Flex-$\pi$ chunks, two rows hold \LaraDctShareTwo\% and \FlexPiDctShareTwo\% of the energy of the six
continuous axes, against \LaraKltShareTwo\% and \FlexPiKltShareTwo\% for the two leading Karhunen--Lo\`eve transform (KLT)
directions pooled over each bank, so the fixed basis is close to the data-dependent optimum of \cref{lem:energy}. Much of
this concentration is the mean of the chunk. On Flex-$\pi$, two rows raise the mean relative error
$\|X-\widetilde X_k\|_F/\|X\|_F$ from \FlexPiReconstructionErrorAllTen\% to \FlexPiReconstructionErrorTwo\%, an empirical
scale of the residual $\zeta_t$ of \cref{prop:iss} on this bank.

Compression shrinks the saving. Counted on the coefficients alone, without the header, segment records and row
indices, two fixed-length rows are \FixedTwoRowSavingEight\% smaller than the eight-step full code and
\FixedTwoRowSavingTen\% smaller than the ten-step one. Priced in tokens of the released FAST+ vocabulary, which already
makes the dropped high rows cheap, two rows save \OftFastTwoSaving\% of the OpenVLA-OFT coefficient stream over
\OftFastChunks{} distinct chunks of Spatial, Goal and LIBERO-10 quantized-full episodes, \LaraFastTwoRowSaving\% on
LaRA-VLA and \FlexPiFastTwoRowSaving\% on Flex-$\pi$. The best general-purpose lossless coder, run as one stream per
episode, keeps a \CoderTwoSaving\% saving for two OpenVLA-OFT coefficient rows. The saving matters on narrow or shared links. On shared
GPUs a call spends a median of \CallTimePolicyMedianMs\,ms in the policy and \CallTimeCodecMedianMs\,ms in the codec, so
below about \LinkBreakEvenKbps\,kbit/s the fixed-length full code spends longer on the link than in the policy, and above
about \CodecBreakEvenMbps\,Mbit/s the codec costs more time than two rows save.

\section{PairGap, a certificate for one state}
\label{sec:certificate}
A fixed row count serves a whole episode, and its few losses sit in a few tasks. A controller that decides at one state
needs evidence about that state. \pairgap{} asks whether the prefix that has arrived can be executed at a restored state
with a guaranteed regret, measured against every option the controller could have waited for.

\subsection{Regret as a shortest-path distance}
\label{sec:pairgap-method}
At a restored state a fixed policy proposes $K$ candidate chunks, each coded by $L$ cosine rows. An \emph{option}
$a=(i,k)$ is the $k$-row prefix of candidate $i$. Executed from the restored state under a continuation seed, it fails
($Y_a=1$) or succeeds ($Y_a=0$), and $\mu_a=\mathbb E[Y_a]$ is its failure rate over seeds. The comparator set
$\mathcal A$ contains every candidate at every evaluated depth, full codes in flight included, and the received bits are
reported as a separate cost. Executing $a$ incurs the regret $\operatorname{reg}(a)=\mu_a-\min_{b\in\mathcal A}\mu_b$. An
$\varepsilon$ certificate for $a$ is a bound $\operatorname{reg}(a)\le\varepsilon$ that holds for every vector of failure rates
consistent with the evidence, so it covers the codes in flight.

\begin{figure}[t]
\centering
\includegraphics[width=\textwidth]{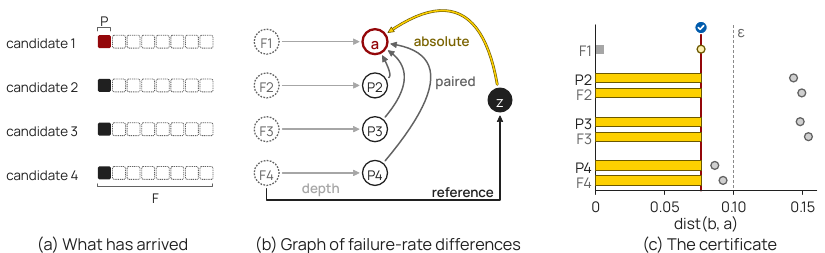}
\caption{\pairgap{} at LaRA-VLA state 1 with 64 rollouts per edge. (a) The one-row prefixes $P$ have arrived and the full codes
$F$ are in flight; $a$ is candidate 1's prefix. (b) Edges bound differences of failure rates; one reference edge is drawn.
(c) $\dist(b,a)$ for every source $b$ in the hue of its last edge, dots for the other route; the certificate is the
largest, $R_w(a)=\CertificateFigureRegret\le\varepsilon=0.10$.}
\label{fig:pairgap}
\end{figure}

Each piece of evidence bounds a difference of failure rates. A directed edge $b\to a$ with weight $w(b,a)$ records
$\mu_a-\mu_b\le w(b,a)$ on the vertex set $\mathcal A\cup\{z\}$, where the reference vertex has $\mu_z=0$. Write
$\overline p(c,n,\delta)$ and $\underline p(c,n,\delta)$ for the upper and lower Clopper--Pearson
limits~\citep{Clopper1934Binomial} of a failure rate after $c$ failures in $n$ rollouts at error $\delta$. Four kinds of
edge enter the graph (\cref{fig:pairgap}):
\emph{reference edges} $b\to z$ of weight $0$ encode $\mu_b\ge0$ and need no data;
the \emph{absolute edge} $z\to a$ carries $U_a=\overline p(c_a,n,\delta_a)$ from $n$ rollouts of $a$ with $c_a$ failures,
at the error $\delta_a$ allocated to it;
a \emph{signed paired edge} $b\to a$ between two received options runs both on the same $n$ seeds and, with $n_+$ seeds on
which only $a$ fails and $n_-$ on which only $b$ fails, carries
$w(b,a)=\overline p(n_+,n,\delta_p/2)-\underline p(n_-,n,\delta_p/2)$ at its error $\delta_p$, which keeps the
within-seed dependence of the pair;
a \emph{depth edge} $(i,L)\to(i,k)$ from an unreceived full code into its received prefix carries a weight $w_d$
calibrated on other states, in our study the upper Clopper--Pearson limit of the pooled count of seeds, over the other
states of the bank, on which a prefix fails and its full code succeeds. Every other ordered pair carries the trivial weight $\bar w=1$. Summing edge inequalities along a path bounds
$\mu_a-\mu_b$ by the length of that path, and the shortest such length $\dist(b,a)$ is the tightest of these bounds, so the
worst-case regret is a shortest-path distance,
\begin{equation}
R_w(a)=\max_{b\in\mathcal A}\dist(b,a),\qquad a\text{ carries an }\varepsilon\text{ certificate}\iff R_w(a)\le\varepsilon .
\label{eq:closure}
\end{equation}
\begin{restatable}[Exact certificate]{theorem}{thmclosure}
\label{thm:closure}
Let $\mathcal A$ be finite, let the graph on $\mathcal A\cup\{z\}$ have a path between every ordered pair of vertices,
and let the set $\mathcal F(w)$ of vectors $\mu$ with $\mu_z=0$ that satisfy every edge be nonempty. Then for every
$a\in\mathcal A$ and $b\in\mathcal A\cup\{z\}$, $\sup_{\mu\in\mathcal F(w)}(\mu_a-\mu_b)=\dist(b,a)$ and
$\sup_{\mu\in\mathcal F(w)}\big(\mu_a-\min_{b\in\mathcal A}\mu_b\big)=R_w(a)$, and the potential
$\mu_v=\dist(b^\star,v)-\dist(b^\star,z)$ with $b^\star\in\arg\max_{b\in\mathcal A}\dist(b,a)$ attains the regret supremum.
\end{restatable}
The maximizing source and its shortest path form the \emph{witness path}, the evidence that limits the certificate.
With one prefix depth and the full depth per candidate, $|\mathcal A|=2K$, and Floyd--Warshall computes all shortest
paths on the $2K+1$ vertices, $729$ relaxations for $K=4$.

\begin{restatable}[Validity]{proposition}{propvalidity}
\label{prop:validity}
If each sampled edge holds with probability at least $1-\delta_e$ and the depth edges of a state drawn exchangeably with
the calibration states hold with probability at least $1-\delta_q$, then $\Pr(\text{\pairgap{} certifies }a\text{ and
}\operatorname{reg}(a)>\varepsilon)\le\delta_q+\sum_e\delta_e$.
\end{restatable}

Which evidence can shorten the bound? The absolute edge alone gives $R_w(a)=U_a$. Relational edges help exactly at the
sources whose reference path $b\to z\to a$ is longer than $\varepsilon$, and each such source then needs a relational path
within $\varepsilon$; an unreceived full code lies at most the depth-edge weight farther from $a$ than its received prefix.
\begin{restatable}[Composition]{proposition}{propcomposition}
\label{prop:composition}
Let every option have a zero-weight edge into $z$, every option-to-option edge weigh at most $\bar w$, and every edge
out of $z$ weigh $\bar w$ except $z\to a$ with weight $U_a\in[0,\bar w]$. Write $\dist_{\neg z}$ for distances over
paths that avoid $z$. (i) $R_w(a)=\max_{b\in\mathcal A}\min\{\dist(b,z)+U_a,\,\dist_{\neg z}(b,a)\}$, and without paired
and depth edges $R_w(a)=U_a$. (ii) If $U_a>\varepsilon$, adding paired and depth edges certifies $a$ exactly when
$\min\{\dist(b,z)+U_a,\dist_{\neg z}(b,a)\}\le\varepsilon$ for every $b\in\mathcal A$. (iii) For an unreceived full code,
$\dist_{\neg z}((i,L),a)\le w_d+\dist_{\neg z}((i,k),a)$. (iv) If $a$ has no failure and a received rival never disagrees
with $a$ on $n$ shared seeds, the paired edge weighs $\overline p(0,n,\delta_p/2)$, which is at least the absolute
weight $\overline p(0,n,\delta_a)$ whenever $\delta_p\le2\delta_a$.
\end{restatable}
Part (iv) predicts what happens on a nearly failure-free panel: a concordant rival adds rollouts and no shorter path.

\subsection{Certifying eight LaRA-VLA states}
\label{sec:certificate-results}
The certificate study replays logged rollouts at eight restored LaRA-VLA states with four candidates each. A candidate's
options are its one-row prefix, \DepthCertificatePrefixBits{} bits, and its full code of eight rows, \DepthCertificateFullBits{} bits, and the executed option is
candidate 1's prefix. Even seeds feed the certificate and odd seeds a held-out audit. The total error
$\delta=\delta_q+\delta_o=0.05$ is split equally between the depth edges, $\delta_q=0.025$, and the sampled edges, whose
online error $\delta_o=0.025$ an arm divides equally among the edges it samples; \pairgap{} samples one absolute and $K-1$
paired edges, so $\delta_a=\delta_p=\delta_o/K$. Four constructed arms isolate the evidence. Absolute$^\dagger$ certifies
through the executed option's absolute edge alone, at error $\delta_o$. Paired and depth$^\dagger$ uses the paired and
depth edges without the absolute edge, each of its $K-1$ paired edges at error $\delta_o/(K-1)$. Receive full$^\dagger$
waits for every full code and certifies candidate 1's full code through its absolute edge, at error $\delta_o$.
One-candidate receipt$^\dagger$ receives every candidate's prefix, fixes candidate 1 before any rollout, bounds its prefix through an absolute edge at error
$\delta_o/2$, and when that bound exceeds $0.05$ requests the candidate's remaining rows, bounds the full code at the same
error and executes the depth with the smaller bound; with no observed failure the bound $1-(\delta_o/2)^{1/n}$ reaches
$0.05$ at 86 rollouts, beyond the largest budget of 64, so all \CertificateReceiptRequests{} states requested the
full code.

\begin{figure}[t]
\centering
\includegraphics[width=\textwidth]{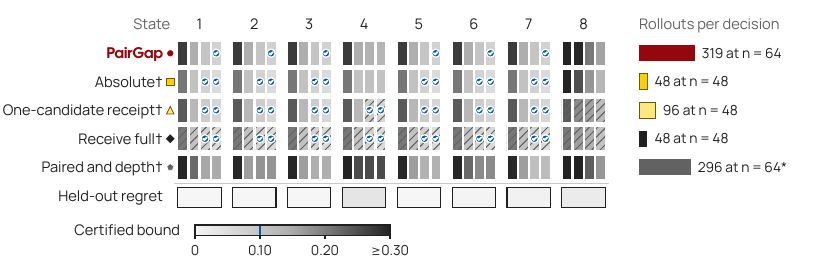}
\caption{The certificate on eight LaRA-VLA states, replayed on logged rollouts at tolerance $\varepsilon=0.10$. Each state
has four cells for budgets of $n=16$, 32, 48 and 64 rollouts per sampled edge, at most $n$ shared seeds on a paired edge; ticks mark bounds within $\varepsilon$, hatches an
executed full code, and the bottom row is the largest regret of any executed option on held-out seeds. At the right,
rollouts per decision at each arm's first certifying $n$ ($^*$no state certified).}
\label{fig:certificate}
\end{figure}

In the replay, comparisons between options decide no certificate (\cref{fig:certificate}). The candidates succeed on
almost every shared seed, so by \cref{prop:composition}(iv) a concordant pair is no shorter than the absolute bound.
\pairgap{} certifies the same six states as Absolute$^\dagger$, one budget later, at \CertificateUnifiedFirstBudget{} rollouts per edge against
\CertificateAbsoluteFirstBudget, and at its first certifying budget spends \CertificateUnifiedRollouts{} rollouts per
decision against \CertificateAbsoluteRollouts. Each of these six certificates is attained on the reference path through
$z\to a$, so it uses the absolute edge alone and none of the depth calibration, and the delay comes from the error of
that edge, $\delta_o/4$ for \pairgap{} against $\delta_o$ for Absolute$^\dagger$, which raises the zero-failure bound at
$n=48$ from \CertificateAbsoluteBoundFortyEight{} to \CertificateUnifiedBoundFortyEight, just above $\varepsilon$.

Paired and depth$^\dagger$ certifies no state. The replay pairs only the seeds that both options logged, so most
paired edges rest on fewer than $n$ shared seeds, which widens their bounds. Rerun with every rival on the executed
option's own seeds at states \SharedSeedCompletedStates, the arm certifies state \SharedSeedCertifiedStates{} through the
depth edge and its pooled calibration and leaves state \SharedSeedOpenStates{} open, where the executed option fails on
\SharedSeedOpenFailures{} of those seeds. One-candidate receipt$^\dagger$ certifies seven
states, six at prefix depth and state 4 through its full code, and Receive full$^\dagger$ certifies the same seven at full
depth and pays for three codes its certificate never uses. State 8 stays open under every arm. On the held-out seeds no
certified option exceeds its tolerance, and the largest held-out regret of any executed option is \CertificateHeldOutMax. For comparisons
to decide, a panel needs what \cref{prop:composition}(ii) asks: rivals that fail alone on enough seeds.

\subsection{Forty LIBERO-10 states and the bank of one chunk}
\label{sec:certificate-banks}
The same rules on the forty Flex-$\pi$ LIBERO-10 states of \cref{sec:restored}, whose options all ran the same 64
continuation seeds, half certifying and half auditing, repeat this picture: with four candidates, comparisons certify no
state that the absolute edge leaves open, at any tolerance. With 32 certification seeds a failure-free
option is bounded at \FlexPiFailureFreeBound{} at best, so no rule certifies at $\varepsilon=0.10$. At $\varepsilon=0.15$
Absolute$^\dagger$ and \pairgap{} each certify \FlexPiUnifiedFifteen{} states, One-candidate receipt$^\dagger$ certifies
\FlexPiReceiptFifteen{} with \FlexPiReceiptBits{} bits per decision against \FlexPiReceiveFullFifteen{} for Receive
full$^\dagger$ at \FlexPiReceiveFullBits, and Paired and depth$^\dagger$ certifies none at $\varepsilon=0.15$ and
\FlexPiPairedDepthTwenty{} at $\varepsilon=0.20$. At the one state where every option fails on most seeds, task \FlexPiHardTask{} at initial state
\FlexPiHardInit, the relational edges lower the bound from \FlexPiHardAbsolute{} to \FlexPiHardUnified, and the executed
option fails alone on enough seeds that no tolerance up to 0.20 certifies it. A panel built to separate the
edges, a Gaussian policy around the OpenVLA-OFT Goal head at ten restored states whose options meet the same continuation
noise, certifies no state under any rule with four candidates and 64 certification seeds, because the sampled rivals disagree with the executed
prefix on many seeds and keep the paired edges wide.

A policy that proposes one chunk per state changes the comparison. Its bank holds the chunk's prefix and the full code it
was cut from, and the depth edge, calibrated on the other states, bounds their gap, so the certificate bounds the regret
against the chunk's own full code. With this edge the graph certifies all eight LaRA-VLA states at
$\varepsilon=0.05$ and all forty Flex-$\pi$ and ten Gaussian OpenVLA-OFT states at $\varepsilon=0.10$, where the executed
option's absolute edge certifies none.

\section{Conclusion}
\label{sec:conclusion}
At an unchanged execution horizon, a robot can act on a low-frequency prefix of each chunk. On LIBERO, every message
shrinks, the call schedule stays the same, and two rows keep the success of the full code on Spatial and Object and four
rows on LIBERO-10, for OpenVLA-OFT and $\pi_{0.5}$. At the same raw size the cosine rows pass where waypoints fail, and
binning passes at the cosine rows' size or larger, at nearly twice their coded bits on LIBERO-10. Shortening the horizon is the costlier lever, since it multiplies the calls, and on LIBERO-10 it also lowers success.
The lost episodes sit in a few tasks, and the events, contact among them, that turn a residual of about one percent of
the energy into a failure are the next target. \pairgap{}
turns rollouts at a restorable state into the exact worst-case regret that the evidence allows over received and
unreceived options. For a policy that proposes one chunk per state, the graph with its depth edge certifies every tested state at
$\varepsilon=0.05$ on LaRA-VLA and at $\varepsilon=0.10$ on the other two banks, where the executed option's own rollouts
certify none.

\section*{Statements}

\subsection*{Ethics}
The experiments run released robot policies in the LIBERO simulator, with no human subjects and no personal data.

\subsection*{Reproducibility}
\Cref{sec:method,sec:certificate} define the code, its accounting and the certificate, and \cref{sec:setup} the episode
protocol and its test. The policies, checkpoints, simulator, task suites and the openpi client are public; the appendix
lists the source revisions of the whole-episode runs, the certificate protocol with its seeds, and the protocol and seeds
of the logged-population study. The code and the machine-readable records behind every number, table and figure are
released at \url{https://github.com/ZenAlexa/act-on-the-low-frequencies}.

\subsection*{Use of generative AI}
Generative AI tools were used in the research and in preparing this paper. The authors verified all content and take full
responsibility for it.

\bibliography{references}

\begin{thebibliography}{93}
\providecommand{\natexlab}[1]{#1}
\providecommand{\url}[1]{\texttt{#1}}
\expandafter\ifx\csname urlstyle\endcsname\relax
  \providecommand{\doi}[1]{doi: #1}\else
  \providecommand{\doi}{doi: \begingroup \urlstyle{rm}\Url}\fi

\bibitem[Ahmed et~al.(1974)Ahmed, Natarajan, and Rao]{Ahmed1974DCT}
N.~Ahmed, T.~Natarajan, and K.~R. Rao.
\newblock {Discrete Cosine Transform}.
\newblock \emph{IEEE Transactions on Computers}, C-23\penalty0 (1):\penalty0
  90--93, 1974.

\bibitem[Badithela et~al.(2025)Badithela, Snyder, Zha, Mikhail, O'Kelly, Dixit,
  and Majumdar]{Badithela2025SureSim}
Apurva Badithela, David Snyder, Lihan Zha, Joseph Mikhail, Matthew O'Kelly,
  Anushri Dixit, and Anirudha Majumdar.
\newblock {Reliable and Scalable Robot Policy Evaluation with Imperfect
  Simulators}.
\newblock \emph{arXiv preprint arXiv:2510.04354}, 2025.

\bibitem[Bai et~al.(2026)Bai, Lyu, Zhou, Li, Wang, Xing, Zhao, Wang, Wang, Chi,
  Chen, and Zhang]{Bai2026LaRA}
Shuanghao Bai, Jing Lyu, Wanqi Zhou, Zhe Li, Dakai Wang, Lei Xing, Xiaoguang
  Zhao, Pengwei Wang, Zhongyuan Wang, Cheng Chi, Badong Chen, and Shanghang
  Zhang.
\newblock {Latent Reasoning VLA: Latent Thinking and Prediction for
  Vision-Language-Action Models}.
\newblock In \emph{Proceedings of the 43rd International Conference on Machine
  Learning}, pp.\  5409--5426, 2026.

\bibitem[Bellman(1958)]{Bellman1958Routing}
Richard Bellman.
\newblock {On a Routing Problem}.
\newblock \emph{Quarterly of Applied Mathematics}, 16\penalty0 (1):\penalty0
  87--90, 1958.

\bibitem[Black et~al.(2025{\natexlab{a}})Black, Brown, Darpinian, Dhabalia,
  Driess, Esmail, Equi, Finn, Fusai, Galliker, et~al.]{Intelligence2025Pi05}
Kevin Black, Noah Brown, James Darpinian, Karan Dhabalia, Danny Driess, Adnan
  Esmail, Michael Equi, Chelsea Finn, Niccolo Fusai, Manuel~Y. Galliker, et~al.
\newblock {$\pi_{0.5}$: A Vision-Language-Action Model with Open-World
  Generalization}.
\newblock In \emph{Proceedings of the 9th Conference on Robot Learning}, pp.\
  17--40, 2025{\natexlab{a}}.

\bibitem[Black et~al.(2025{\natexlab{b}})Black, Brown, Driess, Esmail, Equi,
  Finn, Fusai, Groom, Hausman, Ichter, et~al.]{Black2024Pi0}
Kevin Black, Noah Brown, Danny Driess, Adnan Esmail, Michael Equi, Chelsea
  Finn, Niccolo Fusai, Lachy Groom, Karol Hausman, Brian Ichter, et~al.
\newblock {$\pi_0$: A Vision-Language-Action Flow Model for General Robot
  Control}.
\newblock In \emph{Robotics: Science and Systems XXI}, 2025{\natexlab{b}}.

\bibitem[Black et~al.(2025{\natexlab{c}})Black, Galliker, and
  Levine]{Black2025RTC}
Kevin Black, Manuel~Y. Galliker, and Sergey Levine.
\newblock {Real-Time Execution of Action Chunking Flow Policies}.
\newblock In \emph{Advances in Neural Information Processing Systems},
  volume~38, pp.\  37596--37620, 2025{\natexlab{c}}.

\bibitem[Black et~al.(2025{\natexlab{d}})Black, Ren, Equi, and
  Levine]{Black2025TTRTC}
Kevin Black, Allen~Z. Ren, Michael Equi, and Sergey Levine.
\newblock {Training-Time Action Conditioning for Efficient Real-Time Chunking}.
\newblock \emph{arXiv preprint arXiv:2512.05964}, 2025{\natexlab{d}}.

\bibitem[Cao et~al.(2026)Cao, Wang, Byrnes, Chen, Du, and Garg]{Cao2026CSP}
Shuxin Cao, Liquan Wang, Walker Byrnes, Yiye Chen, Yilun Du, and Animesh Garg.
\newblock {Hierarchical Policy Learning via Spectral Decomposition}.
\newblock \emph{arXiv preprint arXiv:2606.29570}, 2026.

\bibitem[Chen et~al.(2026{\natexlab{a}})Chen, Wang, Chen, Li, He, Zhang, and
  Wu]{Chen2026DynamicCommitment}
Feng Chen, Xianghui Wang, Yuxuan Chen, Boying Li, Yefei He, Zeyu Zhang, and
  Yicheng Wu.
\newblock {Dynamic Execution Commitment of Vision-Language-Action Models}.
\newblock \emph{arXiv preprint arXiv:2605.11567}, 2026{\natexlab{a}}.

\bibitem[Chen et~al.(2026{\natexlab{b}})Chen, Ji, Wheeler, Stocking, and
  Walter]{Chen2026ProAct}
Haoran Chen, Jingtian Ji, Samuel Wheeler, Kaylene~Caswell Stocking, and Matthew
  Walter.
\newblock {Beyond Reconstruction: What Matters in Action Tokenization for Robot
  Policies?}
\newblock \emph{arXiv preprint arXiv:2610.09170}, 2026{\natexlab{b}}.

\bibitem[Chen \& Weng(2026)Chen and Weng]{Chen2026SimToRealBetting}
Yujia Chen and Bowen Weng.
\newblock {Betting for Sim-to-Real Performance Certificates}.
\newblock \emph{arXiv preprint arXiv:2608.21572}, 2026.

\bibitem[Clarke(1981)]{Clarke1981KLT}
R.~J. Clarke.
\newblock {Relation Between the Karhunen--Lo{\`e}ve and Cosine Transforms}.
\newblock \emph{IEE Proceedings F (Communications, Radar and Signal
  Processing)}, 128\penalty0 (6):\penalty0 359--360, 1981.
\newblock \doi{10.1049/ip-f-1.1981.0061}.

\bibitem[Clopper \& Pearson(1934)Clopper and Pearson]{Clopper1934Binomial}
C.~J. Clopper and E.~S. Pearson.
\newblock {The Use of Confidence or Fiducial Limits Illustrated in the Case of
  the Binomial}.
\newblock \emph{Biometrika}, 26\penalty0 (4):\penalty0 404--413, 1934.

\bibitem[Dai et~al.(2025)Dai, Liu, Bai, Liu, Wang, Su, Chen, Lin, and
  Wu]{Dai2025RoVer}
Mingtong Dai, Lingbo Liu, Yongjie Bai, Yang Liu, Zhouxia Wang, Rui Su, Chunjie
  Chen, Liang Lin, and Xinyu Wu.
\newblock {RoVer: Robot Reward Model as Test-Time Verifier for
  Vision-Language-Action Model}.
\newblock \emph{arXiv preprint arXiv:2510.10975}, 2025.

\bibitem[Dong et~al.(2026)Dong, Niu, Liu, Lin, Wang, and Pang]{Dong2026TFGCA}
Shengye Dong, Haochen Niu, Hao Liu, Peiwen Lin, Chuang Wang, and Shanmin Pang.
\newblock {Time-Frequency Geometric Cross-Attention for Chunked
  Vision-Language-Action Models}.
\newblock \emph{arXiv preprint arXiv:2609.09925}, 2026.

\bibitem[Even-Dar et~al.(2006)Even-Dar, Mannor, and
  Mansour]{EvenDar2006ActionElimination}
Eyal Even-Dar, Shie Mannor, and Yishay Mansour.
\newblock {Action Elimination and Stopping Conditions for the Multi-Armed
  Bandit and Reinforcement Learning Problems}.
\newblock \emph{Journal of Machine Learning Research}, 7:\penalty0 1079--1105,
  2006.

\bibitem[Fan(1949)]{Fan1949}
Ky~Fan.
\newblock On a theorem of {Weyl} concerning eigenvalues of linear
  transformations {I}.
\newblock \emph{Proceedings of the National Academy of Sciences}, 35\penalty0
  (11):\penalty0 652--655, 1949.
\newblock \doi{10.1073/pnas.35.11.652}.

\bibitem[Fei et~al.(2026)Fei, Wang, Shi, Dai, Cai, Qian, Ji, He, Zhang, Fei,
  Fu, Gong, and Qiu]{Fei2025LIBEROPlus}
Senyu Fei, Siyin Wang, Junhao Shi, Zihao Dai, Jikun Cai, Pengfang Qian, Li~Ji,
  Xinzhe He, Shiduo Zhang, Zhaoye Fei, Jinlan Fu, Jingjing Gong, and Xipeng
  Qiu.
\newblock {LIBERO-Plus: A Progressive Robustness Benchmark for
  Visual-Language-Action Models}.
\newblock In \emph{IEEE/CVF Conference on Computer Vision and Pattern
  Recognition}, pp.\  38574--38583, 2026.

\bibitem[Floyd(1962)]{Floyd1962Shortest}
Robert~W. Floyd.
\newblock {Algorithm 97: Shortest Path}.
\newblock \emph{Communications of the ACM}, 5\penalty0 (6):\penalty0 345, 1962.
\newblock \doi{10.1145/367766.368168}.

\bibitem[Gabillon et~al.(2012)Gabillon, Ghavamzadeh, and
  Lazaric]{Gabillon2012UGapE}
Victor Gabillon, Mohammad Ghavamzadeh, and Alessandro Lazaric.
\newblock {Best Arm Identification: A Unified Approach to Fixed Budget and
  Fixed Confidence}.
\newblock In \emph{Advances in Neural Information Processing Systems},
  volume~25, pp.\  3212--3220, 2012.

\bibitem[{Gemini Robotics Team} et~al.(2025){Gemini Robotics Team}, Abeyruwan,
  Ainslie, Alayrac, Arenas, Armstrong, Balakrishna, Baruch, Bauza, Blokzijl,
  et~al.]{GeminiRobotics2025}
{Gemini Robotics Team}, Saminda Abeyruwan, Joshua Ainslie, Jean-Baptiste
  Alayrac, Montserrat~Gonzalez Arenas, Travis Armstrong, Ashwin Balakrishna,
  Robert Baruch, Maria Bauza, Michiel Blokzijl, et~al.
\newblock {Gemini Robotics: Bringing AI into the Physical World}.
\newblock \emph{arXiv preprint arXiv:2503.20020}, 2025.

\bibitem[Glasserman \& Yao(1992)Glasserman and
  Yao]{Glasserman1992CommonRandomNumbers}
Paul Glasserman and David~D. Yao.
\newblock {Some Guidelines and Guarantees for Common Random Numbers}.
\newblock \emph{Management Science}, 38\penalty0 (6):\penalty0 884--908, 1992.

\bibitem[G{\"u}nd{\"u}z et~al.(2023)G{\"u}nd{\"u}z, Qin, Aguerri, Dhillon,
  Yang, Yener, Wong, and Chae]{Gunduz2023BeyondBits}
Deniz G{\"u}nd{\"u}z, Zhijin Qin, I{\~n}aki~Estella Aguerri, Harpreet~S.
  Dhillon, Zhaohui Yang, Aylin Yener, Kai~Kit Wong, and Chan-Byoung Chae.
\newblock {Beyond Transmitting Bits: Context, Semantics, and Task-Oriented
  Communications}.
\newblock \emph{IEEE Journal on Selected Areas in Communications}, 41\penalty0
  (1):\penalty0 5--41, 2023.

\bibitem[Guo et~al.(2026{\natexlab{a}})Guo, Chen, Mao, Kenny, Wu, Li, Cai,
  Chen, Ban, Chen, Dou, Yang, Liu, Zhao, and Li]{Guo2026FAFM}
Jianing Guo, Fangzheng Chen, Zihao Mao, Wong Lik~Hang Kenny, Zhenhong Wu,
  Yu~Li, Yishuai Cai, Yuanpei Chen, Yikun Ban, Kai Chen, Qi~Dou, Yaodong Yang,
  Xianglong Liu, Huijie Zhao, and Simin Li.
\newblock {Frequency-Aware Flow Matching for Continuous and Consistent Robotic
  Action Generation}.
\newblock \emph{arXiv preprint arXiv:2606.20135}, 2026{\natexlab{a}}.

\bibitem[Guo et~al.(2026{\natexlab{b}})Guo, Lu, Deng, Wu, Tang, and
  Wang]{Guo2025VLAReasoner}
Wenkai Guo, Guanxing Lu, Haoyuan Deng, Zhenyu Wu, Yansong Tang, and Ziwei Wang.
\newblock {VLA-Reasoner: Empowering Vision-Language-Action Models with
  Reasoning via Online Monte Carlo Tree Search}.
\newblock In \emph{IEEE International Conference on Robotics and Automation},
  pp.\  4990--4997, 2026{\natexlab{b}}.

\bibitem[Jang et~al.(2026)Jang, Kim, Kim, Kim, and
  Shin]{Jang2025VerifierFreeSelection}
Suhyeok Jang, Dongyoung Kim, Changyeon Kim, Youngsuk Kim, and Jinwoo Shin.
\newblock {Verifier-free Test-Time Sampling for Vision-Language-Action Models}.
\newblock In \emph{International Conference on Learning Representations}, 2026.

\bibitem[Jawaid \& Xiang(2026)Jawaid and Xiang]{Jawaid2026NAC}
Ahad Jawaid and Yu~Xiang.
\newblock {NAC: Neural Action Codec for Vision-Language-Action Models}.
\newblock \emph{arXiv preprint arXiv:2606.21372}, 2026.

\bibitem[Jiang \& Wang(2001)Jiang and Wang]{Jiang2001ISS}
Zhong-Ping Jiang and Yuan Wang.
\newblock {Input-to-State Stability for Discrete-Time Nonlinear Systems}.
\newblock \emph{Automatica}, 37\penalty0 (6):\penalty0 857--869, 2001.

\bibitem[Jing et~al.(2025)Jing, Wang, Liu, Tang, Sun, Yao, Wei, Liu, Lu, and
  Ding]{Jing2025MoH}
Dong Jing, Gang Wang, Jiaqi Liu, Weiliang Tang, Zelong Sun, Yunchao Yao, Zhenyu
  Wei, Yunhui Liu, Zhiwu Lu, and Mingyu Ding.
\newblock {Mixture of Horizons in Action Chunking}.
\newblock \emph{arXiv preprint arXiv:2511.19433}, 2025.

\bibitem[Jourdan et~al.(2023)Jourdan, Degenne, and Kaufmann]{Jourdan2023EBTC}
Marc Jourdan, R{\'e}my Degenne, and Emilie Kaufmann.
\newblock {An $\varepsilon$-Best-Arm Identification Algorithm for
  Fixed-Confidence and Beyond}.
\newblock In \emph{Advances in Neural Information Processing Systems},
  volume~36, pp.\  16578--16649, 2023.

\bibitem[Kalyanakrishnan et~al.(2012)Kalyanakrishnan, Tewari, Auer, and
  Stone]{Kalyanakrishnan2012LUCB}
Shivaram Kalyanakrishnan, Ambuj Tewari, Peter Auer, and Peter Stone.
\newblock {PAC Subset Selection in Stochastic Multi-armed Bandits}.
\newblock In \emph{Proceedings of the 29th International Conference on Machine
  Learning}, 2012.

\bibitem[Karnin et~al.(2013)Karnin, Koren, and
  Somekh]{Karnin2013SequentialHalving}
Zohar Karnin, Tomer Koren, and Oren Somekh.
\newblock {Almost Optimal Exploration in Multi-Armed Bandits}.
\newblock In \emph{Proceedings of the 30th International Conference on Machine
  Learning}, pp.\  1238--1246, 2013.

\bibitem[Kim et~al.(2024)Kim, Pertsch, Karamcheti, Xiao, Balakrishna, Nair,
  Rafailov, Foster, Sanketi, Vuong, Kollar, Burchfiel, Tedrake, Sadigh, Levine,
  Liang, and Finn]{Kim2025OpenVLA}
Moo~Jin Kim, Karl Pertsch, Siddharth Karamcheti, Ted Xiao, Ashwin Balakrishna,
  Suraj Nair, Rafael Rafailov, Ethan~P. Foster, Pannag~R. Sanketi, Quan Vuong,
  Thomas Kollar, Benjamin Burchfiel, Russ Tedrake, Dorsa Sadigh, Sergey Levine,
  Percy Liang, and Chelsea Finn.
\newblock {OpenVLA: An Open-Source Vision-Language-Action Model}.
\newblock In \emph{Proceedings of the 8th Conference on Robot Learning}, pp.\
  2679--2713, 2024.

\bibitem[Kim et~al.(2025)Kim, Finn, and Liang]{Kim2025OpenVLAOFT}
Moo~Jin Kim, Chelsea Finn, and Percy Liang.
\newblock {Fine-Tuning Vision-Language-Action Models: Optimizing Speed and
  Success}.
\newblock In \emph{Robotics: Science and Systems XXI}, 2025.

\bibitem[Kim \& Nelson(2006)Kim and Nelson]{Kim2006SelectingBestSystem}
Seong-Hee Kim and Barry~L. Nelson.
\newblock {Selecting the Best System}.
\newblock In Shane~G. Henderson and Barry~L. Nelson (eds.), \emph{Simulation},
  volume~13 of \emph{Handbooks in Operations Research and Management Science},
  chapter~17, pp.\  501--534. Elsevier, 2006.

\bibitem[Kwok et~al.(2025)Kwok, Agia, Sinha, Foutter, Li, Stoica, Mirhoseini,
  and Pavone]{Kwok2025RoboMonkey}
Jacky Kwok, Christopher Agia, Rohan Sinha, Matt Foutter, Shulu Li, Ion Stoica,
  Azalia Mirhoseini, and Marco Pavone.
\newblock {RoboMonkey: Scaling Test-Time Sampling and Verification for
  Vision-Language-Action Models}.
\newblock In \emph{Proceedings of the 9th Conference on Robot Learning}, pp.\
  3200--3217, 2025.

\bibitem[Kwok et~al.(2026)Kwok, Zhang, Xu, Liu, Mirhoseini, Finn, and
  Pavone]{Kwok2026ScalingVerification}
Jacky Kwok, Xilun Zhang, Mengdi Xu, Yuejiang Liu, Azalia Mirhoseini, Chelsea
  Finn, and Marco Pavone.
\newblock {Scaling Verification Can Be More Effective than Scaling Policy
  Learning for Vision-Language-Action Alignment}.
\newblock In \emph{European Conference on Computer Vision}, pp.\  37--55, 2026.

\bibitem[Lei et~al.(2018)Lei, G'Sell, Rinaldo, Tibshirani, and
  Wasserman]{Lei2018Conformal}
Jing Lei, Max G'Sell, Alessandro Rinaldo, Ryan~J. Tibshirani, and Larry
  Wasserman.
\newblock {Distribution-Free Predictive Inference for Regression}.
\newblock \emph{Journal of the American Statistical Association}, 113\penalty0
  (523):\penalty0 1094--1111, 2018.

\bibitem[Li et~al.(2026{\natexlab{a}})Li, Swamy, Bisk, and
  Bajcsy]{Li2026ELASTIC}
Andrew~Zou Li, Gokul Swamy, Yonatan Bisk, and Andrea Bajcsy.
\newblock {ELASTIC: Efficiently Learning to Adaptively Scale Test-Time Compute
  for Generative Control Policies}.
\newblock In \emph{Proceedings of the 10th Conference on Robot Learning},
  2026{\natexlab{a}}.

\bibitem[Li et~al.(2026{\natexlab{b}})Li, Chen, and Huang]{Li2026AVAIVAT}
Boning Li, Yu~Chen, and Longbo Huang.
\newblock {AV-AIVAT: 74x Cheaper Agent Evaluation with Certified Anytime-Valid
  Stopping in Imperfect-Information Games}.
\newblock \emph{arXiv preprint arXiv:2608.06362}, 2026{\natexlab{b}}.

\bibitem[Li et~al.(2026{\natexlab{c}})Li, Mei, Majumdar, Chen, Jiang, and
  Zhu]{Li2026LIBEROVPro}
Huiqiong Li, Zhiting Mei, Anirudha Majumdar, Jingjing Chen, Yu-Gang Jiang, and
  Bin Zhu.
\newblock {LIBERO-VPro: Benchmarking Closed-Loop Visual Robustness of Robotic
  Foundation Models}.
\newblock \emph{arXiv preprint arXiv:2609.24350}, 2026{\natexlab{c}}.

\bibitem[Li et~al.(2026{\natexlab{d}})Li, Su, Cao, Xu, Xia, You, Chen, and
  Xu]{Li2026VLAATTC}
Wenhao Li, Xiu Su, Yichao Cao, Hongyan Xu, Xiaobo Xia, Shan You, Yi~Chen, and
  Chang Xu.
\newblock {VLA-ATTC: Adaptive Test-Time Compute for VLA Models with Relative
  Action Critic Model}.
\newblock In \emph{Proceedings of the 43rd International Conference on Machine
  Learning}, pp.\  69960--69971, 2026{\natexlab{d}}.

\bibitem[Li et~al.(2026{\natexlab{e}})Li, Tang, and Liu]{Li2026FlashVLA}
Zekai Li, Jiaming Tang, and Zhijian Liu.
\newblock {FlashVLA: Streaming Action Decoding for Fast and Asynchronous VLA
  Inference}.
\newblock In \emph{Proceedings of the 10th Conference on Robot Learning},
  2026{\natexlab{e}}.

\bibitem[Lian et~al.(2026)Lian, Yu, Shen, Lin, Du, Zhang, Yang, and
  Chen]{Lian2026ActionPiece}
Shijie Lian, Bin Yu, Zhaolong Shen, Xiaopeng Lin, Yichao Du, Zhirui Zhang,
  Laurence~T. Yang, and Kai Chen.
\newblock {ActionPiece: Rethinking Action Tokenization for Autoregressive
  Vision-Language-Action Models}.
\newblock \emph{arXiv preprint arXiv:2609.18487}, 2026.

\bibitem[Liang et~al.(2026)Liang, Wang, Wang, Wang, Peng, Chen, Chua, and
  Vadakkepat]{Liang2026AAC}
Yuanchang Liang, Xiaobo Wang, Kai Wang, Shuo Wang, Xiaojiang Peng, Haoyu Chen,
  David Kim~Huat Chua, and Prahlad Vadakkepat.
\newblock {Adaptive Action Chunking at Inference-time for
  Vision-Language-Action Models}.
\newblock \emph{arXiv preprint arXiv:2604.04161}, 2026.

\bibitem[Liao et~al.(2026)Liao, Cui, Desingh, and
  Deshwal]{Liao2026ActiveFactorEval}
Andrew Liao, Hanchen Cui, Karthik Desingh, and Aryan Deshwal.
\newblock {Active Real-World Factor-Based Evaluation for Generalist Robot
  Policies}.
\newblock \emph{arXiv preprint arXiv:2607.14439}, 2026.

\bibitem[Liu et~al.(2023)Liu, Zhu, Gao, Feng, Liu, Zhu, and
  Stone]{Liu2023LIBERO}
Bo~Liu, Yifeng Zhu, Chongkai Gao, Yihao Feng, Qiang Liu, Yuke Zhu, and Peter
  Stone.
\newblock {LIBERO: Benchmarking Knowledge Transfer for Lifelong Robot
  Learning}.
\newblock In \emph{Advances in Neural Information Processing Systems},
  volume~36, pp.\  44776--44791, 2023.

\bibitem[Liu et~al.(2026{\natexlab{a}})Liu, Poe, Di, Trivisonno, and
  Caire]{Liu2026ComVLA}
Boliang Liu, Wint~Yi Poe, Jingyun Di, Riccardo Trivisonno, and Giuseppe Caire.
\newblock {ComVLA: Communication-Aware Split Inference for VLA Models in
  6G-Connected Robotics}.
\newblock In \emph{IEEE Global Communications Conference}, 2026{\natexlab{a}}.

\bibitem[Liu et~al.(2026{\natexlab{b}})Liu, Han, Gao, Zhao, Chen, and
  Du]{Liu2026OAT}
Chaoqi Liu, Xiaoshen Han, Jiawei Gao, Yue Zhao, Haonan Chen, and Yilun Du.
\newblock {OAT: Ordered Action Tokenization}.
\newblock In \emph{Robotics: Science and Systems XXII}, 2026{\natexlab{b}}.

\bibitem[Liu et~al.(2026{\natexlab{c}})Liu, Sun, Chao, Yang, Xie, Zhang, Li,
  Tang, Chen, Zhang, and Ding]{Liu2026CheckVLA}
Yushan Liu, Peibo Sun, Xintao Chao, Zhenyang Yang, Yifan Xie, Lingfeng Zhang,
  Shoujie Li, Chenyu Tang, Fang Chen, Xiao-Ping Zhang, and Wenbo Ding.
\newblock {CheckVLA: Execution-Time Verification with Action-Conditioned World
  Model for Long-Horizon Mobile Manipulation}.
\newblock \emph{arXiv preprint arXiv:2607.26789}, 2026{\natexlab{c}}.

\bibitem[Lou et~al.(2026)Lou, Zhang, Yue, Jing, and Si]{Lou2026SafeLoop}
Zeyu Lou, Tianran Zhang, Xinquan Yue, Ya~Jing, and Chenyang Si.
\newblock {SafeLoop: Risk-Aware Rollback for Vision-Language-Action
  Manipulation}.
\newblock In \emph{IEEE/RSJ International Conference on Intelligent Robots and
  Systems}, 2026.

\bibitem[Luo et~al.(2026)Luo, Watson, Sharma, Yang, Qi, Schmerling, Veer,
  Ivanovic, and Pavone]{Luo2026X4Val}
Rachel Luo, Michael Watson, Apoorva Sharma, Heng Yang, Han Qi, Edward
  Schmerling, Sushant Veer, Boris Ivanovic, and Marco Pavone.
\newblock {X4Val: Learning Neural Surrogates for Variance-Reduced Policy
  Evaluation}.
\newblock \emph{arXiv preprint arXiv:2606.05159}, 2026.

\bibitem[Luong et~al.(2026)Luong, Nguyen, Ta, Tran, and Dam]{Luong2026VarDE}
Khang Luong, Nam Nguyen, Hoang Ta, Hung~The Tran, and Tuan~Quang Dam.
\newblock {Variance Driven Exploration: A Provable and Efficient Methodology
  for Pure Exploration in Highly Stochastic Environments}.
\newblock In \emph{Proceedings of the 43rd International Conference on Machine
  Learning}, pp.\  83248--83283, 2026.

\bibitem[Lyu et~al.(2026)Lyu, Chen, and Beksi]{Chen2026ReconVLA}
Zongyao Lyu, Lingling Chen, and William~J. Beksi.
\newblock {SentryVLA: Uncertainty-Guided Action Selection and State Monitoring
  for Vision-Language-Action Robotic Control}.
\newblock In \emph{Proceedings of the 10th Conference on Robot Learning}, 2026.

\bibitem[Maurer \& Pontil(2009)Maurer and Pontil]{Maurer2009EmpiricalBernstein}
Andreas Maurer and Massimiliano Pontil.
\newblock {Empirical Bernstein Bounds and Sample Variance Penalization}.
\newblock In \emph{Proceedings of the 22nd Annual Conference on Learning
  Theory}, 2009.

\bibitem[M{\'e}nard(2019)]{Menard2019GRAD}
Pierre M{\'e}nard.
\newblock {Gradient Ascent for Active Exploration in Bandit Problems}.
\newblock \emph{arXiv preprint arXiv:1905.08165}, 2019.

\bibitem[Nair et~al.(2007)Nair, Fagnani, Zampieri, and Evans]{Nair2007DataRate}
Girish~N. Nair, Fabio Fagnani, Sandro Zampieri, and Robin~J. Evans.
\newblock {Feedback Control Under Data Rate Constraints: An Overview}.
\newblock \emph{Proceedings of the IEEE}, 95\penalty0 (1):\penalty0 108--137,
  2007.

\bibitem[Nakamoto et~al.(2024)Nakamoto, Mees, Kumar, and
  Levine]{Nakamoto2024VGPS}
Mitsuhiko Nakamoto, Oier Mees, Aviral Kumar, and Sergey Levine.
\newblock {Steering Your Generalists: Improving Robotic Foundation Models via
  Value Guidance}.
\newblock In \emph{Proceedings of the 8th Conference on Robot Learning}, pp.\
  4996--5013, 2024.

\bibitem[Peng et~al.(2026)Peng, Ma, Wang, Wang, and Ma]{Peng2026CloudEdgeVLA}
Daojie Peng, Fulong Ma, Bingtao Wang, Sheng Wang, and Jun Ma.
\newblock {Latency-Tolerant Cloud-Edge Collaborative Vision-Language-Action
  Models via Emergent Representational Specialization}.
\newblock \emph{arXiv preprint arXiv:2608.00569}, 2026.

\bibitem[Pertsch et~al.(2025)Pertsch, Stachowicz, Ichter, Driess, Nair, Vuong,
  Mees, Finn, and Levine]{Pertsch2025FAST}
Karl Pertsch, Kyle Stachowicz, Brian Ichter, Danny Driess, Suraj Nair, Quan
  Vuong, Oier Mees, Chelsea Finn, and Sergey Levine.
\newblock {FAST: Efficient Action Tokenization for Vision-Language-Action
  Models}.
\newblock In \emph{Robotics: Science and Systems XXI}, 2025.

\bibitem[{Physical Intelligence}(2025)]{PhysicalIntelligence2025Openpi}
{Physical Intelligence}.
\newblock {openpi}, 2025.
\newblock URL \url{https://github.com/Physical-Intelligence/openpi}.

\bibitem[Regan \& Boutilier(2009)Regan and
  Boutilier]{Regan2009RegretElicitation}
Kevin Regan and Craig Boutilier.
\newblock {Regret-based Reward Elicitation for Markov Decision Processes}.
\newblock In \emph{Proceedings of the 25th Conference on Uncertainty in
  Artificial Intelligence}, pp.\  444--451, 2009.

\bibitem[R{\"o}mer et~al.(2026)R{\"o}mer, Seeliger, Liu, Sturgis, Bagatella,
  Marta, Krause, and Schoellig]{Roemer2026FlowVLAUncertainty}
Ralf R{\"o}mer, Maximilian Seeliger, Saida Liu, Ben Sturgis, Marco Bagatella,
  Daniel Marta, Andreas Krause, and Angela~P. Schoellig.
\newblock {Uncertainty Quantification for Flow-Based Generalist Robot
  Policies}.
\newblock \emph{arXiv preprint arXiv:2606.18043}, 2026.

\bibitem[Russo(2020)]{Russo2020TTTS}
Daniel Russo.
\newblock {Simple Bayesian Algorithms for Best-Arm Identification}.
\newblock \emph{Operations Research}, 68\penalty0 (6):\penalty0 1625--1647,
  2020.

\bibitem[Salla et~al.(2026)Salla, Saravanan, and Stepputtis]{Salla2026POGP}
Rohit~Kumar Salla, Manoj Saravanan, and Simon Stepputtis.
\newblock {Learning When to Stop: Prefix-Optimal Dynamic Diffusion Policies for
  Continuous Control}.
\newblock \emph{arXiv preprint arXiv:2608.05084}, 2026.

\bibitem[Sendai et~al.(2025)Sendai, Alvarez, Matsushima, Matsuo, and
  Iwasawa]{Sendai2025A2C2}
Kohei Sendai, Maxime Alvarez, Tatsuya Matsushima, Yutaka Matsuo, and Yusuke
  Iwasawa.
\newblock {Leave No Observation Behind: Real-time Correction for VLA Action
  Chunks}.
\newblock \emph{arXiv preprint arXiv:2509.23224}, 2025.

\bibitem[Shao et~al.(2022)Shao, Mao, and Zhang]{Shao2022TaskOriented}
Jiawei Shao, Yuyi Mao, and Jun Zhang.
\newblock {Learning Task-Oriented Communication for Edge Inference: An
  Information Bottleneck Approach}.
\newblock \emph{IEEE Journal on Selected Areas in Communications}, 40\penalty0
  (1):\penalty0 197--211, 2022.

\bibitem[Singh et~al.(2026)Singh, Happi, Relan, Nair, and
  Baradwaj]{Singh2026BOKBO}
Anya Singh, Cabrel Happi, Jai Relan, Varun Nair, and Vidyut Baradwaj.
\newblock {BOKBO (Best of K Bad Options): Calibrated Abstention for VLA
  Policies}.
\newblock \emph{arXiv preprint arXiv:2605.30660}, 2026.

\bibitem[Snyder et~al.(2026)Snyder, Badithela, Matni, Pappas, Majumdar, Itkina,
  and Nishimura]{Snyder2026BeyondBinary}
David Snyder, Apurva Badithela, Nikolai Matni, George~J. Pappas, Anirudha
  Majumdar, Masha Itkina, and Haruki Nishimura.
\newblock {Beyond Binary Success: Sample-Efficient and Statistically Rigorous
  Robot Policy Comparison}.
\newblock In \emph{Robotics: Science and Systems XXII}, 2026.

\bibitem[Song et~al.(2025)Song, Chen, Ding, Zhao, Zhao, Zhong, Ge, Li, Wang,
  Ma, Wang, and Li]{Song2025PDVLA}
Wenxuan Song, Jiayi Chen, Pengxiang Ding, Han Zhao, Wei Zhao, Zhide Zhong,
  Zongyuan Ge, Zhijun Li, Donglin Wang, Jun Ma, Lujia Wang, and Haoang Li.
\newblock {PD-VLA: Accelerating Vision-Language-Action Model Integrated with
  Action Chunking via Parallel Decoding}.
\newblock \emph{arXiv preprint arXiv:2503.02310}, 2025.

\bibitem[Su et~al.(2025)Su, Liu, Chen, Zhao, Wu, Li, Xu, Che, and
  Tang]{Su2025FreqPolicy}
Yifei Su, Ning Liu, Dong Chen, Zhen Zhao, Kun Wu, Meng Li, Zhiyuan Xu,
  Zhengping Che, and Jian Tang.
\newblock {FreqPolicy: Efficient Flow-based Visuomotor Policy via Frequency
  Consistency}.
\newblock \emph{arXiv preprint arXiv:2506.08822}, 2025.

\bibitem[Sun et~al.(2026)Sun, Guo, Sun, Wang, Lu, Ji, Ji, Xiong, and
  Meng]{Sun2026PreVLA}
Zhen Sun, Yongjian Guo, Haoran Sun, Luqiao Wang, Wei Lu, Jiachi Ji, Shengzhe
  Ji, Junwu Xiong, and Zhijun Meng.
\newblock {Pre-VLA: Preemptive Runtime Verification for Reliable
  Vision-Language-Action and World-Model Rollouts}.
\newblock \emph{arXiv preprint arXiv:2605.22446}, 2026.

\bibitem[Tang et~al.(2025)Tang, Sun, Zhao, Yang, Lin, Zhang, Hou, Lu, Liu, and
  Han]{Tang2025VLASH}
Jiaming Tang, Yufei Sun, Yilong Zhao, Shang Yang, Yujun Lin, Zhuoyang Zhang,
  James Hou, Yao Lu, Zhijian Liu, and Song Han.
\newblock {VLASH: Real-Time VLAs via Future-State-Aware Asynchronous
  Inference}.
\newblock \emph{arXiv preprint arXiv:2512.01031}, 2025.

\bibitem[Tatikonda \& Mitter(2004)Tatikonda and
  Mitter]{Tatikonda2004CommunicationConstraints}
Sekhar Tatikonda and Sanjoy Mitter.
\newblock {Control Under Communication Constraints}.
\newblock \emph{IEEE Transactions on Automatic Control}, 49\penalty0
  (7):\penalty0 1056--1068, 2004.

\bibitem[Wan et~al.(2026)Wan, Yue, Liu, Chu, Wang, Chen, Jiang, Zhu, Dong, and
  Zhu]{Wan2026NoFreeChecker}
Yang Wan, Xihang Yue, Zhirui Liu, Ziyuan Chu, Shuxun Wang, Yuhan Chen, Xiaonan
  Jiang, Xukun Zhu, Yubo Dong, and Linchao Zhu.
\newblock {No Free Checker: A Survey of Verifiers for Robot Policies}.
\newblock \emph{arXiv preprint arXiv:2609.09250}, 2026.

\bibitem[Wang et~al.(2026{\natexlab{a}})Wang, Xu, Hu, Lin, and
  Gao]{Wang2026SimRealCorrelation}
Shuo Wang, Hanyuan Xu, Yingdong Hu, Fanqi Lin, and Yang Gao.
\newblock {A Practical Recipe Towards Improving Sim-and-Real Correlation for
  VLA Evaluation}.
\newblock \emph{arXiv preprint arXiv:2606.10366}, 2026{\natexlab{a}}.

\bibitem[Wang et~al.(2026{\natexlab{b}})Wang, Lin, Li, Zhang, Yang, Mi, and
  Wei]{Wang2026SVVLA}
Zihua Wang, Zhitao Lin, Ruibo Li, Yu~Zhang, Xu~Yang, Siya Mi, and Xiu-Shen Wei.
\newblock {Open-Loop Planning, Closed-Loop Verification: Speculative
  Verification for VLA}.
\newblock \emph{arXiv preprint arXiv:2604.02965}, 2026{\natexlab{b}}.

\bibitem[Waudby-Smith \& Ramdas(2024)Waudby-Smith and
  Ramdas]{WaudbySmith2024Betting}
Ian Waudby-Smith and Aaditya Ramdas.
\newblock {Estimating Means of Bounded Random Variables by Betting}.
\newblock \emph{Journal of the Royal Statistical Society Series B: Statistical
  Methodology}, 86\penalty0 (1):\penalty0 1--27, 2024.

\bibitem[{XPolicyLab Community} et~al.(2026){XPolicyLab Community}, Chen, Chen,
  Nian, Cai, Chen, Lin, Liang, Chen, Xiang, et~al.]{XPolicyLab2026XPolicyLab}
{XPolicyLab Community}, Tianxing Chen, Yue Chen, Tian Nian, Zijian Cai, Guangyu
  Chen, Wenwei Lin, Qiwei Liang, Zanxin Chen, Peicheng Xiang, et~al.
\newblock {XPolicyLab: A Unified Standard and Open Ecosystem for Robot Policy
  Evaluation and Deployment}.
\newblock \emph{arXiv preprint arXiv:2608.09892}, 2026.

\bibitem[Xu et~al.(2026{\natexlab{a}})Xu, Yu, Xuan, and Lu]{Xu2026VGAS}
Changhua Xu, En~Yu, Junyu Xuan, and Jie Lu.
\newblock {VGAS: Value-Guided Action-Chunk Selection for Few-Shot
  Vision-Language-Action Adaptation}.
\newblock In \emph{Proceedings of the 35th International Joint Conference on
  Artificial Intelligence}, pp.\  5110--5118, 2026{\natexlab{a}}.

\bibitem[Xu et~al.(2026{\natexlab{b}})Xu, Shan, Dai, Wang, and
  Yu]{Xu2026AdaptiveChunking}
Runze Xu, Xiaolong Shan, Shuang Dai, Yu~Wang, and Jincheng Yu.
\newblock {Knowing When to Stop: Adaptive Action Chunking via Internal
  Cross-Attention Dynamics in VLAs}.
\newblock \emph{arXiv preprint arXiv:2609.00908}, 2026{\natexlab{b}}.

\bibitem[Yan et~al.(2026)Yan, Liu, Fan, Cai, Liao, Zhang, and
  Fox]{Yan2026FlexPi}
Ge~Yan, Jinghao Liu, Yuzhi Fan, Lei Cai, Minwen Liao, Jesse Zhang, and Dieter
  Fox.
\newblock {Flex-$\pi$: A Multi-Stream World-Action Model with Compute
  Flexibility}.
\newblock \emph{arXiv preprint arXiv:2608.10860}, 2026.

\bibitem[Yang et~al.(2026)Yang, Zhang, Zhang, Tao, Guo, Ran, Bian, and
  Li]{Yang2026FabriVLA}
Shiyuan Yang, Borong Zhang, Jizheng Zhang, Zhijia Tao, Junfei Guo, Donglai Ran,
  Xu~Bian, and Qingbiao Li.
\newblock {FabriVLA: A Lightweight Vision-Language-Action Model with Conformal
  Action Chunk Uncertainty}.
\newblock \emph{arXiv preprint arXiv:2607.08575}, 2026.

\bibitem[Yang et~al.(2025)Yang, Zhang, He, Pan, Li, Bai, and
  Li]{Yang2025AntiExploration}
Siyuan Yang, Yang Zhang, Haoran He, Ling Pan, Xiu Li, Chenjia Bai, and Xuelong
  Li.
\newblock {Steering Vision-Language-Action Models as Anti-Exploration: A
  Test-Time Scaling Approach}.
\newblock \emph{arXiv preprint arXiv:2512.02834}, 2025.

\bibitem[Yuan et~al.(2026)Yuan, Wu, and Bajcsy]{Yuan2026UPS}
Jessie Yuan, Yilin Wu, and Andrea Bajcsy.
\newblock {When to Act, Ask, or Learn: Uncertainty-Aware Policy Steering}.
\newblock In \emph{Robotics: Science and Systems XXII}, 2026.

\bibitem[Zeng et~al.(2026)Zeng, Agarwal, Bati, Lee, Ancha, and
  Tedrake]{Zeng2026OpenLoop}
Michael Zeng, Abhinav Agarwal, Ajay Bati, Brian Lee, Siddharth Ancha, and Russ
  Tedrake.
\newblock {Revisiting Open-Loop Execution in Robotics: Toward Reactive,
  Higher-Performing Policies}.
\newblock \emph{arXiv preprint arXiv:2608.15938}, 2026.

\bibitem[Zhang et~al.(2025)Zhang, Pfrommer, Pan, Matni, and
  Simchowitz]{Zhang2025Chunking}
Thomas~T. Zhang, Daniel Pfrommer, Chaoyi Pan, Nikolai Matni, and Max
  Simchowitz.
\newblock {Action Chunking and Exploratory Data Collection Yield Exponential
  Improvements in Behavior Cloning for Continuous Control}.
\newblock \emph{arXiv preprint arXiv:2507.09061}, 2025.

\bibitem[Zhao et~al.(2023)Zhao, Kumar, Levine, and Finn]{Zhao2023ACT}
Tony~Z. Zhao, Vikash Kumar, Sergey Levine, and Chelsea Finn.
\newblock {Learning Fine-Grained Bimanual Manipulation with Low-Cost Hardware}.
\newblock In \emph{Robotics: Science and Systems XIX}, 2023.

\bibitem[Zhao et~al.(2026)Zhao, Bogdanovic, Sohal, Tao, Darvish, Aspuru-Guzik,
  Shkurti, and Garg]{Zhao2026DynamicHorizon}
Yuchi Zhao, Miroslav Bogdanovic, Arjun Sohal, Liyu Tao, Kourosh Darvish,
  Al{\'a}n Aspuru-Guzik, Florian Shkurti, and Animesh Garg.
\newblock {Dynamic Execution Horizon Prediction for Chunk-based Robot
  Policies}.
\newblock \emph{arXiv preprint arXiv:2606.11408}, 2026.

\bibitem[Zheng et~al.(2026)Zheng, Cao, Chen, Tian, Li, Li, Sun, Luo, and
  Chen]{Zheng2026RoboECC}
Zihao Zheng, Hangyu Cao, Jiayu Chen, Sicheng Tian, Chenyue Li, Maoliang Li,
  Xinhao Sun, Guojie Luo, and Xiang Chen.
\newblock {RoboECC: Multi-Factor-Aware Edge-Cloud Collaborative Deployment for
  VLA Models}.
\newblock \emph{arXiv preprint arXiv:2603.20711}, 2026.

\bibitem[Zhou et~al.(2025)Zhou, Liao, Huang, Tang, Otto, Jia, Jiang, Hilber,
  Li, Wang, Ya{\u{g}}murlu, Blank, Reuss, and Lioutikov]{Zhou2025BEAST}
Hongyi Zhou, Weiran Liao, Xi~Huang, Yucheng Tang, Fabian Otto, Xiaogang Jia,
  Xinkai Jiang, Simon Hilber, Ge~Li, Qian Wang, {\"O}mer~Erdin{\c{c}}
  Ya{\u{g}}murlu, Nils Blank, Moritz Reuss, and Rudolf Lioutikov.
\newblock {BEAST: Efficient Tokenization of B-Splines Encoded Action Sequences
  for Imitation Learning}.
\newblock \emph{arXiv preprint arXiv:2506.06072}, 2025.

\bibitem[Zouhar et~al.(2026)Zouhar, Kreutzer, Lavie, Kocmi, Post, Bojar, and
  Sachan]{Zouhar2026DynamicAllocation}
Vil{\'e}m Zouhar, Julia Kreutzer, Alon Lavie, Tom Kocmi, Matt Post,
  Ond{\v{r}}ej Bojar, and Mrinmaya Sachan.
\newblock {Dynamically Allocating Evaluation Effort for Model Ranking}.
\newblock In \emph{Findings of the Association for Computational Linguistics:
  EMNLP 2026}, pp.\  7533--7556, 2026.

\end{thebibliography}
\bibliographystyle{iclr2027_conference}

\appendix
\section{Proofs}
\label{app:proofs}
This appendix proves the results of \cref{sec:method,sec:certificate} in the order they appear, then the results that
the certificate protocol of \cref{app:certificate} and the logged-population study of \cref{app:logged} use.

\subsection{Serialized payload\texorpdfstring{ (\cref{prop:payload})}{}}
\proppayload*
\begin{proof}
The stream opens with a four-byte magic string, a version byte, a level count, two 16-bit dimensions, an eight-byte
scale and a 32-byte candidate binding, 50 bytes in all. Geometric bands end at rows $1,2,4,\dots$, so the first
$k\in\{1,2,4,8\}$ rows occupy $1+\log_2k$ bands, each opened by an 11-byte segment record. Each row carries a two-byte
frequency index and its $m$ coefficients. A zigzag varint maps $q$ to $2q$ for $q\ge0$ and to $-2q-1$ for $q<0$ and
writes one byte exactly when $-64\le q\le63$. The total is $8\,[50+11(1+\log_2k)+(2+m)k]$ bits, which is
\cref{eq:exact-price} with the stated widths, and no field depends on the coefficient values beyond the one-byte
condition. The leading $d$ steps coded with horizon $d$ produce the bands ending at $1,2,\dots,d$ and $d$ rows of $m$
coefficients, the field counts of a $d$-row prefix, so they cost $\ell(d)$.
\end{proof}

\subsection{Calls and bits\texorpdfstring{ (\cref{prop:counts})}{}}
\propcounts*
\begin{proof}
The controller executes the arrived steps before it requests the next chunk, so the arrivals partition the $H$ executed
steps into blocks of $d$ steps and a final partial block: $\lceil H/d\rceil$ arrivals, one call each, and $\ell(d)$ bits
per arrival. When $d$ and $L$ divide $H$, the call counts are $H/d$ and $H/L$, with ratio $L/d$, and the bit totals are
$H\ell(d)/d$ and $H\ell(L)/L$, with ratio $(L/d)\,\ell(d)/\ell(L)$, which is at most one exactly when $\ell(d)/d\le \ell(L)/L$.
\end{proof}

\subsection{Energy of a frequency prefix\texorpdfstring{ (\cref{lem:energy})}{}}
\lemenergy*
\begin{proof}
Write $X=C^{\top}Z$ with $C$ orthogonal. Then $X-C^{\top}\Pi_kZ=C^{\top}(I-\Pi_k)Z$, and orthogonality gives
$\|X-C^{\top}\Pi_kZ\|_F^2=\|(I-\Pi_k)Z\|_F^2=\sum_{r>k}\|Z_r\|^2$. Since $Z_r=\phi_r^{\top}X$, $\mathbb
E\|Z_r\|^2=\phi_r^{\top}\Sigma \phi_r=e_r$, which gives the expectation. For a rank-$k$ orthogonal projection $Q$ along time,
$\mathbb E\|X-QX\|_F^2=\operatorname{tr}\Sigma-\operatorname{tr}(Q\Sigma)$, and the Ky Fan maximum
principle~\citep[Theorem~1]{Fan1949} gives $\operatorname{tr}(Q\Sigma)\le\sum_{r\le k}\lambda_r$, with equality for the
projection onto the leading $k$ eigenvectors. A set $S$ of $k$ cosine rows has expected error $\sum_{r\notin S}e_r$,
which non-increasing energies minimize at $S=\{1,\dots,k\}$.
\end{proof}
A cosine prefix approaches the lower bound of the lemma when the chunk covariance is close to that of a strongly
correlated first-order Markov signal~\citep{Ahmed1974DCT,Clarke1981KLT}.

\subsection{Deviation of every-chunk truncation\texorpdfstring{ (\cref{prop:iss})}{}}
\propiss*
\begin{proof}
Write $x_{t+1}=F(x_t,u_k(x_t))$ and $\bar x_{t+1}=G(\bar x_t)$ under the same exogenous randomness, and
$\xi_t=\|x_t-\bar x_t\|_M$. Adding and subtracting $F(x_t,\bar u(x_t))=G(x_t)$,
\begin{equation*}
\xi_{t+1}\le\|F(x_t,u_k(x_t))-F(x_t,\bar u(x_t))\|_M+\|G(x_t)-G(\bar x_t)\|_M\le\gamma\,\zeta_t+\rho\,\xi_t .
\end{equation*}
With $\xi_0=0$, induction gives $\xi_t\le\gamma\sum_{t'<t}\rho^{t-1-t'}\zeta_{t'}$; bounding each $\zeta_{t'}$ by
its supremum and summing the geometric series gives $\gamma\sup_{t'}\zeta_{t'}/(1-\rho)$. The assumption
supplies both inequalities at every step and for every realization, since both trajectories stay in $\mathcal S$.
\end{proof}

\subsection{Exact certificate\texorpdfstring{ (\cref{thm:closure})}{}}
\thmclosure*
\begin{proof}
Summing the edge inequalities along a walk $b=v_0\to\cdots\to v_r=a$ gives $\mu_a-\mu_b\le\sum_t w(v_t,v_{t+1})$, and
minimizing over walks gives $\mu_a-\mu_b\le \dist(b,a)$. Feasibility excludes negative cycles, so $\dist(b,b)=0$. To attain
the bound, set $\mu_v=\dist(b,v)-\dist(b,z)$. The triangle inequality $\dist(b,v)\le \dist(b,u)+w(u,v)$ gives $\mu_v-\mu_u\le
w(u,v)$ for every edge, and $\mu_z=0$, so $\mu\in\mathcal F(w)$ and $\mu_a-\mu_b=\dist(b,a)-\dist(b,b)=\dist(b,a)$. Every feasible
vector has regret $\mu_a-\min_b\mu_b=\max_b(\mu_a-\mu_b)\le\max_b\dist(b,a)=R_w(a)$, and choosing $b=b^\star$ in the
construction attains it.
\end{proof}

\subsection{Validity\texorpdfstring{ (\cref{prop:validity})}{}}
\propvalidity*
\begin{proof}
Let $\mathcal E$ be the event that every sampled edge and every depth edge of the state holds; reference and trivial edges hold
deterministically. On $\mathcal E$ the true vector of failure rates lies in $\mathcal F(w)$, so \cref{thm:closure} gives $\operatorname{reg}(a)\le R_w(a)$, and a
certificate $R_w(a)\le\varepsilon$ implies $\operatorname{reg}(a)\le\varepsilon$. The event of the proposition lies in the complement of $\mathcal E$,
whose probability is at most $\delta_q+\sum_e\delta_e$ by Boole's inequality. Each sampled edge uses an error allocated
at a budget $n$ fixed before any rollout, and the executed option is fixed in advance, so no edge is chosen after its
outcomes are seen; a receiver that tests several budgets in sequence must divide each edge's error over those budgets.
\end{proof}

\subsection{Composition\texorpdfstring{ (\cref{prop:composition})}{}}
\propcomposition*
\begin{proof}
(i) A path fragment $u\to z\to v$ with $u,v\in\mathcal A$ and $v\ne a$ weighs $\bar w$, at least the direct edge $u\to
v$, so replacing such fragments removes every edge out of $z$ other than $z\to a$ from shortest option-to-option paths;
a shortest path into $z$ can end at its first visit to $z$ because feasibility excludes negative cycles. A path from $b$
to $a$ that visits $z$ ends with $z\to a$ after its last visit and has length at least $\dist(b,z)+U_a$, attained by a
shortest path into $z$ followed by $z\to a$; every other path avoids $z$ and has length at least $\dist_{\neg z}(b,a)$.
Taking the smaller class and the largest source gives the formula, and $\dist(b,z)\le0$ because $b\to z$ weighs $0$.
Without paired and depth edges, $\dist(b,z)=0$ and $\dist_{\neg z}(b,a)=\bar w\ge U_a$ for $b\ne a$, and $\dist_{\neg
z}(a,a)=0$; at least one rival exists, so $R_w(a)=U_a$. (ii) Adding edges never lengthens a shortest path; with
$U_a>\varepsilon$, the certificate holds after the addition exactly when every term of the maximum in (i) is at most
$\varepsilon$. (iii) The path $(i,L)\to(i,k)$ followed by a shortest $z$-avoiding path from $(i,k)$ to $a$ avoids $z$ and
has length $w_d+\dist_{\neg z}((i,k),a)$. (iv) No failure of $a$ and no discordance give $n_+=n_-=0$, so the paired weight
is $\overline p(0,n,\delta_p/2)-\underline p(0,n,\delta_p/2)=1-(\delta_p/2)^{1/n}$, and the absolute weight $\overline
p(0,n,\delta_a)=1-\delta_a^{1/n}$ is decreasing in the error, so $\delta_p/2\le\delta_a$ gives the claim.
\end{proof}

\subsection{Results used in the appendices}
\label{app:proofs-appendix}
The depth edge of \cref{app:certificate} can be calibrated by split conformal prediction.
\begin{proposition}[Conformal depth budget]
\label{prop:conformal}
Let $g(x)=\max_i(\mu_{(i,k)}-\mu_{(i,L)})$, let a per-state score $S(x)\in[0,\bar w]$ satisfy $\Pr(g(x)>S(x))\le\delta_s$,
and let the calibration and test scores be exchangeable. With $w_d$ the rank-$\lceil(M+1)(1-\delta_c)\rceil$ score of $M$
calibration states, or $w_d=\bar w$ when that rank exceeds $M$, $\Pr(g(x)>w_d)\le\delta_s+\delta_c$.
\end{proposition}
\begin{proof}
Split conformal calibration~\citep{Lei2018Conformal} gives $\Pr(S(x)>w_d)\le\delta_c$, and the support bound covers
$w_d=\bar w$. The event $g(x)>w_d$ lies in the union of $g(x)>S(x)$ and $S(x)>w_d$.
\end{proof}

The logged-population study of \cref{app:logged} certifies over a fixed population of $N$ outcomes per option.
\begin{proposition}[Finite-population range]
\label{prop:finite-range}
Let fixed outcomes $Y_{a,1},\dots,Y_{a,N}\in[y^-,y^+]$ be revealed on an index set $I_a$ with $n_a=|I_a|$ and
$T_a=\sum_{j\in I_a}Y_{a,j}$. Under any adaptive allocation, the population mean lies in
$[\underline\mu_a,\overline\mu_a]=[(T_a+(N-n_a)y^-)/N,\ (T_a+(N-n_a)y^+)/N]$, and revealing all $N$ outcomes makes both
ends equal to it.
\end{proposition}
\begin{proof}
Each unrevealed summand lies in $[y^-,y^+]$, pathwise, whatever the order of revelation.
\end{proof}
For a same-seed pair $(a,b)$ observed on $I_{ab}$ with joint support $S_{ab}$ of $(Y_a,Y_b)$, the logged paired edge is
$w(b\to a)=N^{-1}\bigl(\sum_{j\in I_{ab}}(Y_{a,j}-Y_{b,j})+(N-|I_{ab}|)\kappa_{ab}\bigr)$ with $\kappa_{ab}=\sup_{(y,y')\in
S_{ab}}(y-y')$.
\begin{proposition}[Rectangle support]
\label{prop:rectangle}
If $I_a=I_b=I_{ab}$ and $S_{ab}$ is the product of the marginal supports $[y_a^-,y_a^+]\times[y_b^-,y_b^+]$, then
$w(b\to a)$ equals the difference of the range ends of \cref{prop:finite-range} under the marginal supports,
$\overline\mu_a-\underline\mu_b$; for $|I_{ab}|<N$ the paired edge is strictly tighter exactly when
$\kappa_{ab}<y_a^+-y_b^-$.
\end{proposition}
\begin{proof}
The observed terms coincide, and each unobserved seed contributes $\kappa_{ab}/N$ to the paired edge and $(y_a^+-y_b^-)/N$ to
$\overline\mu_a-\underline\mu_b$; a rectangle attains $\sup(y-y')=y_a^+-y_b^-$.
\end{proof}
\begin{proposition}[Resolution]
\label{prop:resolution}
Let $Y\in\{0,1\}$ and every observed pair be concordant. If the unobserved seeds admit a discordant outcome, then
$w(b\to a)=(N-|I_{ab}|)/N$, and a certificate at $\varepsilon<1/N$ requires $|I_{ab}|=N$.
\end{proposition}
\begin{proof}
Concordant observations contribute zero and each unobserved seed contributes $1/N$.
\end{proof}
\begin{proposition}[Saturated incumbent]
\label{prop:saturation}
In the two-level setting with $K\ge2$ candidates, nonnegative bit prices and $\mu_{P_1}\le\varepsilon<\bar w/N$ for the
first prefix $P_1$, the prefix-first finite-population rule uses exactly $N$ rollouts and returns $P_1$.
\end{proposition}
\begin{proof}
Initially $P_1$ has the smallest upper endpoint under the prefix-first tie rule. After $n$ of its outcomes,
$U_{P_1}(n)=(T(n)+(N-n)\bar w)/N\le\bar w$, and unqueried prefixes keep upper endpoint $\bar w$ and full candidates at
least $\bar w$, so $P_1$ stays the incumbent. For $n<N$, $U_{P_1}(n)\ge\bar w/N>\varepsilon$, and an unqueried prefix keeps
lower endpoint zero, so it stays active; $K\ge2$ supplies it. At $n=N$, $U_{P_1}(N)=\mu_{P_1}\le\varepsilon$ and every
competing lower endpoint is nonnegative, so the certificate holds exactly at exhaustion under every reveal order.
\end{proof}
\begin{theorem}[Fresh-certificate sample size]
\label{thm:fresh}
Let fresh paired differences $\Delta=Y_a-Y_b$ take values in $[-\kappa,\kappa]$, with $0<2\varepsilon<\kappa$ and $0<\alpha<1$. A rule
that certifies $\mu_a-\mu_b\le\varepsilon$ on the all-nonpositive record of $n$ pairs with error at most $\alpha$ needs
$n\ge\log(1/\alpha)/(-\log(1-2\varepsilon/\kappa))$. The rule that certifies exactly when all $n$ differences are nonpositive
has error at most $\alpha$ once $n\ge\log(1/\alpha)/(-\log(1-\varepsilon/\kappa))$. As $\varepsilon/\kappa\to0$ the two thresholds are
asymptotic to $(\kappa/(2\varepsilon))\log(1/\alpha)$ and $(\kappa/\varepsilon)\log(1/\alpha)$.
\end{theorem}
\begin{proof}
For the lower bound, let $\Delta=\kappa$ with probability $p=2\varepsilon/\kappa$ and $\Delta=0$ otherwise, so
$\mu_a-\mu_b=2\varepsilon>\varepsilon$. Independent pairs give the all-nonpositive record with probability $(1-p)^n$, and a rule
that certifies on it errs at least that often; error at most $\alpha$ forces $(1-p)^n\le\alpha$. For the upper bound,
let $\mathbb E\Delta\ge\varepsilon$. Since $\Delta\le\kappa$, $\varepsilon\le\kappa\Pr(\Delta>0)$, so
$\Pr(\Delta\le0)\le1-\varepsilon/\kappa$, and the zero-record rule errs with probability at most $(1-\varepsilon/\kappa)^n\le\alpha$.
\end{proof}
\begin{proposition}[Deterministic same-seed envelope]
\label{prop:envelope}
Suppose state distances satisfy $\xi_{t+1}\le\theta_1\xi_t+\theta_2e_t$ and absolute stage-cost differences are at most
$\theta_3\xi_t+\theta_4e_t$ for nonnegative constants. With $E_0=\xi_0$, $E_{t+1}=\theta_1E_t+\theta_2e_t$ and
$G_T=\sum_{t<T}(\theta_3E_t+\theta_4e_t)$, two same-seed rollouts started at distance $\xi_0$ satisfy $|Y_a-Y_b|\le
G_T$, and with zero initial distance and constant action error $r$, $G_T(r)=rG_T(1)$.
\end{proposition}
\begin{proof}
Induction gives $\xi_t\le E_t$; summing the stage-cost bounds gives $|Y_a-Y_b|\le G_T$, and linearity in $r$ gives the
scaling.
\end{proof}
\begin{proposition}[Betting failure probability]
\label{prop:betting}
A two-ended ordered-grid scan with two unit-capital pivots, predictable solvent bets and budgets $\alpha_1,\alpha_2\ge0$
has failure probability at most $\alpha_1+\alpha_2$ on a finite rational history tree, under the declared
conditional-mean model.
\end{proposition}
\begin{proof}
For a budget $\alpha>0$, let $M_t$ be a nonnegative supermartingale with $M_0=1$ and $\tau=\inf\{t:M_t\ge1/\alpha\}$;
optional stopping at $\tau\wedge T$ gives $\mathbb E M_{\tau\wedge T}\le1$, so $\Pr(\tau\le T)\le\alpha$ by Markov's
inequality. Applying this to the two pivots that bracket the true mean and adding gives $\alpha_1+\alpha_2$; the scan
keeps the true mean whenever both pivots survive, and integer wealth rounded downward keeps the supermartingale
inequality.
\end{proof}

\section{Codec fields, precision and time}
\label{app:codec}
This appendix supports the code and the accounting of \cref{sec:code,sec:accounting}.

\subsection{Rounding, row removal and coefficient width}
\begin{figure}[t]
\centering
\includegraphics[width=\textwidth]{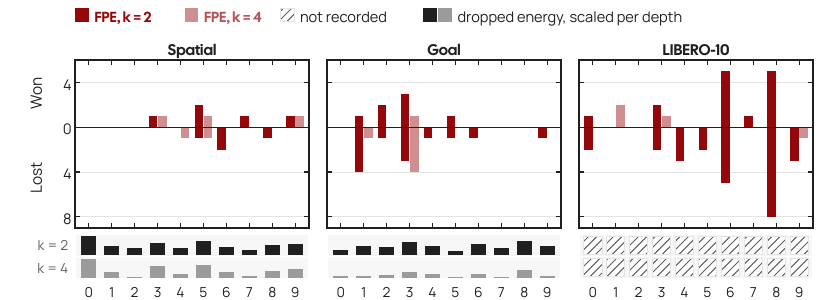}
\caption{Where whole-episode success moves with OpenVLA-OFT at seed 7 (\cref{sec:losses}). Bars count the paired episodes won and lost by FPE
against the quantized full code, 50 pairs per task; meters show the share of the six-axis energy that each depth drops on
the task's quantized-full chunks, scaled per depth (largest 1.2\% at two rows and 0.08\% at four).}
\label{fig:episodes}
\end{figure}

\Cref{eq:quantized-residual} follows from $Z-\Pi_k\widetilde Z=(I-\Pi_k)Z+\Pi_k(Z-\widetilde Z)$: the two summands have
disjoint row supports, and orthogonality keeps their squared norms through the inverse transform. Each kept coefficient
has rounding error at most $1/(2s)$, which gives the bound $km/(4s^2)$. The same argument with $\widetilde X_L=C^\top\widetilde Z$ gives
\cref{eq:quantized-tail}, the quantity that the spectral measurements of \cref{fig:episodes} report. Both identities hold for the decoded chunk
before clipping to the action box and before conversion of the gripper channel; a common Euclidean projection onto a
convex action box keeps the upper bound. A coefficient $q$ is written as a zigzag varint of
$\nu(q)=\max\{1,\lceil\log_2(v(q)+1)/7\rceil\}$ bytes, with $v(q)=2q$ for $q\ge0$ and $v(q)=-2q-1$ for $q<0$, so a
message of the row set $I_k$ spanning $g(k)$ bands has
\begin{equation}
\ell_s(X,k)=400+88\,g(k)+16k+8\sum_{r\in I_k}\sum_{j=1}^{m}\nu\bigl(\operatorname{round}(s(CX)_{rj})\bigr)
\label{eq:variable-payload}
\end{equation}
bits. At $s=10$ every recorded coefficient fits one byte, the condition of \cref{prop:payload}. A larger scale lowers
the worst-case squared rounding error as $s^{-2}$ and lengthens the varints at discrete thresholds.

\begin{figure}[t]
\centering
\includegraphics[width=\textwidth]{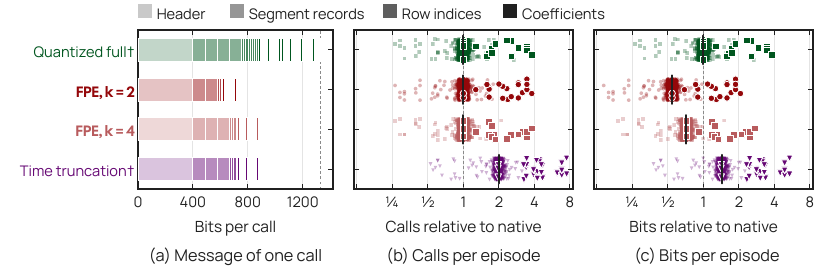}
\caption{What each OpenVLA-OFT execution pays on Spatial and Goal at seed 7 (\cref{sec:accounting}). (a) Bits of one message by field; dashed, native execution priced
as a full-chunk message. (b) Calls and (c) bits of each of the 1,000 episodes of an execution, relative to native
execution from the same task and initial state; solid marks are failures, bars medians.}
\label{fig:payload}
\end{figure}

The 400-bit header is 55.6\% of a two-row message and 30.1\% of an eight-row message (\cref{fig:payload}a); without it the two- and four-row
ratios to the full code would be 0.345 and 0.595, a shorter header needs the shape, scale and candidate binding from an
agreed transport context, and $\beta_t$ transport bits per message move the two-row ratio to
$(720+\beta_t)/(1{,}328+\beta_t)$. At payload
rate $R$ two rows serialize $608/R$ seconds faster than eight per call, 0.608\,ms at 1\,Mbit/s and 0.061\,ms at
10\,Mbit/s; beside the 260--322\,$\mu$s encode time of \cref{tab:precision}, which native execution does not pay, the net
saving turns negative above about \CodecBreakEvenMbps\,Mbit/s.

\subsection{Precision and rows on a frozen two-task panel}
\label{app:precision}
A frozen panel crosses the quantization scale $s\in\{1,10,100\}$ with $k\in\{2,8\}$ on Goal tasks 0 and 9 at official
initial states 0--9 and seed 7, with the checkpoint, revisions and eight-step execution of the primary experiment: 120
episodes, all complete (\cref{tab:precision}). At $s=1$ both row counts fail all twenty episodes, so the failure lies in
full-code quantization; at $s=10$ and $s=100$ both succeed on all twenty, with no discordant pair. At $s=100$ the
varints lengthen to 738 and 1,347 bits per call. Above $s=1$ the panel sits at its ceiling, so it locates
the precision failure at $s=1$.

\begin{table}[t]
\centering\small\setlength{\tabcolsep}{4pt}
\caption{The frozen precision panel at three scales $s$, twenty matched Goal episodes per condition. Error is
$\|X-\widetilde X_L\|_F$ (quantization) and $\|X-\widetilde X_k\|_F$ (total) before clipping and gripper conversion,
averaged over the calls of an episode and then the episodes; codec times are elapsed microseconds per call on the
evaluation CPU.}
\label{tab:precision}
\begin{tabular}{@{}rlrrrrrrr@{}}
\toprule
& & \multicolumn{3}{c}{Twenty episodes} & \multicolumn{2}{c}{Error} & \multicolumn{2}{c}{Codec ($\mu$s)} \\
\cmidrule(lr){3-5}\cmidrule(lr){6-7}\cmidrule(l){8-9}
$s$ & Execution & Successes & Calls & Bits per call & Quantization & Total & Encode & Decode \\
\midrule
1 & FPE, $k=2$ & 0 & 760 & 720 & 0.778 & 0.793 & 261.4 & 54.1 \\
1 & Quantized full$^{\dagger}$ & 0 & 760 & 1,328 & 0.779 & 0.779 & 321.7 & 79.1 \\
\midrule
10 & FPE, $k=2$ & 20 & 351 & 720 & 0.121 & 0.173 & 269.0 & 56.9 \\
10 & Quantized full$^{\dagger}$ & 20 & 348 & 1,328 & 0.122 & 0.122 & 260.1 & 80.1 \\
\midrule
100 & FPE, $k=2$ & 20 & 348 & 738 & 0.020 & 0.127 & 275.0 & 58.0 \\
100 & Quantized full$^{\dagger}$ & 20 & 354 & 1,347 & 0.020 & 0.020 & 262.4 & 83.0 \\
\bottomrule
\end{tabular}

\end{table}

Encoding covers the cosine transform, rounding, the progressive message and the selected prefix; decoding covers header
validation, parsing and the inverse transform; policy inference, simulator steps, candidate binding and network transfer
are outside these times. The codec runs locally and hands complete messages to the receiver.

\section{Whole-episode registration and controls}
\label{app:registered}
This appendix gives the design behind \cref{sec:episodes}, the further seeds, the second policy family, the codes of
matched size behind \cref{sec:matched} and three further arms: the constant row, a discrete gripper and the trailing steps
of each chunk.

\subsection{The registered design}
The closed-loop runs use the official OpenVLA-OFT LIBERO runner, LIBERO and the released checkpoints at the revisions of
\cref{tab:revisions}. Each episode records its success, its policy calls and the serialized bits of every call. The
runner truncates a chunk before it fills the action queue, so time truncation executes indices 0--3 and the full arms
0--7; the upstream option that executes indices 4--7 of an eight-step chunk is the trailing-steps control of
\cref{app:trailing}.

\begin{table}[t]
\centering\small\setlength{\tabcolsep}{6pt}
\caption{Source revisions of the whole-episode runs.}
\label{tab:revisions}
\begin{tabular}{@{}ll@{}}
\toprule
Component & Revision \\
\midrule
OpenVLA-OFT LIBERO runner & \texttt{e4287e94541f459edc4feabc4e181f537cd569a8} \\
LIBERO & \texttt{8f1084e3132a39270c3a13ebe37270a43ece2a01} \\
Spatial checkpoint & \texttt{6d0231af0e48c5985f1ff86908f4674b84bc049b} \\
Goal checkpoint & \texttt{c2d0f9fbbd82674683b397ff923168a12f6a307b} \\
LIBERO-10 checkpoint & \texttt{95220f9a3421a7ff12d4218e73d09ade830fa9a3} \\
Object checkpoint & \texttt{4c89574e1c538b6c102f43f0526d60a9d3650148} \\
openpi, serving $\pi_{0.5}$ & \texttt{215abfb217dbac7d5f1273282331b9b1866c0479} \\
\bottomrule
\end{tabular}
\end{table}

The every-chunk comparison was registered for Spatial and Goal and frozen under version control before its first
frequency-arm episode: seed 7, ten tasks per suite, fifty official initial states per task, arms paired by suite, task,
initial-state index and seed. The primary contrasts are each FPE arm
against the quantized full code, which isolates row removal from quantization, and the quantized full code against
native execution (\cref{tab:episodes-full}), each tested by \cref{eq:sign-flip}. The enumeration is exact conditional on the observed magnitudes; reading it at the null boundary needs joint invariance under sign
changes, for which independent, symmetric task differences suffice. The operating-point rule keeps two rows unless Goal
success falls more than two points below the quantized full code; it selects a row count from the observed change, and
the test is a separate assessment. LIBERO-10 ran later under the same test. Its time-truncation arm ran on 25 October 2026 under a protocol frozen before
the run, in a runtime rebuilt with the package versions of the other LIBERO-10 arms and OSMesa from Mesa 23.2.1. In that
runtime a native rerun of all 500 episodes repeats the outcome and the call count of \OftLiberoTenRerunIdentical{} stored
native episodes and reaches \OftLiberoTenRerunSuccess\% success, with \OftLiberoTenRerunWins{} paired wins and
\OftLiberoTenRerunLosses{} losses against the stored episodes. Against this rerun, in the same runtime, time truncation
changes success by \OftLiberoTenTimeRerunDelta{} points with \OftLiberoTenTimeRerunWins{} paired wins and
\OftLiberoTenTimeRerunLosses{} losses. On
LIBERO-10 two arms were tested, and the four-row $P=0.0020$ stays below $0.0125$, the Bonferroni level for two tests.

\begin{table}[t]
\centering\small\setlength{\tabcolsep}{3.6pt}
\caption{Every whole-episode execution of OpenVLA-OFT at seed 7 with its paired contrast: FPE against the quantized full
code, that code and time truncation against native execution, on the same task, initial state and seed; $P$ is the
two-point test of \cref{eq:sign-flip}. $^\ddagger$Priced by \cref{eq:exact-price}.}
\label{tab:episodes-full}
\begin{tabular}{@{}lrrrrrrrr@{}}
\toprule
& & \multicolumn{4}{c}{Paired contrast} & \multicolumn{3}{c}{Cost} \\
\cmidrule(lr){3-6}\cmidrule(l){7-9}
Execution & Success & $\Delta$ & Wins & Losses & $P$ & Calls & Bits per call & Bit ratio \\
& (\%) $\uparrow$ & (points) & & & & & & $\downarrow$ \\
\midrule
\multicolumn{9}{@{}l}{\emph{Spatial}} \\
Native full & 98.4 & -- & -- & -- & -- & 6,652 & 1,328\rlap{$^{\ddagger}$} & 1.00 \\
Quantized full$^{\dagger}$ & 98.6 & $+0.2$ & 6 & 5 & 0.0039\,\pass & 6,665 & 1,328 & 1.00 \\
FPE, $k=2$ & 98.8 & $+0.2$ & 5 & 4 & 0.012\,\pass & 6,646 & 720 & 0.54 \\
FPE, $k=4$ & 98.8 & $+0.2$ & 3 & 2 & 0.0020\,\pass & 6,637 & 952 & 0.72 \\
Time truncation$^{\dagger}$ & 98.2 & $-0.2$ & 7 & 8 & -- & 13,164 & 952\rlap{$^{\ddagger}$} & 1.42 \\
\midrule
\multicolumn{9}{@{}l}{\emph{Object}} \\
Native full & 97.6 & -- & -- & -- & -- & 8,809 & 1,328\rlap{$^{\ddagger}$} & 1.00 \\
Quantized full$^{\dagger}$ & 94.8 & $-2.8$ & 3 & 17 & 0.83\,\fail & 9,147 & 1,328 & 1.04 \\
FPE, $k=2$ & 95.8 & $+1.0$ & 13 & 8 & $9.8\times10^{-4}$\,\pass & 8,988 & 720 & 0.55 \\
FPE, $k=4$ & 94.8 & $0.0$ & 4 & 4 & 0.027\,\fail & 9,096 & 952 & 0.74 \\
\midrule
\multicolumn{9}{@{}l}{\emph{Goal}} \\
Native full & 97.0 & -- & -- & -- & -- & 7,295 & 1,328\rlap{$^{\ddagger}$} & 1.00 \\
Quantized full$^{\dagger}$ & 97.6 & $+0.6$ & 12 & 9 & 0.0020\,\pass & 7,317 & 1,328 & 1.00 \\
FPE, $k=2$ & 96.6 & $-1.0$ & 7 & 12 & 0.14\,\fail & 7,396 & 720 & 0.55 \\
FPE, $k=4$ & 96.8 & $-0.8$ & 1 & 5 & 0.074\,\fail & 7,365 & 952 & 0.72 \\
Time truncation$^{\dagger}$ & 95.6 & $-1.4$ & 13 & 20 & -- & 14,924 & 952\rlap{$^{\ddagger}$} & 1.47 \\
\midrule
\multicolumn{9}{@{}l}{\emph{LIBERO-10}} \\
Native full & 93.8 & -- & -- & -- & -- & 16,840 & 1,328\rlap{$^{\ddagger}$} & 1.00 \\
Quantized full$^{\dagger}$ & 92.2 & $-1.6$ & 19 & 27 & 0.45\,\fail & 17,148 & 1,328 & 1.02 \\
FPE, $k=2$ & 90.0 & $-2.2$ & 14 & 25 & 0.66\,\fail & 17,515 & 720 & 0.56 \\
FPE, $k=4$ & 92.6 & $+0.4$ & 3 & 1 & 0.0020\,\pass & 17,128 & 952 & 0.73 \\
Time truncation$^{\dagger}$ & 87.4 & $-6.4$ & 11 & 43 & -- & 36,782 & 952\rlap{$^{\ddagger}$} & 1.57 \\
\bottomrule
\end{tabular}

\end{table}

The rank association between a task's dropped share and its success change in \cref{sec:losses} is Spearman's
coefficient with a permutation $P$ over 100,000 seeded permutations of the task labels, $P=\EnergySuccessPTwo$ at two rows
and $\EnergySuccessPFour$ at four; a rule fixed before the computation reads $|\rho_S|<0.3$ at both depths as no
detectable association.

\subsection{Further seeds}
\label{app:seeds}
The second seed sets the evaluation seed and the LIBERO environment seed to 8; the runner fixes the environment seed at 0
and the released heads are deterministic, so the evaluation seed alone would repeat every rollout. It ran the quantized
full code, two rows and four rows on all 500 official episodes of Spatial, Goal and LIBERO-10 in one rebuilt runtime,
each under a decision rule fixed before it ran: Spatial replicates if both prefixes pass at seed 8, LIBERO-10 if four rows
pass, and Goal if two rows pass. Spatial and LIBERO-10 replicate and Goal does not (\cref{tab:replication}). Goal at seed 9
ran under the rule, fixed before its run, that two rows keep Goal success if they pass over the thirty task--seed units of the seed-7 run, seed 8 and seed 9; they do. The seed-8 rule also reported two rows
pooled with the seed-7 Goal arms of the matched-size runtime (rebuilt 7), and the seed-9 rule reports the same pooling
beside its own. Each pooled row sums the paired wins and losses of its seeds, and its $\Delta$ is the mean success change
over the pooled episodes.

\begin{table}[t]
\centering\small
\caption{The two-point test of OpenVLA-OFT prefixes against the quantized full code beyond seed 7, at single seeds over ten
task units and pooled over task--seed units with the seed-7 runs of \cref{tab:episodes-full}.}
\label{tab:replication}
\begin{tabular}{@{}lrrrr@{}}
\toprule
Suite and code & Seeds & $\Delta$ (points) & Wins / losses & $P$ \\
\midrule
\multicolumn{5}{@{}l}{\emph{One seed, 10 task units}} \\
Spatial, two rows & 8 & $+1.0$ & 8 / 3 & 0.0020\,\pass \\
Spatial, four rows & 8 & $\ensuremath{-}0.2$ & 3 / 4 & 0.016\,\pass \\
LIBERO-10, two rows & 8 & $\ensuremath{-}1.2$ & 20 / 26 & 0.32\,\fail \\
LIBERO-10, four rows & 8 & $+0.4$ & 3 / 1 & $9.8\times10^{-4}$\,\pass \\
Goal, two rows & 8 & $\ensuremath{-}0.6$ & 10 / 13 & 0.17\,\fail \\
Goal, four rows & 8 & $\ensuremath{-}0.4$ & 2 / 4 & 0.018\,\pass \\
Goal, two rows & 9 & $+1.2$ & 15 / 9 & 0.0039\,\pass \\
Goal, four rows & 9 & $+0.2$ & 3 / 2 & $9.8\times10^{-4}$\,\pass \\
\midrule
\multicolumn{5}{@{}l}{\emph{Pooled task--seed units}} \\
Spatial, two rows & 7, 8 & $+0.6$ & 13 / 7 & $3.8\times10^{-5}$\,\pass \\
Spatial, four rows & 7, 8 & $0.0$ & 6 / 6 & $5.7\times10^{-5}$\,\pass \\
LIBERO-10, two rows & 7, 8 & $\ensuremath{-}1.7$ & 34 / 51 & 0.40\,\fail \\
LIBERO-10, four rows & 7, 8 & $+0.4$ & 6 / 2 & $1.9\times10^{-6}$\,\pass \\
Goal, two rows & 7, 8 & $\ensuremath{-}0.8$ & 17 / 25 & 0.060\,\fail \\
Goal, two rows, rebuilt 7 & 7, 8 & $\ensuremath{-}0.2$ & 19 / 21 & 0.0088\,\pass \\
Goal, two rows & 7, 8, 9 & $\ensuremath{-}0.1$ & 32 / 34 & 0.0015\,\pass \\
Goal, four rows & 7, 8, 9 & $\ensuremath{-}0.3$ & 6 / 11 & $1.0\times10^{-5}$\,\pass \\
Goal, two rows, rebuilt 7 & 7, 8, 9 & $+0.3$ & 34 / 30 & $1.2\times10^{-4}$\,\pass \\
\bottomrule
\end{tabular}

\end{table}

\subsection{A second policy family}
\label{app:pi05}
The $\pi_{0.5}$ LIBERO checkpoint of openpi (\texttt{pi05\_libero}) is served by openpi and driven by its official LIBERO
client: the 50 official initial states of every task, 224-pixel padded views, the official step limits and five of the
ten predicted steps executed per call. Every call applies the OpenVLA-OFT codec to the five executed steps at scale 10,
and the robot executes the decoded chunk as the client executes the policy's actions; at one byte per coefficient the
quantized full code of five rows sends \PiFiveFullBits{} bits, four rows 952 and two rows 720. Each shard starts a fresh
policy server, so the arms of one task draw the same sampling sequence call by call. The seed sets the client's random
seed and the environment seed, which moves object placements within each initial state. The registered rule names two
rows the operating point of a suite if they pass, otherwise four rows, and the prefix keeps success on a suite with an
operating point. At seed 7 Spatial and Goal have four rows, Object two rows and LIBERO-10 none (\cref{tab:pi05}).
The seed-8 replicate, under the same rule fixed before it ran, has an operating point on every suite, and
pooled over both seeds the operating point is two rows on Spatial, Object and Goal and four rows on
LIBERO-10. The fifth row
holds at most \PiFiveFifthRowShare\% of the quantized energy, so four rows and the quantized full code end identically on every
paired Object, Goal and LIBERO-10 episode at seed 8. The policy server takes \PiFiveSeedSevenPolicyLowMs{} to
\PiFiveSeedSevenPolicyHighMs{}\,ms per call at the median by suite and arm on shared GPUs. Across the forty tasks, the
energy that two rows drop has Spearman $\rho_S=\PiFiveEnergySpearman$ with the task success change at seed 7 (permutation
$P=\PiFiveEnergyP$) and $\PiFiveSeedEightEnergySpearman$ at seed 8 ($P=\PiFiveSeedEightEnergyP$).

\begin{table}[t]
\centering\small
\caption{$\pi_{0.5}$ success in percent under openpi's official LIBERO client, 500 episodes per suite, seed and arm, and the
two-point test of each prefix against the quantized full code.}
\label{tab:pi05}
\begin{tabular}{@{}lrrrrrr@{}}
\toprule
Suite & Native full & Quantized full$^{\dagger}$ & $k=2$ & $P$ & $k=4$ & $P$ \\
\midrule
\multicolumn{7}{@{}l}{\emph{Seed 7}} \\
Spatial & 98.2 & 98.6 & 98.2 & 0.038\,\fail & 98.4 & 0.0020\,\pass \\
Object & 99.6 & 98.4 & 98.8 & 0.016\,\pass & 98.4 & $9.8\times10^{-4}$\,\pass \\
Goal & 95.8 & 96.0 & 96.2 & 0.035\,\fail & 95.6 & 0.0039\,\pass \\
LIBERO-10 & 93.8 & 92.4 & 92.0 & 0.23\,\fail & 92.2 & 0.037\,\fail \\
\midrule
\multicolumn{7}{@{}l}{\emph{Seed 8}} \\
Spatial & 98.6 & 99.0 & 99.0 & 0.0078\,\pass & 98.6 & 0.011\,\pass \\
Object & 99.2 & 98.6 & 98.2 & 0.031\,\fail & 98.6 & $9.8\times10^{-4}$\,\pass \\
Goal & 97.2 & 96.8 & 96.0 & 0.21\,\fail & 96.8 & $9.8\times10^{-4}$\,\pass \\
LIBERO-10 & 92.4 & 93.4 & 93.6 & 0.12\,\fail & 93.4 & $9.8\times10^{-4}$\,\pass \\
\midrule
\multicolumn{7}{@{}l}{\emph{Seeds 7 and 8, 20 task--seed units}} \\
Spatial &  &  &  & $7.2\times10^{-4}$\,\pass &  & $3.8\times10^{-5}$\,\pass \\
Object &  &  &  & $4.9\times10^{-4}$\,\pass &  & $9.5\times10^{-7}$\,\pass \\
Goal &  &  &  & 0.024\,\pass &  & $3.8\times10^{-6}$\,\pass \\
LIBERO-10 &  &  &  & 0.064\,\fail &  & $1.6\times10^{-4}$\,\pass \\
\bottomrule
\end{tabular}

\end{table}

\subsection{Codes of matched size}
\label{app:matched}
The codes of \cref{fig:rate-success} code the same finished OpenVLA-OFT chunk of eight steps and seven action dimensions
under the same header. A prefix of $k$ cosine rows sends $56k$ coefficient bits. The waypoint code sends evenly spaced steps, 56 bits each, and
interpolates linearly between them; one waypoint holds the first step, two the first and last, and four the steps at
positions 0, 2, 5 and 7. Binning sends every step at one to eight bits per value, 56 bits for each bit, with bins uniform between the 1st and
99th percentiles of the checkpoint's action statistics, and decodes bin centers, as OpenVLA's action tokens do at eight bits. Spatial and Goal
ran at seed 7 and LIBERO-10 at seed 8, the seeds of the earlier arms on the same runtime, and every arm is paired with the
quantized full code on the same states (\cref{tab:matched}). Two waypoints on Spatial and Goal and four waypoints ran all
fifty initial states from the start; every other waypoint and binning arm first ran initial states 0--9 of every task and
extended to all fifty exactly when it came within five points of the cosine arm with the smallest payload at or above its own; on Goal, 3-bit binning reached 93\% on its pilot, six points below four rows on
the same initial states, and did not extend.

\begin{table}[t]
\centering\small
\caption{The points of \cref{fig:rate-success}: OpenVLA-OFT success in percent by code and coefficient bits per call, with
the two-point test against the quantized full code; $^{*}$a pilot on ten initial states per task, where no test applies.}
\label{tab:matched}
\begin{tabular}{@{}lrrrr@{}}
\toprule
Code & Bits per call & Spatial & Goal & LIBERO-10 \\
\midrule
One cosine row & 56 & 97.8\rlap{\,\fail} & 97.6\rlap{\,\fail} & 88.2\rlap{\,\fail} \\
Two cosine rows & 112 & 98.8\rlap{\,\pass} & 97.2\rlap{\,\pass} & 90.6\rlap{\,\fail} \\
Four cosine rows & 224 & 98.4\rlap{\,\fail} & 97.4\rlap{\,\pass} & 92.2\rlap{\,\pass} \\
\midrule
One waypoint & 56 & 34.0\rlap{$^{*}$} & 15.0\rlap{$^{*}$} & 6.0\rlap{$^{*}$} \\
Two waypoints & 112 & 97.2\rlap{\,\fail} & 95.8\rlap{\,\fail} & 72.0\rlap{$^{*}$} \\
Four waypoints & 224 & 97.4\rlap{\,\fail} & 96.4\rlap{\,\fail} & 90.2\rlap{\,\fail} \\
\midrule
1-bit binning & 56 & 36.0\rlap{$^{*}$} & 4.0\rlap{$^{*}$} & 0.0\rlap{$^{*}$} \\
2-bit binning & 112 & 88.0\rlap{$^{*}$} & 40.0\rlap{$^{*}$} & 33.0\rlap{$^{*}$} \\
3-bit binning & 168 & 97.8\rlap{\,\fail} & 93.0\rlap{$^{*}$} & 90.4\rlap{\,\fail} \\
4-bit binning & 224 & 98.4\rlap{\,\fail} & 97.6\rlap{\,\pass} & 94.4\rlap{\,\pass} \\
8-bit binning & 448 & 99.0\rlap{\,\pass} & 97.4\rlap{\,\pass} & 95.4\rlap{\,\pass} \\
\midrule
Quantized full$^{\dagger}$ & 448 & 99.0 & 97.0 & 91.8 \\
\bottomrule
\end{tabular}

\end{table}

General-purpose lossless coders price the messages of the \CoderCalls{} calls of the quantized-full episodes of these
runs (\cref{tab:coders}): deflate, LZMA2, zstd and Brotli one message at a time, and deflate, zstd and Brotli as one stream
per episode that keeps its context and flushes every message, with fixed container headers excluded; bzip2 gives the
largest message in every row and is left out. FAST+ prices the same calls over their distinct chunks. The best coder,
deflate over the episode's stream, compresses the full code to \CoderFullBestStream{} bits per call and two rows to
\CoderTwoBestStream, a ratio of \CoderTwoOverFull; FAST+ reaches \CoderFullFast{} and \CoderTwoFast. Two uncoded rows,
112 bits, are smaller than the full code under every general-purpose lossless coder, and at every row count the cosine rows keep within
\CoderKltGapPercent\% of the energy of the chunks' own principal basis. The binned arms' bin indices are priced at the
smaller of their packed and one-byte forms. Over the \CallTimeCalls{} calls of the second-seed and matched-size runs on
shared GPUs the policy takes \CallTimePolicyMedianMs{}\,ms at the median and \CallTimePolicyNinetyMs{}\,ms at the 90th
percentile, and encoding and decoding together take \CallTimeCodecMedianMs{}\,ms.

\begin{table}[t]
\centering\small\setlength{\tabcolsep}{4pt}
\caption{Coefficient bits per call of OpenVLA-OFT messages under lossless coders, one message at a time and as one stream
per episode, and under FAST+; the header, segment records and row indices are excluded.}
\label{tab:coders}
\begin{tabular}{@{}lrrrrrrrrr@{}}
\toprule
& & \multicolumn{4}{c}{One message} & \multicolumn{3}{c}{Stream per episode} & \\
\cmidrule(lr){3-6}\cmidrule(lr){7-9}
Message & Raw & Deflate & LZMA2 & zstd & Brotli & Deflate & zstd & Brotli & FAST+ \\
\midrule
\multicolumn{10}{@{}l}{\emph{Quantized-full calls of Spatial, Goal and LIBERO-10, 38,011 calls}} \\
Full code, eight rows & 448 & 163 & 232 & 266 & 216 & 131 & 204 & 208 & 104 \\
Four rows & 224 & 149 & 218 & 251 & 207 & 120 & 191 & 201 & 91 \\
Two rows & 112 & 116 & 144 & 184 & 143 & 99 & 137 & 135 & 74 \\
One row & 56 & 71 & 88 & 128 & 88 & 66 & 83 & 84 & 46 \\
256-bin action tokens & 448 & 345 & 389 & 463 & 427 & 335 & 367 & 390 & -- \\
\midrule
\multicolumn{10}{@{}l}{\emph{LIBERO-10 calls of the matched-size runs, seed 8}} \\
Four rows & 224 & 139 & 208 & 241 & 196 & 106 & 173 & 186 & -- \\
4-bit binning & 224 & 219 & 253 & 291 & 232 & 192 & 234 & 213 & -- \\
3-bit binning & 168 & 169 & 200 & 240 & 173 & 120 & 169 & 161 & -- \\
2-bit binning & 112 & 122 & 144 & 184 & 144 & 76 & 125 & 127 & -- \\
\bottomrule
\end{tabular}

\end{table}

\subsection{The constant row and a discrete gripper}
\label{app:mechanism}
Two further arms separate the payload from the row count (\cref{tab:mechanism}), each with its own quantized-full
anchor from the same execution batch. The one-row arm executes the constant row alone, the constant chunk at the mean of the
prediction, at 560 bits. The discrete-gripper arm codes the six continuous axes with two rows and packs eight ternary
gripper commands, each the sign of the anchor's gripper value, into sixteen bits, so its payload equals that of the
continuous two-row stream.

\begin{table}[t]
\centering\small\setlength{\tabcolsep}{4.5pt}
\caption{Row count and gripper code, each block against its own anchor from one execution batch. Bold and underline
mark the best and second-best success among the FPE rows of a block.}
\label{tab:mechanism}
\begin{tabular}{@{}lrrrrr@{}}
\toprule
& & \multicolumn{2}{c}{Spatial} & \multicolumn{2}{c}{Goal} \\
\cmidrule(lr){3-4}\cmidrule(l){5-6}
Execution & Bits per call & Success (\%) $\uparrow$ & Calls & Success (\%) $\uparrow$ & Calls \\
\midrule
\multicolumn{6}{@{}l}{\emph{Row count, 500 episodes per arm and suite}} \\
FPE, $k=1$ & 560 & \secondbest{97.8} & 6,730 & \textbf{97.8} & 7,221 \\
FPE, $k=2$ & 720 & \textbf{98.8} & 6,669 & \secondbest{97.0} & 7,320 \\
Quantized full$^{\dagger}$ & 1,328 & 99.0 & 6,632 & 97.0 & 7,327 \\
\midrule
\multicolumn{6}{@{}l}{\emph{Gripper, 400 Spatial and 500 Goal episodes per arm}} \\
FPE, $k=2$, continuous gripper & 720 & \textbf{99.0} & 5,323 & \secondbest{97.0} & 7,320 \\
FPE, $k=2$, ternary gripper & 720 & \secondbest{98.0} & 5,358 & \textbf{97.2} & 7,308 \\
\bottomrule
\end{tabular}

\end{table}

On Goal the constant row wins \OftGoalFreqOneVersusFreqTwoWins{} and loses \OftGoalFreqOneVersusFreqTwoLosses{} paired
episodes against two rows ($P=\OftGoalFreqOneVersusFreqTwoP$, exact sign test) and wins
\OftGoalFreqOneVersusQuantizedWins{} and loses \OftGoalFreqOneVersusQuantizedLosses{} against its anchor
($P=\OftGoalFreqOneVersusQuantizedP$); on Spatial it wins \OftSpatialFreqOneVersusFreqTwoWins{} and loses
\OftSpatialFreqOneVersusFreqTwoLosses{} against two rows ($P=\OftSpatialFreqOneVersusFreqTwoP$). Against the continuous
two-row stream on the same episodes, the ternary gripper changes Spatial success by \GripperSpatialDelta{} points
(\GripperSpatialWins{} gains, \GripperSpatialLosses{} losses, $P=\GripperSpatialP$) and Goal success by
\GripperGoalDelta{} points (\GripperGoalWins{} gain, \GripperGoalLosses{} losses, $P=\GripperGoalP$), by the same exact
sign test.

\subsection{Time truncation and the trailing-steps control}
\label{app:trailing}
Time truncation at $d=4$ calls the policy twice as often (\cref{fig:payload}b), and its bit ratio follows
\cref{prop:counts} because a four-step chunk costs $\ell(4)=952$ bits. Failed episodes run to the step limit, which raises
the call ratio where failures concentrate: Goal task 0 succeeds on 40 of 50 time-truncated episodes against 50 of 50
native ones, at a call ratio of 2.46. Executing each chunk's trailing four steps in place of its
leading four, with the same policy, initial states and schedule, succeeds on 0 of 50 Spatial task 0 episodes (2,750
calls) and on 24 of 50 Goal task 1 episodes (2,325 calls), against 505 and 651 calls at the full horizon, so the
executed steps must be the leading ones; the bit accounting leaves this control out.

\section{Restored states and spectra}
\label{app:restored}
This appendix gives the protocol and the per-depth results behind \cref{sec:restored,sec:spectra}.

\subsection{Single-chunk ladders and the LIBERO-10 panel}
A restored-state test restores a simulator state, executes one truncated chunk and lets the standard policy continue;
the outcome is the task success of the continuation. The ladders fix sixteen states (eight tasks across four suites at
initial states 21 and 28), four candidates per state and four continuation seeds per candidate and depth. LaRA-VLA codes
eight steps and is tested at one, two, four and eight rows; Flex-$\pi$ codes ten steps and is tested at one and ten
rows (\cref{tab:frequency}); the unquantized LaRA-VLA action succeeds on \LadderLaraUnquantizedSuccesses{} of
\LadderLaraTrajectories{}. Ten of the 64 LaRA-VLA candidates equal their complete quantized action at two rows and 61
at four.

\begin{table}[t]
\footnotesize\setlength{\tabcolsep}{3.5pt}
\begin{minipage}[t]{0.69\textwidth}
\centering
\caption{\raggedright Single-chunk tests per depth: continuation success, bits, FAST+ bits of the prefix and
zero-filled tail, and the executed chunk's relative error (64 chunks per bank). Bold and underline mark the best and
second-best success and bits per call among the FPE rows of a block.}
\label{tab:frequency}
\begin{tabular}{@{}lrrrrr@{}}
\toprule
& & & \multicolumn{2}{c}{FAST+ bits} & \\
\cmidrule(lr){4-5}
Execution & Success & Bits per call & Prefix & Zero tail & Error \\
& (\%) $\uparrow$ & $\downarrow$ & & & (\%) \\
\midrule
\multicolumn{6}{@{}l}{\emph{LaRA-VLA ladder, 256 runs}} \\
FPE, $k=1$ & 98.44 & \textbf{560} & 51.0 & 73.0 & 7.9 \\
FPE, $k=2$ & \textbf{99.22} & \secondbest{720} & 79.6 & 89.7 & 4.4 \\
FPE, $k=4$ & \textbf{99.22} & 952 & 90.4 & 101.6 & 3.8 \\
Quantized full$^{\dagger}$ & 99.22 & 1,328 & 101.4 & 101.4 & 3.7 \\
\midrule
\multicolumn{6}{@{}l}{\emph{Flex-$\pi$ ladder, 256 runs}} \\
FPE, $k=1$ & 99.61 & 560 & 52.4 & 74.4 & 11.0 \\
Quantized full$^{\dagger}$ & 98.83 & 1,560 & 117.9 & 117.9 & 3.3 \\
\midrule
\multicolumn{6}{@{}l}{\emph{Flex-$\pi$ LIBERO-10 panel, 10,240 runs}} \\
FPE, $k=2$ & 96.23 & 720 & -- & -- & -- \\
Quantized full$^{\dagger}$ & 95.94 & 1,328 & -- & -- & -- \\
\bottomrule
\end{tabular}

\end{minipage}\hfill
\begin{minipage}[t]{0.27\textwidth}
\centering
\caption{\raggedright Initial states of the forty LIBERO-10 panel states.}
\label{tab:panel-ids}
\begin{tabular}{@{}rrrrr@{}}
\toprule
Task & S1 & S2 & S3 & S4 \\
\midrule
0 & 10 & 15 & 30 & 44 \\
1 & 1 & 17 & 18 & 37 \\
2 & 23 & 25 & 29 & 42 \\
3 & 13 & 21 & 23 & 38 \\
4 & 4 & 22 & 24 & 36 \\
5 & 14 & 15 & 25 & 45 \\
6 & 8 & 17 & 20 & 49 \\
7 & 0 & 3 & 14 & 49 \\
8 & 13 & 21 & 43 & 49 \\
9 & 0 & 5 & 11 & 35 \\
\bottomrule
\end{tabular}

\end{minipage}
\end{table}

The LIBERO-10 panel freezes 160 ten-step candidates of Flex-$\pi$ at one denoising step on forty states
(\cref{tab:panel-ids}). All 160 candidates have two zero rows after quantization, so their eight-row references rebuild
the full quantized action; four candidates keep frequencies 0 and 2 at two rows and the others frequencies 0 and 1. Two
rows give fewer failures than the reference at eight states, equal counts at 27 and more at five, with at most two extra
failures among 256 continuations; the ladders and the panel share LIBERO-10 task 8 at initial state 21, with different
candidates, inference settings and seeds.

\subsection{Row order and reconstruction}
\begin{figure}[t]
\centering
\includegraphics[width=\textwidth]{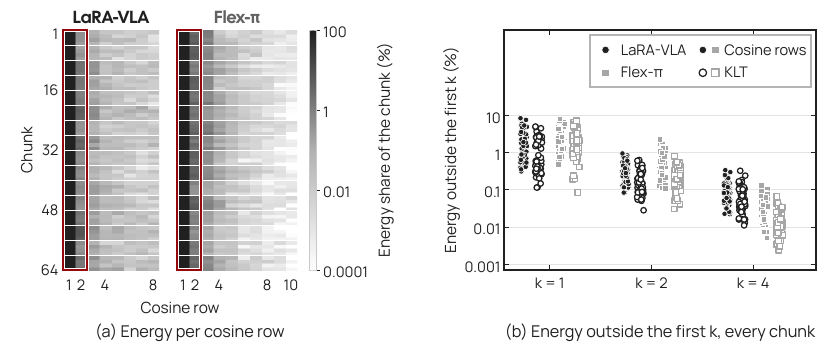}
\caption{Energy of cosine prefixes (\cref{sec:spectra}) on the 64 LaRA-VLA chunks (eight steps) and 64 Flex-$\pi$ chunks (ten steps) of the
restored-state banks, six continuous axes. (a) Energy of each row as a share of its chunk; red frame, the two rows FPE
keeps. (b) Energy outside the first $k$ cosine rows (filled) and outside the $k$ leading Karhunen--Lo\`eve transform (KLT) directions
of the bank (open).}
\label{fig:spectra}
\end{figure}

\begin{figure}[t]
\begin{minipage}[c]{.48\textwidth}
\includegraphics[width=\textwidth]{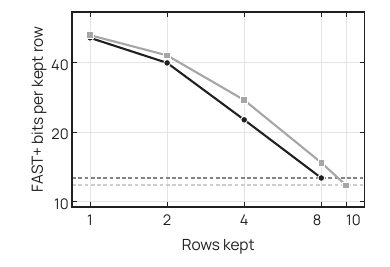}
\end{minipage}\hfill
\begin{minipage}[c]{.48\textwidth}
\caption{\raggedright FAST+ bits per kept row of each prefix of the quantized cosine rows at 11 bits per
token~\citep{Pertsch2025FAST}; dashed, the full code. LaRA-VLA (circles) and Flex-$\pi$ (squares), the 64 chunks of each
restored-state bank.}
\label{fig:fast-price}
\end{minipage}
\end{figure}

Over the 64 chunks of each bank, two cosine rows hold \LaraDctShareTwo\% (LaRA-VLA) and \FlexPiDctShareTwo\% (Flex-$\pi$)
of the pooled six-axis energy (\cref{fig:spectra}), against \LaraKltShareTwo\% and \FlexPiKltShareTwo\% for the two leading KLT directions; on
Flex-$\pi$ the mean relative error of the executed chunk is \FlexPiReconstructionErrorAllTen\% for the full code and
\FlexPiReconstructionErrorTwo\% at two rows. On the ladders two LaRA-VLA rows succeed on
\LadderLaraTwoSuccesses{} of \LadderLaraTrajectories{} continuations, as the full code does, and one Flex-$\pi$ row on
\LadderFlexPiPrefixSuccesses{} against \LadderFlexPiFullSuccesses{} for the ten-row code. The banks send each chunk's quantized rows in decreasing order of energy, where FPE (\cref{sec:code}) sends them in
increasing frequency. For the pooled energies the two orders agree up to six rows; per chunk, they keep the same two
rows for 156 of the 160 LIBERO-10 panel candidates and the same order of all ten rows for 135. Flex-$\pi$ row energies
are monotone, and the eighth LaRA-VLA row exceeds the seventh by less than 0.02\% of total energy, so prefixes of up to
six rows keep the best cosine subset on these data. Removing each chunk's mean lowers the pooled lag-one autocorrelation
of the actions from \LaraLagOneLow--\LaraLagOneHigh{} to \LaraDemeanedLagOneLow--\LaraDemeanedLagOneHigh{} on LaRA-VLA
and from \FlexPiLagOneLow--\FlexPiLagOneHigh{} to \FlexPiDemeanedLagOneLow--\FlexPiDemeanedLagOneHigh{} on Flex-$\pi$.
The squared subspace overlap $\|C_{1:2}V_{1:2}\|_F^2/2$ between the first two cosine rows and the two leading eigenvectors
$V$ of the temporal second-moment matrix is \LaraSubspaceOverlapTwo{} for LaRA-VLA and \FlexPiSubspaceOverlapTwo{} for
Flex-$\pi$. Under FAST+ an executable
prefix also needs the zero-filling interface and the shape of the chunk, since the tokens encode a quantized coefficient
sequence (\cref{tab:frequency}). The largest two-row residual reaches \LaraReconstructionErrorMaxTwo\% of the chunk norm
on LaRA-VLA and \FlexPiReconstructionErrorMaxTwo\% on Flex-$\pi$, against full-code maxima of
\LaraReconstructionErrorMaxEight\% and \FlexPiReconstructionErrorMaxTen\%.

Entropy coding prices every partial prefix above the full code per kept row (\cref{fig:fast-price}): the dropped high rows
are already cheap under FAST+, so a two-row prefix saves less of an entropy-coded coefficient stream than of the fixed-length coefficients. Two
rows save \LaraFastTwoRowSaving\% of the LaRA-VLA coefficient stream, \FlexPiFastTwoRowSaving\% of the Flex-$\pi$ stream
and \OftFastTwoSaving\% of the OpenVLA-OFT stream per call. On the \OftFastChunks{} distinct chunks of the quantized-full
calls of \cref{tab:coders}, FAST+ prices the full code at \OftFastFullBits{} bits, \OftFastFullShare{} of its
\OftFastFixedBits{} fixed-length coefficient bits, and two and four rows at \OftFastTwoRatio{} and \OftFastFourRatio{} of
it.

\section{Certificate arms, algorithms and budgets}
\label{app:certificate}
This appendix gives the edges, arms, algorithms and error allocation of the certificate study of
\cref{sec:certificate-results,sec:certificate-banks}.

\subsection{Edges and their errors}
A signed paired edge keeps the within-seed dependence of its two options, whose positive covariance lowers the
variance of the mean difference, and its coverage follows from its two Clopper--Pearson limits and their errors. The depth edge of a state pools, over the other seven states, the count of seeds on which a prefix fails and its
full code succeeds, and applies the upper Clopper--Pearson limit to that pooled count. \Cref{prop:conformal} gives the
split-conformal calibration of this edge with a joint budget $\delta_q=\delta_s+\delta_c$, which needs at least $1/\delta_c-1$ calibration states
for a finite rank; the implemented conformal calculation sets $\delta_s=\delta_c=0.025$, a joint budget of $0.05$, twice
the $\delta_q=0.025$ of the pooled calculation that \cref{tab:certificate} uses.

\subsection{Arms and protocol}
All arms share the candidate bank, the calibration states, the seeds, the bit accounting and the total error
$\delta=\delta_q+\delta_o=0.05$, split equally. \emph{Absolute}$^\dagger$ keeps the reference and absolute edges: the
certificate that an option's own rollouts give through a Clopper--Pearson interval~\citep{Clopper1934Binomial}, in the
role that betting certificates play in sim-to-real testing~\citep{Chen2026SimToRealBetting}. \emph{Paired and
depth}$^\dagger$ keeps only the paired comparisons that sequential policy and agent comparisons
use~\citep{Snyder2026BeyondBinary,Li2026AVAIVAT}, here with fixed-$n$ Clopper--Pearson limits, composed with the depth
edge. \emph{\pairgap} uses all four edge kinds and splits $\delta_o$ over its $K$ sampled edges. \emph{One-candidate
receipt}$^\dagger$ is \cref{alg:receipt}, which tests absolute edges of candidate 1 at two depths and selects
the prefix on a tie. \emph{Receive full}$^\dagger$ receives every full code and certifies candidate 1's full option
through its absolute edge, as a receiver that waits for complete chunks does~\citep{Black2025RTC,Li2026FlashVLA}.

\begin{algorithm}[t]
\caption{One-candidate receipt at one state}
\label{alg:receipt}
\small
\begin{algorithmic}[1]
\Require candidates $1,\dots,K$, depths $k<L$, budget $n$, tolerance $\varepsilon$, receipt threshold $\eta=0.05$, online error $\delta_o=0.025$
\State fix candidate 1 before any rollout; receive the depth-$k$ prefixes of all $K$ candidates
\State $U_k\gets$ upper Clopper--Pearson bound on $\mu_{(1,k)}$ from $n$ rollouts at error $\delta_o/2$; $a\gets(1,k)$; $R\gets U_k$
\If{$U_k>\eta$}
\State receive the remaining rows of candidate 1; $U_L\gets$ the same bound from $n$ rollouts of $(1,L)$ at error $\delta_o/2$
\If{$U_L<U_k$} \State $a\gets(1,L)$; $R\gets U_L$ \EndIf
\EndIf
\State \Return $a$, the bound $R$, the status $R\le\varepsilon$, the received bits and the rollouts
\end{algorithmic}
\end{algorithm}

LaRA-VLA proposes four candidates at each of eight restored states (official initial state 28 of eight tasks); each
candidate is coded in eight rows, \DepthCertificatePrefixBits{} bits for the one-row prefix and
\DepthCertificateFullBits{} for the full code. Every option has logged continuation rollouts with task failure as the
cost, so $\bar w=1$ and the bit price is zero in the regret. Even seeds feed the certificate and odd
seeds form a held-out audit that recomputes the regret of each executed option. Each budget $n\in\{16,32,48,64\}$ is a
fixed-budget experiment of its own; testing the budgets in sequence would require dividing the error over them. The
replay reads the logged seed lists of the options, which differ: an absolute edge uses the first $n$ even seeds of its
option, and a paired edge the even seeds that both options logged, so \PairedShortFortyEight{} of the
\PairedEdgeCount{} paired edges hold fewer than 48 at $n=48$, besides the shortfall at $n=64$ that
\cref{sec:certificate-results} reports. The rollout counts of \cref{tab:certificate} therefore add the executed
option's outcomes on several seed sets, more than \cref{alg:pairgap} with one shared seed list would spend.

\begin{algorithm}[t]
\caption{\pairgap{} at one arrival}
\label{alg:pairgap}
\small
\begin{algorithmic}[1]
\Require candidates $1,\dots,K$, depths $k<L$, budget $n$, tolerance $\varepsilon$, online error $\delta_o$
\State receive the depth-$k$ prefixes of all $K$ candidates; add reference edges and calibrated depth edges $(i,L)\to(i,k)$
\State fix the executed option $a=(1,k)$ before any rollout and split $\delta_o$ over the sampled edges
\State run $a$ and each received rival $(i,k)$ on the same $n$ seeds; add the absolute edge $z\to a$ and the paired edges
\State compute all shortest paths on $\mathcal A\cup\{z\}$ (Floyd--Warshall) and $R_w(a)=\max_b \dist(b,a)$
\State \Return $a$, the bound $R_w(a)$, the status $R_w(a)\le\varepsilon$, the witness path and the received bits
\end{algorithmic}
\end{algorithm}

\subsection{What decides, and what it costs}
\label{app:witness}
\begin{table}[t]
\centering\small
\caption{Certification on eight LaRA-VLA states at tolerance $0.10$ and total error $0.05$: states certified at 48 and 64
rollouts per sampled edge, at most $n$ shared seeds on a paired edge (none below 48 or at tolerance $0.05$), and, at the first certifying budget $n$, the states
certified at prefix depth with the bits and rollouts per decision. Bold and underline mark the best and second-best
value among the first three arms; the two full-code receipts below them are references, and Receive candidate 1's full
code$^\dagger$ is derived from Receive full$^\dagger$. $^{*}$At $n=64$.}
\label{tab:certificate}
\begin{tabular}{@{}lrrrrrr@{}}
\toprule
& \multicolumn{2}{c}{States at $n=$} & \multicolumn{4}{c}{At the first certifying $n$} \\
\cmidrule(lr){2-3}\cmidrule(l){4-7}
Rule & 48 $\uparrow$ & 64 $\uparrow$ & $n$ $\downarrow$ & Prefix $\uparrow$ & Bits $\downarrow$ & Rollouts $\downarrow$ \\
\midrule
\pairgap & 0 & \secondbest{6} & 64 & 6 & \textbf{2,240} & 319.0 \\
Absolute$^{\dagger}$ & \secondbest{6} & \secondbest{6} & \textbf{48} & 6 & \textbf{2,240} & \textbf{48.0} \\
One-candidate receipt$^{\dagger}$ & \textbf{7} & \textbf{7} & \textbf{48} & 6 & 3,008 & \secondbest{96.0} \\
\midrule
Receive full$^{\dagger}$ & 7 & 7 & 48 & 0 & 5,312 & 48.0 \\
Receive candidate 1's full code$^{\dagger}$ & 7 & 7 & 48 & 0 & 3,008 & 48.0 \\
\midrule
Paired and depth$^{\dagger}$ & 0 & 0 & -- & 0 & 2,240 & 295.8\rlap{$^{*}$} \\
\bottomrule
\end{tabular}

\end{table}

Every certificate on this panel is attained on a reference path $b\to z\to a$, so it rests on the absolute edge. At
$n=64$ the calibrated depth edges are at most \DepthCertificateDepthEdgeHigh, so each unreceived full code sits within
that distance of its received prefix, and the binding sources are the received rivals, whose longest relational paths
under the paired-and-depth error allocation run from \DepthCertificateRelationalLow{} to
\DepthCertificateRelationalHigh{}; the executed option and its rivals succeed on almost every shared seed, and most
paired edges rest on fewer than $n$ shared seeds. On the 80 logged Flex-$\pi$ states of \cref{app:logged}, a prefix and
its full code disagree on a share \RelationDepthDiscordance{} of seeds, against \RelationCandidateDiscordance{} for two
candidates at one depth, and \RelationSaturatedStates{} of the 80 states are saturated, so the depth edge is the
cheapest edge to calibrate. A paired edge turns negative once the rival fails alone on enough seeds, and \cref{prop:composition}(ii) then
certifies $a$ even where its absolute bound stays open; the matched panel has too few such seeds.

\pairgap{} and Paired and depth$^\dagger$ spend 319.0 and 295.8 rollouts per decision at $n=64$ (\cref{tab:certificate}).
One shared seed list for the executed option and three rivals would spend 256, and the replay adds the executed option's
outcomes on the seeds of each pair. At $n=64$ the replay's \PairedEdgeCount{} paired edges hold \PairedSeedsLow{} to
\PairedSeedsHigh{} shared seeds, \PairedShortSixtyFour{} of them fewer than 64. At states \SharedSeedCompletedStates{} we ran every rival one-row prefix on those of the executed option's
first 64 even seeds that it had not logged, \SharedSeedNewRollouts{} new rollouts of the 384 that the three rivals need
at the two states, and \SharedSeedControlsReproduced{} of
\SharedSeedControls{} reruns of logged seeds of the executed option reproduced their outcome and end state. When the
executed option has no failure on its list, a signed edge weighs at most $\overline p(0,64,\delta_p/2)=0.082$ whatever
the rival does, and the depth edge, which pools the logged even seeds of all four candidates at the other seven
states, adds at most \DepthCertificateDepthEdgeHigh, which gives the bound \SharedSeedCertifiedBound{} of state
\SharedSeedCertifiedStates{} (held-out regret \SharedSeedCertifiedAudit) with \SharedSeedCertifiedRollouts{} rollouts per
decision; state \SharedSeedOpenStates{} stays open at \SharedSeedOpenBound, where the executed option fails on
\SharedSeedOpenFailures{} of those seeds. The depth edge reuses logged outcomes of other states once per panel.
The one-candidate receipt saves 2,304 bits against receiving every full code and spends 48 more rollouts per decision.
In a serial model with mean rollout time $t_r$, payload rate $R$ and added codec and certificate time $\Delta t_c$, that
saving lowers latency exactly when $48t_r+\Delta t_c<2{,}304/R$, whose right side is 2.304\,ms at 1\,Mbit/s. With no observed failure, the one-sided exact bound at edge error
$\alpha$ is $U(n)=1-\alpha^{1/n}$, so $U(n)\le\varepsilon$ needs $n\ge\lceil\log\alpha/\log(1-\varepsilon)\rceil$ rollouts:
72 at $\alpha=0.025$ and 86 at $\alpha=0.0125$ for $\varepsilon=0.05$, and 42 at $\alpha=0.0125$ for $\varepsilon=0.10$, which
the budget of 48 covers.

\subsection{The LIBERO-10 and Gaussian banks and the one-candidate bank}
\label{app:banks}
The two further four-candidate banks of \cref{sec:certificate-banks} run every option on one shared list of continuation
seeds, so every paired edge holds its full budget. The Flex-$\pi$ LIBERO-10 bank holds the two-row prefix and the
quantized full code of each of four candidates at the forty panel states, with 64 seeds per option; even seeds certify and
odd seeds audit, the executed option is candidate 1's prefix, and the error split, the edges and the analysis are those
of the LaRA-VLA study, with the depth calibration leaving out the certified state. With 32 certification seeds and the
online error split over the sampled edges, the absolute edge of \pairgap{} is at least 0.147 and a concordant paired edge
at least 0.165, or 0.157 under the error split of Paired and depth$^\dagger$, so no paired path certifies at 0.15 whatever the data, and $\varepsilon=0.20$ is the tolerance at which
the paired edges can act. The relational edges certify no state
that the absolute edge leaves open at any tolerance (\cref{tab:banks}); 25 of the 40 states have no failure on any option.
At $\varepsilon=0.20$ Absolute$^\dagger$ certifies task 6 at initial state 8 and \pairgap{} does not, because \pairgap{}
splits the online error over four bounds.

The Gaussian panel was built to give the relational edges the regime they need. A Gaussian policy whose mean is the
released OpenVLA-OFT Goal head adds noise of scale $0.4\,(q_{99}-q_{01})/2$ to the six arm axes of every chunk, the
smallest scale whose full-code success fell in $[0.40,0.80]$ on a pilot outside the panel. At official initial state 0 of
each Goal task it proposes four candidates, and the bank holds each candidate's two-row prefix and quantized full code.
Every option runs on 128 continuation seeds; after it, the unmodified Gaussian policy continues with noise drawn from
the seed, so every option meets the same noise; even seeds certify and odd seeds audit, and twenty reruns in separate
processes reproduce each success and end state. With four candidates no rule certifies at $\varepsilon\le0.20$, so by the
registered rule the regime is absent at this noise: rival options disagree with the executed prefix on up to 34 of 64
certifying seeds, against 3 to 9 for its own full code.

The one-candidate bank keeps candidate 1's prefix and its own quantized full code, the bank of a deterministic policy, and
runs the analysis unchanged with its candidate count set to one. At the first state of the Gaussian panel, on the 32 certifying seeds of
its first pass, the prefix and its own full code disagreed on 3 seeds and the rival prefixes on 7 to 15. The depth edge pools the other states' counts of seeds on which the prefix fails and
its full code succeeds, here over candidate 1 alone; it stays at or below \OneCandidateLaraDepthMax{} on LaRA-VLA, \OneCandidateFlexPiDepthMax{} on
Flex-$\pi$ and \OneCandidateGaussianDepthMax{} on the Gaussian panel, so it certifies wherever the prefix and its full code
agree, failures included: at task \FlexPiHardTask{}, initial state \FlexPiHardInit{} of Flex-$\pi$ both fail on most seeds
and the graph certifies the prefix against its own full code; its held-out regret against the rival prefixes of the
four-candidate bank is 0.125.

\begin{table}[t]
\centering\small\setlength{\tabcolsep}{4pt}
\caption{Certified states at each tolerance on the three restored-state banks (LaRA-VLA, \mbox{Flex-$\pi$} LIBERO-10 and the
Gaussian OpenVLA-OFT Goal panel) at their largest budget, with four candidates and with the one-candidate bank, and bits
per decision on LaRA-VLA and on the other two banks. Bold and
underline mark the best and second-best value among the compared rules of a block; Receive full$^\dagger$ is a reference.}
\label{tab:banks}
\begin{tabular}{@{}l*{8}{>{\raggedleft\arraybackslash}p{19.5pt}}rr@{}}
\toprule
& \multicolumn{2}{c}{LaRA-VLA, 8} & \multicolumn{3}{c}{Flex-$\pi$ LIBERO-10, 40} & \multicolumn{3}{c}{Gaussian Goal, 10} & \multicolumn{2}{c}{Bits} \\
\cmidrule(lr){2-3}\cmidrule(lr){4-6}\cmidrule(lr){7-9}\cmidrule(l){10-11}
Rule & 0.05 & 0.10 & 0.10 & 0.15 & 0.20 & 0.10 & 0.15 & 0.20 & LaRA & Others \\
\midrule
\multicolumn{11}{@{}l}{\emph{Four candidates}} \\
Receive full$^{\dagger}$ & 0 & 7 & 0 & 32 & 34 & 0 & 0 & 0 & 5,312 & 5,312 \\
One-candidate receipt$^{\dagger}$ & 0 & \textbf{7} & 0 & \textbf{34} & \textbf{35} & 0 & 0 & 0 & 3,008 & 3,488 \\
Absolute$^{\dagger}$ & 0 & \secondbest{6} & 0 & \textbf{34} & \textbf{35} & 0 & 0 & 0 & \textbf{2,240} & \textbf{2,880} \\
Paired and depth$^{\dagger}$ & 0 & 0 & 0 & 0 & 34 & 0 & 0 & 0 & \textbf{2,240} & \textbf{2,880} \\
\pairgap & 0 & \secondbest{6} & 0 & \textbf{34} & 34 & 0 & 0 & 0 & \textbf{2,240} & \textbf{2,880} \\
\midrule
\multicolumn{11}{@{}l}{\emph{One candidate, the bank of a deterministic policy}} \\
Absolute$^{\dagger}$ & 0 & \secondbest{6} & 0 & \secondbest{34} & \secondbest{35} & 0 & 0 & 0 & 560 & 720 \\
\pairgap & \textbf{8} & \textbf{8} & \textbf{40} & \textbf{40} & \textbf{40} & \textbf{10} & \textbf{10} & \textbf{10} & 560 & 720 \\
\bottomrule
\end{tabular}

\end{table}

\section{The logged-population allocation study}
\label{app:logged}
This study applies the shortest-path closure of \cref{sec:certificate}, with finite-population edges, to fixed, already
logged populations of outcomes. It asks
how few rollouts a certificate needs with the finite-population bounds of
\cref{prop:finite-range,prop:rectangle,prop:resolution,prop:saturation} and an optional bit price, and how the selections
fare on fresh seeds.

\subsection{Setup}
Flex-$\pi$~\citep{Yan2026FlexPi} contributes 80 restored states: two tasks from each of Spatial, Object, Goal and
LIBERO-10, drawn by a seeded hash over task names before any outcome was observed, at official initial states 20--29.
LaRA-VLA contributes the sixteen ladder states of \cref{app:restored}. Each state has a nested bank of two and four candidates at prefix and full depth, and each option has eight
logged continuation outcomes with cost $Y=1-g+0.01\min\{h,h_{\max}\}/h_{\max}\in[0,1.01]$ for satisfied-goal fraction $g$,
executed steps $h$ and step limit $h_{\max}$, so $\varepsilon=0.01$ equals the full step-time term. Every selector reads the same logged
populations in the order of a fixed permutation (allocation seed 17; seeds 29 and 43 test the order), certifies regret
over the population at $\varepsilon=0.01$ with bit price $\lambda_b=5\times10^{-8}$ per bit, and freezes its selection;
four fresh continuation seeds per option then give 320 Flex-$\pi$ and 64 LaRA-VLA held-out outcomes per selector. The
Flex-$\pi$ four-candidate population holds 5,120 logged outcomes (2,560 in the nested two-candidate bank) and 2,560 fresh
ones, and the LaRA-VLA population 1,024 logged and 512 fresh ones.
\pairgap{} here reads the incumbent prefix first. Fixed prefix$^\dagger$ and fixed full$^\dagger$ always choose one
depth; tree refinement$^\dagger$ applies confidence-interval action elimination~\citep{EvenDar2006ActionElimination} in a
stop-and-refine tree with $\varepsilon/2$ for the candidate and $\varepsilon/2$ for the depth; uniform$^\dagger$ advances every
population in lockstep; the incumbent--rival rule$^\dagger$ samples the incumbent and the competitor that limits its
certificate; the oracle$^\dagger$ reads every population; and the first-candidate control$^\dagger$ fixes the action
before any rollout. Random draws come from SHA-256 digests of a purpose label and the state, replicates vary the
continuation label, and development states are disjoint from the evaluated ones.

\subsection{Rollouts to a certificate}
\begin{figure}[t]
\centering
\includegraphics[width=\textwidth]{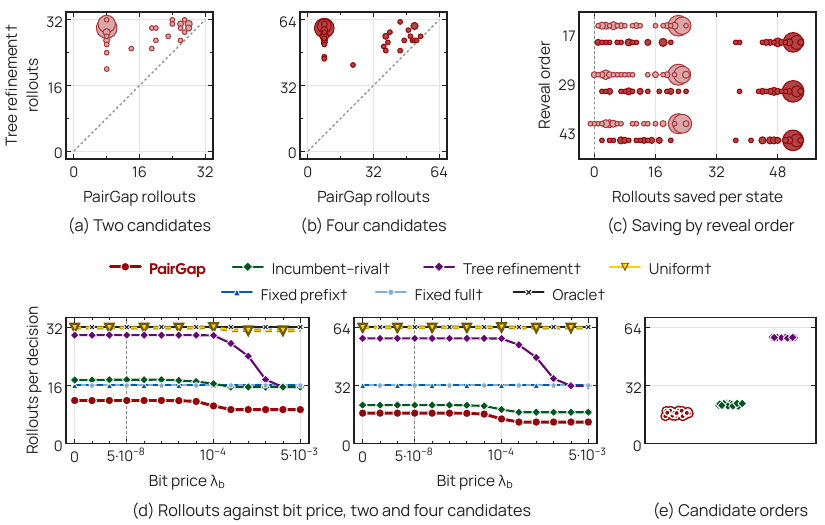}
\caption{Rollouts to an $\varepsilon=0.01$ certificate on the logged Flex-$\pi$ populations, 80 states with eight outcomes per
option, at bit price $5\times10^{-8}$ unless swept. (a, b) \pairgap{} against tree refinement per state under reveal order
17; (c) rollouts saved per state under three reveal orders, two candidates light and four dark, mark area counting states;
(d) rollouts per decision at fourteen bit prices placed at equal steps, two candidates left and four right, the operating
price dashed; (e) mean rollouts for each of 24 candidate orders.}
\label{fig:logged-rollouts}
\end{figure}

At two and four candidates, \pairgap{} certifies with 11.9 and 16.8 rollouts per decision against 29.8 and 57.8 for
tree refinement and 17.6 and 21.2 for the incumbent--rival rule (\cref{fig:logged-rollouts}, \cref{tab:rollouts}). It reads exactly eight
rollouts, the incumbent's whole population, at 61 of 80 states, as \cref{prop:resolution,prop:saturation} predict at
$\varepsilon<1/N$. The saving over tree refinement holds at every state under all three reveal orders at four candidates
and at 236 of 240 state--order pairs at two, with three ties and one state where tree refinement spends one rollout
fewer. Across the fourteen prices \pairgap{} spends the fewest rollouts of the certificate-producing selectors and the
fixed controls; a one-round point rule spends fewer and is $\varepsilon$-optimal at 68 and 66 of 80 states. At the operating price, fixed prefix and fixed full are
$\varepsilon$-optimal at 73 and 79 of 80 states with four candidates, and every certificate-producing selector at all 80.

\begin{table}[t]
\centering\footnotesize\setlength{\tabcolsep}{4pt}
\caption{Rollouts per decision of the logged-population selectors under one stopping rule, by tolerance and over 24
candidate orders at $\varepsilon=0.01$ (mean; share keeping the selection). Bold and underline mark the fewest and
second-fewest rollouts.}
\label{tab:rollouts}
\begin{tabular}{@{}lrrrrrrrr@{}}
\toprule
& \multicolumn{3}{c}{Flex-$\pi$, $\varepsilon=$} & \multicolumn{3}{c}{LaRA-VLA, $\varepsilon=$} & \multicolumn{2}{c}{24 candidate orders} \\
\cmidrule(lr){2-4}\cmidrule(lr){5-7}\cmidrule(l){8-9}
Selector & $0.0025$ & $0.005$ & $0.01$ & $0.0025$ & $0.005$ & $0.01$ & Mean & Same (\%) \\
\midrule
\pairgap & \secondbest{36.6} & \textbf{20.9} & \textbf{16.8} & \textbf{35.2} & \textbf{13.6} & \textbf{8.0} & \textbf{16.7} & 36.6 \\
Incumbent--rival$^{\dagger}$ & \textbf{36.5} & \secondbest{23.0} & \secondbest{21.2} & \secondbest{36.6} & \secondbest{19.5} & \secondbest{18.4} & \secondbest{21.5} & 46.5 \\
Tree refinement$^{\dagger}$ & 57.8 & 57.8 & 57.8 & 59.8 & 59.8 & 59.8 & 58.1 & 91.0 \\
Uniform$^{\dagger}$\rlap{$^{*}$} & 63.5 & 63.5 & 63.5 & 64.0 & 64.0 & 64.0 & 63.6 & 97.9 \\
EB-TC$_\varepsilon$~\citep{Jourdan2023EBTC} & 49.3 & 50.0 & 50.6 & 54.5 & 55.1 & 55.6 & 50.8 & 84.2 \\
Top-two Thompson~\citep{Russo2020TTTS} & 50.3 & 50.2 & 50.2 & 51.2 & 51.2 & 51.2 & 50.0 & 28.6 \\
UGapE~\citep{Gabillon2012UGapE} & 51.4 & 51.3 & 51.3 & 55.6 & 55.6 & 55.6 & 51.6 & 61.1 \\
\bottomrule
\end{tabular}

\footnotesize $^{*}$Under the common stopping rule; \cref{tab:comparison} and \cref{fig:logged-rollouts} stop
uniform allocation by its fixed-confidence rule.
\end{table}

Given one exact stopping rule, modern $\varepsilon$-best-arm rules that balance a leader and a challenger,
EB-TC$_\varepsilon$, top-two Thompson sampling and UGapE, need three to seven times the rollouts (\cref{tab:rollouts}),
because the certificate here closes as soon as the incumbent's own population is read. On LaRA-VLA every state is a saturated incumbent, and \pairgap{} spends 8.0 rollouts
and executes the first candidate.

\subsection{Held-out outcomes}
\begin{figure}[t]
\centering
\includegraphics[width=\textwidth]{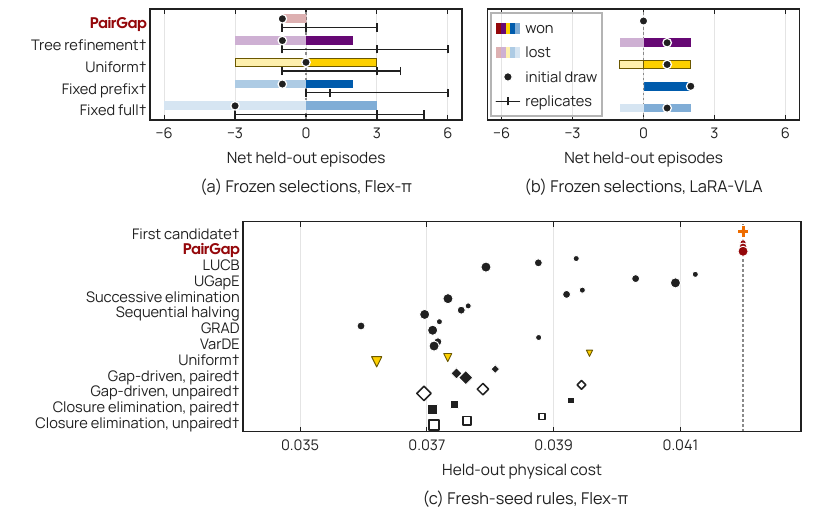}
\caption{Held-out outcomes. (a, b) Frozen selections against the first-candidate control, paired on state and fresh seed over
four seeds per state: wins (hue), losses (tint), the net (dot) and on Flex-$\pi$ the net of three further replicates
(ticks). (c) Held-out physical cost of the fresh-seed rules, which acquire on half of each option's fresh seeds and are
scored on the other half; marks grow with the budget of 32, 64 and 96 rollouts; dashed, the first candidate.}
\label{fig:held-out}
\end{figure}

Frozen selections keep every selector of \cref{tab:comparison} within three Flex-$\pi$ episodes of one another, 307 to
310 of 320 (\cref{fig:held-out}a, b); against the first-candidate control
\pairgap{} loses one episode and wins none, tree refinement and fixed prefix each win two and lose three, and all three
exact binomial tests give $P=1$. Three further continuation-seed replicates give \pairgap{} paired differences of
$-1$, $+3$ and $0$ (304, 310 and 303 successes), a spread as large as any selector's effect. On LaRA-VLA \pairgap{} and the control succeed on the same 62 of 64 outcomes, and fixed prefix reaches 64
with two paired gains and no loss ($P=0.50$). Physical cost,
which also counts partial goals and executed steps, places \pairgap{} last on Flex-$\pi$: its frozen selections cost
more than tree refinement at 39 states, less at 14 and the same at 27.

\begin{table}[t]
\centering\footnotesize\setlength{\tabcolsep}{2.5pt}
\caption{Held-out success, acquisition per decision and physical cost at $\varepsilon=0.01$; $\Delta$ is held-out wins minus losses
against the first-candidate control, and cost the mean over four fresh seeds per state. Bold and underline mark the best
and second-best value of a column within a block.}
\label{tab:comparison}
\begin{tabular}{@{}lrrrrrrrrrr@{}}
\toprule
& \multicolumn{5}{c}{Flex-$\pi$, 320 outcomes} & \multicolumn{5}{c}{LaRA-VLA, 64 outcomes} \\
\cmidrule(lr){2-6}\cmidrule(l){7-11}
Selector & Success & $\Delta$ & Rollouts & Bits & Cost & Success & $\Delta$ & Rollouts & Bits & Cost \\
 & (\%) $\uparrow$ & & $\downarrow$ & $\downarrow$ & $\downarrow$ & (\%) $\uparrow$ & & $\downarrow$ & $\downarrow$ & $\downarrow$ \\
\midrule
\multicolumn{11}{@{}l}{\emph{Certificate-producing selectors}} \\
\pairgap & \secondbest{96.6} & $-1$ & \textbf{16.8} & \textbf{2,340} & 0.0337 & 96.9 & 0 & \textbf{8.0} & \textbf{2,240} & 0.0257 \\
Tree refinement$^{\dagger}$ & \secondbest{96.6} & $-1$ & \secondbest{57.8} & \secondbest{2,415} & \secondbest{0.0316} & \textbf{98.4} & $+1$ & \secondbest{59.8} & \secondbest{2,288} & \textbf{0.0204} \\
Uniform$^{\dagger}$ & \textbf{96.9} & 0 & 63.8 & 2,478 & \textbf{0.0284} & \textbf{98.4} & $+1$ & 64.0 & \secondbest{2,288} & \textbf{0.0204} \\
\midrule
\multicolumn{11}{@{}l}{\emph{Fixed-action controls}} \\
Fixed prefix$^{\dagger}$ & \secondbest{96.6} & $-1$ & \secondbest{32.0} & \textbf{2,240} & \textbf{0.0316} & \textbf{100.0} & $+2$ & \secondbest{32.0} & \textbf{2,240} & \textbf{0.0047} \\
Fixed full$^{\dagger}$ & 95.9 & $-3$ & \secondbest{32.0} & 6,240 & \secondbest{0.0326} & \secondbest{98.4} & $+1$ & \secondbest{32.0} & 5,312 & \secondbest{0.0204} \\
First candidate$^{\dagger}$ & \textbf{96.9} & 0 & \textbf{0.0} & \textbf{2,240} & 0.0327 & 96.9 & 0 & \textbf{0.0} & \textbf{2,240} & 0.0257 \\
\bottomrule
\end{tabular}

\end{table}

\begin{figure}[t]
\centering
\includegraphics[width=\textwidth]{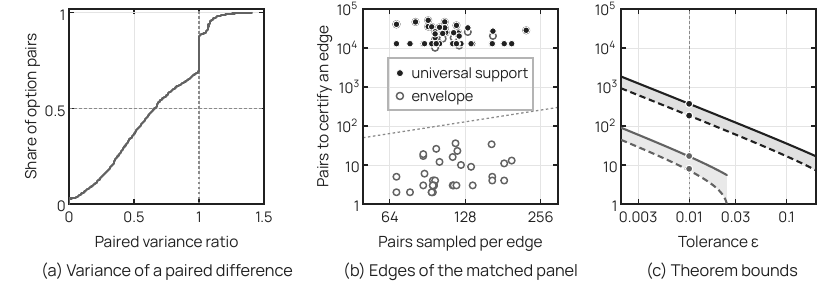}
\caption{Pairing and the fresh-certificate price. (a) $\operatorname{Var}(Y_b-Y_a)/(\operatorname{Var}Y_a+\operatorname{Var}Y_b)$
over shared seeds for 2,239 Flex-$\pi$ option pairs (1 without pairing). (b) Pairs an empirical-Bernstein certificate at
$\varepsilon=0.01$ needs on each of the 56 edges of the matched LaRA-VLA panel, under the universal support
$[-1.01,1.01]$ (ink) and the data-fitted envelope (grey); diagonal, needed equals sampled. (c) Pair counts of the fresh
certificate of \cref{thm:fresh} at $\alpha=1/40$, lower dashed and upper solid.}
\label{fig:pairing}
\end{figure}

\subsection{Fresh sampling and the price of pairing}
\label{app:fresh-sampling}
With fresh seeds (\cref{fig:held-out}c) we run eleven sampling rules, LUCB~\citep{Kalyanakrishnan2012LUCB},
UGapE~\citep{Gabillon2012UGapE}, successive elimination~\citep{EvenDar2006ActionElimination}, sequential
halving~\citep{Karnin2013SequentialHalving}, GRAD~\citep{Menard2019GRAD}, VarDE~\citep{Luong2026VarDE}, uniform
allocation, and gap-driven and closure-elimination rules in paired and unpaired form. Ten of them lower held-out cost
from 0.0420 for the first candidate to 0.0362--0.0379 at 96 rollouts, UGapE stays at 0.0403--0.0412, and paired and unpaired
variants of a rule track one another. Pairing lowers the variance of a difference, with median ratio 0.66 over
2,239 option pairs (\cref{fig:pairing}a), and the width $\kappa$ of the paired support sets the fresh price: \cref{thm:fresh}
gives 185 and 371 pairs per edge at $\alpha=1/40$ under the universal support $\kappa=1.01$, and 8 and 17 under an
envelope $\kappa=0.05$ such as \cref{prop:envelope} supplies. On the matched LaRA-VLA panel (index 28, 12,528 trajectories under seven
sampling rules, 99.2--100\% held-out success at 222--224 rollouts and 5,312 bits), a time-uniform empirical-Bernstein recomputation~\citep{Maurer2009EmpiricalBernstein} certifies 0 of 8 states (a
zero-variance edge needs 12,885 pairs over 112 directional edges) and 2 of 8 with an envelope fitted to the differences. The betting grid~\citep{WaudbySmith2024Betting} of \cref{prop:betting} needs 529 pairs for one edge at
$\alpha=1/40$ and 878 for 28 edges at $\alpha=1/1120$ (empirical Bernstein 10,824 and 12,532, the zero-record rule 371
and 706).

\subsection{Selection at the arrival state}
Selecting at the state the robot reaches during transmission (eight
Flex-$\pi$ states, four arrival ticks, 1,056 assignments) succeeds on 84 of 96 delayed-tick outcomes against 86 at the
request state, with 39.7\% (ticks 2, 5) and 52.0\% (tick 8) less communication in the study's own code; Spatial task 4
loses a quarter of its matched successes and task 5 gains a twelfth.

\end{document}